\documentclass[9.5pt,journal]{IEEEtran}
\usepackage{booktabs}    
\usepackage{amsmath,amsfonts}
\allowdisplaybreaks 
\usepackage{algorithmic}
\usepackage{array}
\usepackage{hyperref}
\usepackage{tabularx} 
\usepackage[caption=false,font=normalsize,labelfont=sf,textfont=sf]{subfig}
\usepackage{textcomp}
\usepackage{stfloats}
\usepackage{url}
\usepackage{verbatim}
\usepackage{graphicx}
\usepackage{mathtools}
\usepackage{pifont}
\usepackage{tikz}
\usepackage{cite}
\usepackage{graphicx}
\usepackage{forest}
\usepackage{listings}
\usepackage{xcolor}
\usepackage{graphicx}
\usepackage{amssymb}
\usepackage[table]{xcolor} 
\forestset{
  forked edges/.style={
    for tree={
      grow=east,
      parent anchor=east,
      child anchor=west,
      edge path'={
        (!u.parent anchor) -- +(5pt,0) |- (.child anchor)
      },
      if level=0{}{
        if={n==1}{edge path'={(!u.parent anchor) -- +(5pt,0) |- (.child anchor)}}{}
      }
    }
  }
}

\makeatletter
\def\IEEEauthorrefmark#1{\raisebox{0pt}[0pt][0pt]{\textsuperscript{#1}}}
\makeatother

\def\BibTeX{{\rm B\kern-.05em{\sc i\kern-.025em b}\kern-.08em
    T\kern-.1667em\lower.7ex\hbox{E}\kern-.125emX}}
\usepackage{balance}
\begin{document}
\title{A Unified Study of LoRA Variants: Taxonomy, Review, Codebase, and Empirical Evaluation}

\author{
\vspace{-8pt}
\IEEEauthorblockN{
Haonan He\IEEEauthorrefmark{1,}\IEEEauthorrefmark{2}\textsuperscript{$^\dagger$},
Jingqi Ye\IEEEauthorrefmark{1,}\IEEEauthorrefmark{2}\textsuperscript{$^\dagger$},
Minglei Li\IEEEauthorrefmark{1,}\IEEEauthorrefmark{3}\textsuperscript{$^\dagger$},
Zhengbo Wang\IEEEauthorrefmark{2,}\IEEEauthorrefmark{5},
Mengqi Li\IEEEauthorrefmark{6}, \\
Tao Chen\IEEEauthorrefmark{3},
Lei Bai\IEEEauthorrefmark{1},
Peng Ye\IEEEauthorrefmark{1,}\IEEEauthorrefmark{3,}\IEEEauthorrefmark{4}\textsuperscript{$^*$}
} \\
\IEEEauthorblockA{
\small
\IEEEauthorrefmark{1}Shanghai Artificial Intelligence Laboratory, Shanghai 200233, China\\
\IEEEauthorrefmark{2}University of Science and Technology of China, Hefei 230026, China\\
\IEEEauthorrefmark{3}Fudan University, Shanghai 200433, China\\
\IEEEauthorrefmark{4}The Chinese University of Hong Kong, Hong Kong SAR 999077, China \\
\IEEEauthorrefmark{5}Institute of Automation, Chinese Academy of Sciences, Beijing 100190, China\\
\IEEEauthorrefmark{6}The Chinese University of Hong Kong, Shenzhen, Shenzhen 518172, China\\
}
\thanks{$^\dagger$ Equal contribution.}
\thanks{$^*$ Correspondence: yepeng@pjlab.org.cn.}
\vspace{-25pt}
}

\markboth{Journal of \LaTeX\ Class Files,~Vol.~18, No.~9, September~2020}%
{How to Use the IEEEtran \LaTeX \ Templates}

\maketitle

\begin{abstract}

Low-Rank Adaptation (LoRA) is a fundamental parameter‑efficient fine‑tuning method that balances efficiency and performance in large-scale neural networks. However, the proliferation of LoRA variants has led to fragmentation in methodology, theory, code, and evaluation. To this end, this work presents the first unified study of LoRA variants, offering a systematic taxonomy, unified theoretical review, structured codebase, and standardized empirical assessment. First, we categorize LoRA variants along four principal axes: rank, optimization dynamics, initialization, and integration with Mixture‑of‑Experts. Then, we review their relationships and evolution within a common theoretical framework focused on low‑rank update dynamics. Further, we introduce LoRAFactory, a modular codebase that implements variants through a unified interface, supporting plug‑and‑play experimentation and fine‑grained analysis. Last, using this codebase, we conduct a large‑scale evaluation across natural language generation, natural language understanding, and image classification tasks, systematically exploring key hyperparameters. Our results uncover several findings, notably: LoRA and its variants exhibit pronounced sensitivity to the choices of learning rate compared to other hyperparameters; moreover, with proper hyperparameter configurations, LoRA consistently matches or surpasses the performance of most of its variants. All code and configurations are publicly available at this \href{https://anonymous.4open.science/r/MyTransformers-4EC3}{link}.
\end{abstract}
\section{Introduction}

\IEEEPARstart{L}{arge-scale} models with billions of parameters, such as large language models (LLMs), which are pretrained on massive corpora, have demonstrated remarkable performance across diverse tasks, transforming fields ranging from natural language processing to multimodal reasoning~\cite{liang2024mllm-survey,li2025lvm-survey,zhao2023llm-survey}. However, full fine-tuning large-scale models is highly resource-intensive, primarily due to the substantial GPU memory required to store optimizer states. To alleviate this burden, numerous parameter-efficient fine-tuning (PEFT) methods have been proposed~\cite{gao2023llama-adapeter-v2,liu2021p-tuning,zaken2021bitfit,karimi2021compacter,hu2022lora}. These approaches drastically reduce memory usage by either minimizing the number of trainable parameters or optimizing the management of optimizer states, especially for adaptive optimizers~\cite{kingma2014adam,loshchilov2017adamw}. 
Consequently, PEFT methods also enhance the training efficiency under distributed frameworks, such as ZeRO~\cite{rajbhandari2020zero} and FSDP~\cite{zhao2023fsdp}, by reducing communication overhead.

Low-Rank Adaptation (\textsc{LoRA})~\cite{hu2022lora} has emerged as one of the most widely adopted PEFT methods. Its popularity stems from strong empirical performance, implementation simplicity, and broad generalization across domains, including parametric knowledge memory~\cite{su2025parametric, zweiger2025self}, multimodal learning~\cite{wei2025moka,wang2025vision-as-lora}, and federated learning~\cite{babakniya2023slora,qi2024fdlora}. 
Despite its efficiency and effectiveness, such as fine-tuning 32B-scale models on a consumer-level GPU through quantization methods~\cite{dettmers2023qlora,li2023loftq}, \textsc{LoRA} still exhibits limitations, such as the low-rank structure, which often results in a performance gap compared to full fine-tuning, particularly on complex downstream tasks.

To bridge this gap, numerous variants of \textsc{LoRA} have been developed, which can be broadly classified as follows: \textit{Rank Adjustment Based Variants} (Section~\ref{rank_variants}) include methods such as \textsc{ReLoRA}~\cite{lialin2307relora}, which composes multiple low-rank update subspaces; \textsc{AdaLoRA}~\cite{zhang2023adalora}, which dynamically masks less important ranks; 
and \textsc{RandLoRA}~\cite{albert2025randlora}, which enables high-rank training via rank-sharing strategies. 
\textit{Optimization Process Adjustment Based Variants} (Section~\ref{training_dynamics_variants}) cover approaches like \textsc{LoRA+}~\cite{hayou2024lora+}, which decouples the learning rates of low-rank weights for optimization stability; 
\textsc{LoRA-Pro}~\cite{wang2024lora-pro}, which reduces the discrepancy with full fine-tuning via parameter update space alignment. \textit{Initialization Adjustment Based Variants} (Section~\ref{initialization_variants}) comprise techniques such as \textsc{PiSSA}~\cite{meng2024pissa}, which applies Singular Value Decomposition (SVD) on pretrained weights to extract dominant features for initializing low-rank weights, and \textsc{LoRA-GA}~\cite{wang2024lora-ga}, which performs SVD on the gradients of pretrained weights for initialization. Lastly, \textit{Mixture-of-Experts (MoE) Integration Based Variants} (Section~\ref{moe_variants}) combine \textsc{LoRA} with MoE mechanisms to enable adaptive parameter activation, as exemplified by \textsc{Mixture-of-LoRAs}~\cite{feng2024mixture-of-lora}, which distributes low-rank updates across multiple conditionally activated experts. Despite the rapid development, critical gaps remain in the field.

\textbf{First}, existing taxonomies in the general field of PEFT or \textsc{LoRA}
outline broad and superficial organization, 
and thus fail to render a fine-grained and systematic framework focused on \textsc{LoRA} variants based on their principal operational axes. \textbf{Second,} there is a lack of an in-depth review. Surveys on \textsc{LoRA} do not provide a thorough review of the theoretical foundations, design principles, and operational mechanisms that distinguish LoRA variants. 
This limitation, combined with the mathematical sophistication of many proposals, impedes accessibility, especially for non-specialists. \textbf{Third,} code support is fragmented and unwieldy. While the popular PEFT library~\cite{peft} provides a basic \textsc{LoRA} implementation with useful features (e.g., multi-LoRA serving), it supports only a limited set of variants. Worse, its codebase has become cluttered with 
deeply nested logic and tight interdependencies, making it difficult to read and extend. \textbf{Fourth,} evaluations are inconsistent and limited in scope. The original \textsc{LoRA} paper conducts the evaluation using models like RoBERTa~\cite{liu2019roberta} (GLUE~\cite{wang2018glue}), GPT-2~\cite{radford2019gpt2} (E2E NLG~\cite{novikova2017e2e}), and GPT-3~\cite{brown2020gpt3} (WikiSQL~\cite{zhong2017seq2sql}, MNLI~\cite{wang2018glue}, SAMSum~\cite{gliwa2019samsum}). Recent works now use large models such as LLaMA3~\cite{dubey2024llama3} and Qwen3~\cite{yang2025qwen3} for evaluation, creating a comparison gap. 
Moreover, evaluations remain largely confined to language tasks, despite the growing use of \textsc{LoRA} in various domains.

\definecolor{hidden-draw}{RGB}{177, 177, 177}
\definecolor{hidden-pink}{RGB}{255, 192, 203} 
\tikzset{
  my-box/.style={
    rectangle,
    draw=hidden-draw,
    rounded corners,
    text opacity=1,
    minimum height=1.5em,
    minimum width=5em,
    inner sep=2pt,
    align=center,
    fill opacity=.5,
    line width=0.8pt,
  },
  leaf/.style={
    my-box, 
    minimum height=1.5em,
    fill=hidden-pink!80, 
    text=black, 
    align=left,
    font=\normalsize,
    inner xsep=2pt,
    inner ysep=4pt,
    line width=0.8pt,
  }
}
\begin{figure*}[t!]
    \centering
    \resizebox{\textwidth}{!}{ 
        \begin{forest}
            forked edges,
            for tree={
                grow=east,
                reversed=true,
                anchor=base west,
                parent anchor=east,
                child anchor=west,
                base=center,
                font=\large,
                rectangle,
                draw=hidden-draw,
                rounded corners,
                align=center, 
                text centered, 
                minimum width=8em,
                edge+={darkgray, line width=1pt},
                s sep=3pt,
                inner xsep=2pt,
                inner ysep=3pt,
                line width=0.8pt,
                text width=6em, 
                ver/.style={rotate=90, child anchor=north, parent anchor=south, anchor=center},
            },
            where level=1{text width=8em,font=\normalsize,align=center}{}, 
            where level=2{text width=16em,font=\normalsize,align=center}{},  
            where level=3{text width=16em,font=\normalsize,align=center}{},  
            where level=4{text width=16em,font=\normalsize,align=center}{},
            [
                \textbf{LoRA Variants}, ver, text width=25em
                [
                    \textbf{Rank Adjustment Based} \\ \textbf{LoRA Variants (\S \ref{rank_variants})} 
                    , fill=green!10, text width=17em
                    [
                        \textbf{Rank Expansion Methods} (\S \ref{Rank Compositional Methods}), fill=green!10, text width=20em
                        [
                            \scriptsize \textbf{ReLoRA~\cite{lialin2307relora}, PeriodicLoRA~\cite{meng2024periodiclora},
                            CoLA~\cite{xia2024chain},
                            XGBLoRA~\cite{zhang2024xgblora},
                            MeLoRA~\cite{ren2024melora},
                            LoHA~\cite{hyeon2021fedpara}}\\
                            \scriptsize \textbf{LoKr~\cite{yeh2023lokr},  
                            HiRA~\cite{huang2024hira},
                            MoRA~\cite{jiang2024mora}},fill=green!10, text width=34em
                        ]
                    ]
                    [
                        \textbf{Rank Sharing Methods} (\S \ref{Rank Sharing Methods}), fill=green!10, text width=20em
                        [
                            \scriptsize \textbf{ShareLoRA~\cite{song2024sharelora}, VeRA~\cite{kopiczko2023vera}, 
                            RaSA~\cite{he2025rasa},
                            RandLoRA~\cite{albert2025randlora},
                            DenseLoRA~\cite{mu2025denselora},
                            ProLoRA~\cite{wang2024prolora}} \\
                            \scriptsize \textbf{BSLoRA~\cite{zhoubslora},
                            TiedLoRA~\cite{renduchintala2023tied-lora},
                            VB-LoRA~\cite{li2024vblora},
                            E\(^2\)LoRA~\cite{li2025e2lora}},
                            fill=green!10, text width=34em
                        ]
                    ]
                    [
                        \textbf{Rank Budgeting Methods} (\S \ref{subsubsec:Rank Redistribution Methods}), fill=green!10, text width=20em
                        [
                            \scriptsize \textbf{AdaLoRA~\cite{zhang2023adalora}, SaLoRA~\cite{hu2023structure}, 
                            SoRA~\cite{ding2023sora},
                            AutoLoRA~\cite{zhang2024autolora},
                            IncreLoRA~\cite{zhang2023increlora},
                            ALoRA~\cite{liu2024alora},} \\
                            \scriptsize \textbf{
                            BiLoRA~\cite{qiang2024bilora},
                            GoRA~\cite{he2025gora},
                            RaLoRA(-Pro)~\cite{ye2025gradient},
                            EVA~\cite{paischer2024eva}},
                            fill=green!10, text width=34em
                        ]
                    ]
                ]
                [
                    \textbf{Optimization Process Adjustment} \\ 
                    \textbf{Based LoRA Variants (\S \ref{training_dynamics_variants})}, 
                    fill=pink!10, text width=17em
                    [
                        \textbf{Stability Enhancing Methods}(\S \ref{subsubsec: stability_enhancement}), fill=pink!10, text width=20em
                        [
                            \scriptsize \textbf{
                            rsLoRA~\cite{kalajdzievski2023rslora},
                            LoRA+~\cite{hayou2024lora+},
                            RPLoRA~\cite{zhang2024riemannian}},
                            fill=pink!10, text width=34em
                        ]
                    ]
                    [
                        \textbf{Alignment Enhancing Methods}(\S \ref{subsubsec: update_decomposition}), 
                        fill=pink!10, text width=20em
                        [
                            \scriptsize \textbf{
                            DoRA~\cite{liu2024dora},
                            DeLoRA~\cite{bini2025delora},
                            DuDE~\cite{han2025dude},
                            LoRA-Pro~\cite{wang2024lora-pro},
                            RPLoRA~\cite{zhang2024riemannian},
                            FLoRA (Hao et al.)~\cite{hao2024flora}} \\
                            \scriptsize \textbf{FLoRA (Si et al.)~\cite{si2024flora},
                            Aurora~\cite{dong2025aurora},
                            SineLoRA~\cite{ji2024SineLoRA},
                            LoRAN~\cite{li2024loran},
                            LoDA~\cite{liu2025loda}},
                            fill=pink!10, text width=34em
                        ]
                    ]
                ]
                [
                    \textbf{Initialization Adjustment Based} 
                    \\ 
                    \textbf{LoRA Variants (\S \ref{initialization_variants})},
                    fill=violet!10 , text width=17em
                    [
                        \textbf{Data-independent Init Methods} (\S \ref{subsubsec: random_init}), 
                        fill=violet!10, text width=20em
                        [
                            \scriptsize \textbf{
                            NZLoRA~\cite{li2025beyond-zero-initialization},
                            Hayou et al.~\cite{hayou2024impact-of-initialization}
                            PiSSA~\cite{meng2024pissa},
                            MiLoRA~\cite{wang2024milora}, 
                            OLoRA~\cite{wang2024prolora},
                            NoRA~\cite{lin2024nora}} \\
                            \scriptsize \textbf{
                            NLoRA~\cite{guo2025nlora},
                            SORSA~\cite{cao2024sorsa}}, fill=violet!10, text width=34em
                        ]
                    ]
                    [
                        \textbf{Gradient-driven Init Methods}(\S \ref{subsubsec:Gradient-based initialization}), fill=violet!10, text width=20em
                        [
                            \scriptsize \textbf{LoRA-GA~\cite{wang2024lora-ga},
                            LoRA-One~\cite{zhang2025loraone}, 
                            GORA~\cite{he2025gora},
                            LoRA-TSD~\cite{si2024lora-tsd}}, fill=violet!10, text width=34em
                        ]
                    ]
                    [
                        \textbf{Acitivation-aware Init Methods}(\S \ref{subsubsec:Activation-based initialization}), fill=violet!10, text width=20em
                        [
                            \scriptsize \textbf{CorDA~\cite{yang2024corda},
                            EVA~\cite{paischer2024eva}}, fill=violet!10, text width=34em
                        ]
                    ]
                ]
                [
                    \textbf{Mixture-of-Experts Intergration} \\ \textbf{Based LoRA Variants (\S \ref{moe_variants})}, fill=orange!5 , text width=17em
                    [
                        \textbf{Loss Modification Methods} (\S \ref{subsubsec: loss_modification_methods}), 
                        fill=orange!5, text width=20em
                        [
                            \scriptsize \textbf{
                            MoELoRA~\cite{luo2024moelora},
                            LoRAMoE~\cite{dou2023loramoe}}, fill=orange!5, text width=34em
                        ]
                    ]
                    [
                        \textbf{Router Modification Methods} (\S \ref{subsubsec: router_modification_methods}), 
                        fill=orange!5, text width=20em
                        [
                            \scriptsize \textbf{
                            MoA~\cite{feng2024mixture},
                            AdaMoLE~\cite{liu2024adamole}}, fill=orange!5, text width=34em
                        ]
                    ]
                    [
                        \textbf{Expert Modification Methods} (\S \ref{subsubsec: expert_modification_methods}), 
                        fill=orange!5, text width=20em
                        [
                            \scriptsize \textbf{
                            MoLA~\cite{gao2024higher},
                            GOAT~\cite{fan2025make},
                            Hydra-LoRA~\cite{tian2024hydralora},
                            MoSLoRA~\cite{wu2024mixture},
                            MulhtiLoRA~\cite{wang2023multilora},
                            Sira~\cite{zhu2023sira}
                            }  \\ \scriptsize \textbf{
                            MtLoRA~\cite{agiza2024mtlora},
                            Yang et al.~\cite{yang2024multi},
                            Llava-mole~\cite{chen2024llava}, 
                            Moka~\cite{wei2025moka}}, fill=orange!5, text width=34em
                        ]
                    ]
                ]
            ]
        \end{forest}
    }
    \vspace{-5mm}
    \caption{Hierarchical taxonomy of LoRA variants based on four core principle operational axes.}
    \vspace{-5mm}
    \label{fig:taxonomy}
\end{figure*}
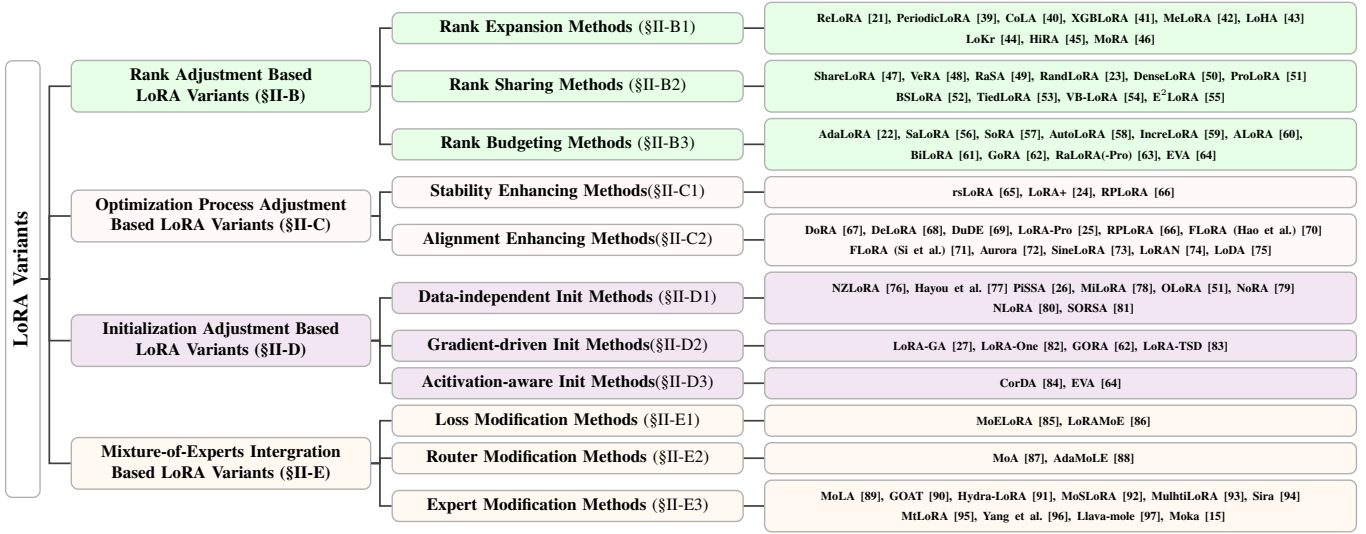

To address these challenges, this work presents the \textbf{first} unified study of \textsc{LoRA} variants: (1) We propose a structured and fine-grained taxonomy (Figure~\ref{fig:taxonomy}) focused on \textsc{LoRA} variants according to their operational principles; (2) Building upon the taxonomy, we conduct an in-depth review in Section~\ref{review_lora_and_variants} grounded in a unified theoretical framework; (3) Further, we provide a clean, modular codebase detailed in Section~\ref{lorafactory} that implements variants as subclasses of a \textsc{LoRA} base class, thereby significantly enhancing readability and extensibility; (4) Building on these infrastructures, we launch a large-scale empirical study across three domains: natural language generation, natural language understanding, and image classification, evaluating 20 representative variants that have been accepted in top AI/ML venues under extensive hyperparameter sweeping. We discover several important key findings as shown in Section~\ref{benchmark}, especially, \textsc{LoRA} can match or outperform most of its variants with appropriate hyperparameter configurations.
Our work provides a solid foundation for future research. The contributions are summarized as follows:

\begin{enumerate}
\item We formulate a structured taxonomy of \textsc{LoRA} variants, providing a \textbf{fine-grained systematic framework} based on the principal operational axes of \textsc{LoRA} variants. 

\item We present a theoretical review of \textsc{LoRA} variants, which establishes a \textbf{unified foundation} rooted in low-rank adaptation dynamics to promote understanding. 

\item We introduce \textbf{LoRAFactory}, which implements over 50 \textsc{LoRA} variants and functions beyond a toolkit by enabling standardized and extensible evaluations. 

\item We conduct \textbf{large-scale evaluations} with over 3,000 experiments across 3 model architectures and 22 tasks, spanning natural language generation, natural language understanding, and image classification.

\item We uncover several \textbf{key findings}, with two being particularly noteworthy: (1) \textsc{LoRA} and its variants are highly sensitive to the learning rate compared to other hyperparameters; (2) \textsc{LoRA} can match or outperform its most variants with proper hyperparameter configurations. 

\end{enumerate}
\vspace{-5mm}
\section{Review of \textsc{LoRA} and Its Variants}
In this section, we conduct a theoretical review; details of the notations we used can be found in the Appendix B. 
\vspace{-5mm}
\label{review_lora_and_variants}
\subsection{Overview of \textsc{LoRA}}
\label{overview_of_lora}
\subsubsection{Mechanism of \textsc{LoRA}}
\label{machanism_of_lora}
\textsc{LoRA} is grounded in the hypothesis that the updates to pretrained weights during fine-tuning possess \emph{low intrinsic ranks}, aligning with observations that over-parameterized models often reside on a low intrinsic dimension~\cite{li2018measuring-intrinsic}. Specifically, at each fine-tuning step \(t\), for a pretrained weight matrix \( \widetilde{W} \in \mathbb{R}^{m \times n} \), \textsc{LoRA} approximates the corresponding update \( \Delta W_t \) using low-rank matrices \( A_t \in \mathbb{R}^{m \times r} \) and \( B_t \in \mathbb{R}^{r \times n} \), with $r \ll \min(m, n)$. Formally:

\begin{equation}
W_t = \widetilde{W} + \Delta W_t = \widetilde{W} + \frac{\alpha}{r}A_tB_t.
\end{equation}

Here, the product \( A_tB_t \) is normalized by the rank \( r \) and scaled by a hyperparameter \( \alpha \). This design ensures that the magnitude of $\Delta W_t$ depends primarily on \( \alpha \) rather than the rank \( r \), allowing for more controllable fine-tuning. However, empirical studies~\cite{biderman2024lora-learn-less, shuttleworth2024lora-vs-fft} suggest setting \(\alpha \) to \(2r\), as a constant \( \alpha \) may lead LoRA to converge to low-rank solutions even under large-\(r\) settings.

\subsubsection{Comparison between \textsc{LoRA} and Full Fine-tuning}
\label{lora_vs_fft}
\textsc{LoRA} is fundamentally related to full fine-tuning, though still demonstrates differences in both optimization dynamics and final performance. Mathematically, the gradients of a low-rank adapter at the \(t\)-th step are expressed as:
\begin{equation}
\label{gradient}
\nabla A_t = \frac{\alpha}{r} \nabla \widetilde{W}_t B_t^\top , \quad \nabla B_t = \frac{\alpha}{r} A_t^\top  \nabla \widetilde{W}_t.
\end{equation}

As demonstrated in prior research~\cite{hao2024flora,he2025gora}, under standard \textsc{LoRA} initialization with either small learning rates or a frozen $A$ matrix (as in \textsc{LoRA-FA}~\cite{zhang2023lora-fa,zhu2024asymmetry-low-rank}), the update can be approximated, with exact simplification for frozen $A$ or approximate simplification under small learning rates as:
\begin{equation}
\label{gradient_compression}
\Delta W_t = A_tB_t = - \eta \frac{\alpha}{r} \sum_{i=0}^{t-1} A_0 A_0^\top \nabla \widetilde{W}_i.
\end{equation}

Moreover, the step-wise update obtained from the low-rank adapter can be expressed as:
\begin{equation}
\label{low-rank-gradient}
\begin{aligned}
    &\Delta W_{t+1} -\Delta W_t = (A_{t+1}-\eta\nabla A_t)(B_{t+1} - \eta\nabla B_t) - A_tB_t \\
    & = -\eta A_t\nabla B_t- \eta\nabla A_t B_t+\eta^2\nabla A_t\nabla B_t \\
    &\approx-\frac{\eta\alpha}{r}(A_tA_t^\top\nabla \widetilde{W}_t+\nabla \widetilde{W}_tB_t^\top B_t),
\end{aligned}
\end{equation}
where the approximation holds under the assumption of a small learning rate, such that the $\mathcal{O}(\eta^2)$ term is negligible.

Eqs.~\eqref{gradient_compression}-\eqref{low-rank-gradient} uncover the relationship between 
\textsc{LoRA} adapters and the gradients of pretrained weights. Especially, Eq.~\eqref{gradient_compression} reveals that a \textsc{LoRA} adapter essentially functions as a gradient compressor, which first compresses the gradient of the corresponding pretrained weight through $A^\top$, and then decompresses it via $A$.

Despite the connection, LoRA differs in its optimization dynamics, final performance, and applicable use cases. Ghosh et al.~\cite{ghosh2024closer-look-it} empirically show that during instruction tuning, models fine-tuned with \textsc{LoRA} retain closer alignment with the pretrained knowledge, whereas full fine-tuning tends to fit the instruction data closer. Specifically, \textsc{LoRA} results in a reduced token-level distribution shift compared to full fine-tuning. It learns localized adaptations, such as sentence initiation, leading to a more concentrated distribution shift.
As further validated by Biderman et al.~\cite{biderman2024lora-learn-less} and Shuttleworth et al.~\cite{shuttleworth2024lora-vs-fft}, LoRA better mitigates catastrophic forgetting~\cite{mccloskey1989catastrophic}. Additionally, Biderman et al.~\cite{biderman2024lora-learn-less}, and Schulman et al.~\cite{schulman2025lora-without-regret} find that \textsc{LoRA} is more sensitive to hyperparameters than full fine-tuning, especially to the learning rate.

\subsubsection{Advantages of \textsc{LoRA}}
\label{advantages_of_lora}
\textsc{LoRA}'s primary benefit lies in its ability to significantly reduce the memory footprint of optimizer states—particularly in mixed-precision training, where stateful optimizers~\cite{kingma2014adam,loshchilov2017adamw,shazeer2018adafactor} require storing states in 32-bit precision. Contrary to popular belief, \textsc{LoRA} introduces additional FLOPs in both training and inference (without merging).
This overhead is especially noticeable in single-GPU and non-offload setups. However, in distributed training settings, \textsc{LoRA} can reduce communication costs between devices and nodes, especially for optimizer offloading strategies such as \textsc{ZeRO-Offload}~\cite{ren2021zero-offload} and data parallelism strategies such as \textsc{ZeRO}~\cite{rajbhandari2020zero} and \textsc{FSDP}~\cite{zhao2023fsdp}, leading to faster overall training processes.

\begin{figure}[ht!]
\begin{center}
\includegraphics[width=\columnwidth]{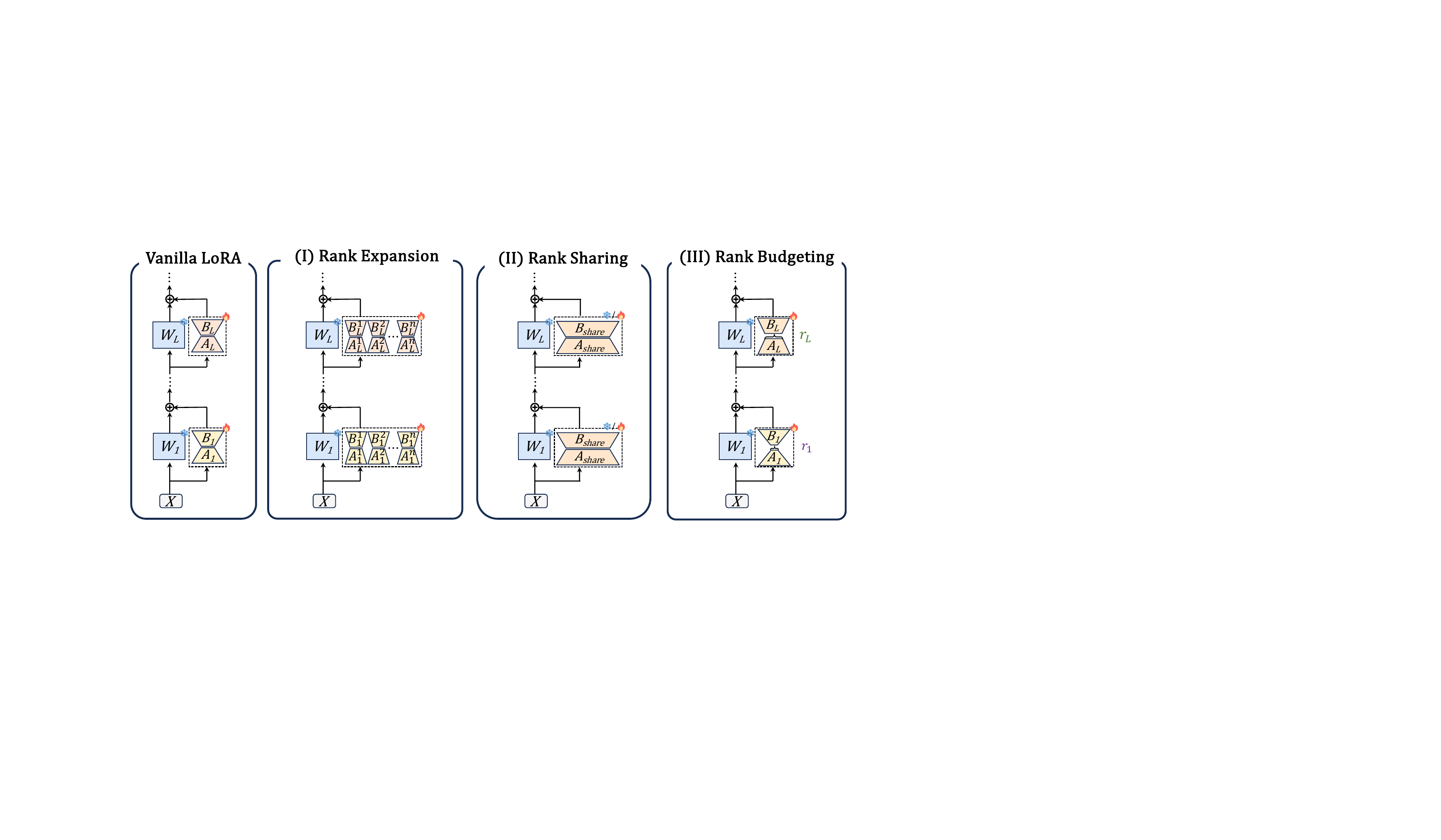}
\end{center}
\vspace{-3mm}
\caption{Illustration of rank adjustment based \textsc{LoRA} variants.}
\vspace{-5mm}
\label{fig: rank_methods}
\end{figure}
\subsection{Rank Adjustment Based \textsc{LoRA} Variants}
\label{rank_variants}

Vanilla \textsc{LoRA} applies a uniform small rank to all adapters for simplicity, though this design can hinder both expressiveness and parameter efficiency. Since different modules and layers contribute unevenly to downstream performance, a fixed small-rank allocation can be inherently suboptimal. 

As shown in Figure~\ref{fig: rank_methods}, recent research investigates three approaches: (a) Rank Expansion Methods, which composite low-rank matrices through linear algebraic principles; (b) Rank Sharing Methods, which share low-rank parameters across adapters to enable larger rank configurations; (c) Rank Budgeting Methods, which dynamically allocate rank across modules during or before training.


\subsubsection{Rank Expansion Methods}
\label{Rank Compositional Methods}
Rank expansion methods share a common objective: to preserve the parameter efficiency (not equal to computational efficiency) of \textsc{LoRA} while expanding the effective ranks. At their core, several well-known rank inequalities and identities from linear algebra provide theoretical justification for their effectiveness. These include:
\begin{gather}
\mathcal{R}(M_1+M_2)\leq\mathcal{R}(M_1)+\mathcal{R}(M_2), \label{eq:subadditivity} \\
\mathcal{R}(M_1\odot M_2) \le \mathcal{R}(M_1) \cdot \mathcal{R}(M_2), \label{eq:hadamard} \\
\mathcal{R}(M_1 \otimes M_2) = \mathcal{R}(M_1) \cdot \mathcal{R}(M_2), \label{eq:kronecker} \\
\mathcal{R}(M_1M_2) \le \min(\mathcal{R}(M_1), \mathcal{R}(M_2)), \\
\max(\mathcal{R}(M_1),\mathcal{R}(M_2)) \leq \mathcal{R}([M_1|M_2]) \\
\mathcal{R}([M_1|M_2]) \leq\mathcal{R}(M_1)+\mathcal{R}(M_2), \label{eq:concat_bound} \\
\mathcal{R}\left(\bigoplus_{i=0}^{k-1} M_i\right)=\sum_{i=1}^n\mathcal{R}(M_i), \label{eq:blkdiagnal}
\end{gather}
where $\mathcal{R}(M)$ denotes the rank of matrix $M$, $\odot$ the Hadamard product, $\otimes$ the Kronecker product, and \(\bigoplus\) represents block-diagonal concatenation. These guide the construction of composite structures, enabling richer representations without a proportional increase in trainable parameters.

Inspired by Eq.~\eqref{eq:subadditivity}, \textsc{\textbf{ReLoRA}}~\cite{lialin2307relora} introduces a \emph{merge-and-reinit} strategy to construct higher-rank updates by accumulating low-rank subspaces.

Training is divided into $N$ phases, each consisting of $T$ steps. At step \(t\), the current phase index is defined as $\tau = \lfloor (t-1)/T \rfloor + 1$. Let $t_\tau := \tau T$ denote the final step of phase $\tau$, with $t_0 := 0$.  At the end of phase $\tau$, the low-rank update is merged into the base weight:
\begin{equation}
    W_{t_\tau} = 
    \begin{cases}
        \widetilde{W}, & \tau = 0, \\
        W_{t_{\tau-1}} + \dfrac{\alpha}{r} A_{t_\tau} B_{t_\tau}, & \tau = 1, 2, \dots, N,
    \end{cases}
\end{equation}
Within phase $\tau$ (i.e., for $t_{\tau-1} < t \leq t_\tau$), the model weight is:
\begin{equation}
    W_t = W_{t_{\tau-1}} + \frac{\alpha}{r} A_t B_t.
\end{equation}
After merging, low-rank matrices are reinitialized, and their optimizer states are reset, enabling the exploration of a new low-rank subspace. To mitigate instability from optimizer resets, a jagged learning rate schedule, which re-warmups the learning rate from zero at the next phase, is adopted.

\textsc{\textbf{PeriodicLoRA}}~\cite{meng2024periodiclora} implements a method similar to \textsc{ReLoRA} but introduces a momentum-based merging mechanism to enhance training stability. Additionally, \textbf{\textsc{COLA}}~\cite{xia2024chain} also proposes a similar approach with a motivation from the Frank Wolfe algorithm~\cite{frank1956frank-wolfe-algorithm}. Moreover, the \emph{merge-and-reinit} method can also be viewed as a gradient boosting (GB) method. From this perspective, Zhang et al.~\cite{zhang2024xgblora} draw inspiration from GB algorithms such as GBDT~\cite{chen2016xgboost}, proposing \textsc{\textbf{XGBLoRA}}, which randomly selects 2 layers to be trained by rank-one adapters at each \emph{merge-and-reinit} phase.

However, the method of \textsc{ReLoRA} lacks a lower bound of effective rank. Furthermore, as the \emph{merge-and-reinit} process directly modifies the pretrained weights, these weights must be saved after training, leading to substantial storage requirements compared to \textsc{LoRA}. To address these issues, \textsc{\textbf{MELoRA}}~\cite{ren2024melora} draws inspiration from Eq.~\eqref{eq:blkdiagnal}. Specifically, it partitions both $A_t$ and $B_t$ into $k$ mini blocks and stacks them along the diagonal to form the overall update:
\begin{gather}
\label{eq: melora}
    \Delta W_t = \left( \bigoplus_{i=1}^{k} A_t^i \right)
                 \left( \bigoplus_{i=1}^{k} B_t^i \right), \\
    A_t^i \in \mathbb{R}^{\frac{m}{k} \times r'}, \quad
    B_t^i \in \mathbb{R}^{r' \times \frac{n}{k}}.
\end{gather}
This construction ensures that the effective rank is equal to $kr^\prime$. When $r^{\prime}=r$, \textsc{MELoRA} increases the rank from $r$ to $kr$ without increasing the trainable parameter count. Conversely, when $r^{\prime}=\frac{r}{k}$, \textsc{MELoRA} reduces the trainable parameter count by a factor of $k$, while preserving the rank of \(r\).

Hyeon-Woo et al.~\cite{hyeon2021fedpara} draw inspiration from the Hadamard product to enhance the expressiveness of \textsc{LoRA}, proposing \textsc{FedPara} (also known as \textsc{\textbf{LoHa}}~\cite{yeh2023lokr}). Specifically, a \textsc{LoHa} adapter reparameterizes the update to a pretrained weight matrix via the Hadamard product of two low-rank matrix pairs: $A_t^1 \in \mathbb{R}^{m \times r}$, $A_t^2 \in \mathbb{R}^{m \times r}$ and $B_t^1 \in \mathbb{R}^{r \times n}$, $B_t^2 \in \mathbb{R}^{r \times n}$. The adaptation update can be formally expressed as:
\begin{equation}
    \Delta W_t = \frac{\alpha}{r}(A_t^1B_t^1 \odot A_t^2B_t^2).
\end{equation}
As shown in Eq.~\eqref{eq:hadamard}, \textsc{LoHa} enables an upper effective rank bound of $r^2$, while only doubling the trainable parameter count compared to \textsc{LoRA} with the same hyperparameter $r$.

To approximate even higher-rank updates, \textsc{\textbf{HiRA}}~\cite{huang2024hira} constructs the update to a pretrained weight via the Hadamard product between the pretrained weight and the low-rank update. This update is formally expressed as:
\begin{equation}
    \Delta W_t = \widetilde W \odot \frac{\alpha}{r}A_tB_t.
\end{equation}
By leveraging the multiplicative interaction, \textsc{HiRA} allows potential high-rank update bounding by the product of the rank of the pretrained weight and the rank of the low-rank update.

Inspired by \textsc{LoHa} and \textsc{\textbf{KronA}}~\cite{edalati2025krona}, which employ Kronecker products for matrix decomposition, Yeh et al.~\cite{yeh2023lokr} propose \textsc{\textbf{LoKr}}, which can be formally expressed as:
\begin{gather}
    m_d = \max(u \le\min(k, \sqrt{m})|m\text{ mod }u=0), \\
    n_d = \max(u \le\min(k, \sqrt{n})|n\text{ mod }u=0), \\
    A_t \in \mathbb{R}^{m_d\times r}, \quad B_t \in \mathbb{R}^{r\times n_d}, \quad C_t \in \mathbb{R}^{m/m_d \times n/n_d}, \\
    \Delta W_t =\frac{\alpha}{r}A_tB_t \otimes C_t,
\end{gather}
where $k$ is a hyperparameter. \textsc{LoKr} maintains comparable parameter counts to \textsc{LoRA} while significantly increasing effective ranks.

\subsubsection{Rank Sharing Methods}
\label{Rank Sharing Methods}
Parameter sharing is a widely adopted strategy for neural networks~\cite{dehghani2018universal-transformer, takase2021lessons-sharing}. Recently, it has also become prevalent for improving the parameter efficiency of \textsc{LoRA}, as it allows for sharing low-rank weights across modules, thereby reducing the number of trainable parameters~\cite{song2024sharelora,kopiczko2023vera,li2024vblora, li2025e2lora}. For example, \textsc{\textbf{VB-LoRA}}~\cite{li2024vblora} implements an extreme parameter efficiency method using a shared vector bank strategy to composite low-rank matrices. In this paper, we focus on another function of parameter sharing, i.e., increasing the overall rank of adapters by sharing, and we refer to this as \textit{rank sharing strategies}.

An intuitive parameter sharing strategy is to share the trainable low-rank matrices across all modules. Following this idea, \textsc{\textbf{ShareLoRA}}~\cite{song2024sharelora} investigates the performance of sharing different components of the low-rank matrices, namely, matrix $A$, matrix $B$, or both, across modules. Empirical results show that sharing both $A$ and $B$ significantly reduces the trainable parameter count but incurs a noticeable performance drop. In contrast, sharing matrix $A$ achieves performance on par with vanilla \textsc{LoRA} while halving the trainable parameter count. 

This observation suggests a practical strategy: by sharing matrix $A$ and doubling the rank $r$, one can \textbf{potentially} surpass the performance of standard \textsc{LoRA} with the same trainable parameter count. Formally, when both low-rank matrices $A$ and $B$ are shared across targeted modules,  the update of a \textsc{ShareLoRA} adapter is defined as:
\begin{gather}
    A_t^S \in \mathbb{R}^{m_{\text{max}} \times r}, \quad B_t^S \in \mathbb{R}^{r \times n_{\text{max}}}, \\
    \label{slice operation}
    \Delta W_t = \frac{\alpha}{r} A_t^S[:m,:]B_t^S[:,:n], \quad \Delta W_t \in \mathbb{R}^{m \times n},
\end{gather}
where $A_t^S,B_t^S$ are shared low-rank matrices at step \(t\), $m_{\text{max}}$ and $n_{\text{max}}$ denote the maximum input and output dimensions across all adapted modules.

Building upon shared low-rank matrices, \textsc{\textbf{VeRA}}~\cite{kopiczko2023vera} proposes a vector-based fine-tuning approach. It shares and fixes randomly initialized low-rank matrices across modules, while introducing trainable scaling vectors to modulate the adaptation. This design is motivated by findings that tuning small, strategically chosen parts of randomly initialized models can yield surprisingly strong performance~\cite{lu2022frozen-transformer, schrimpf2021neural-arch, frankle2020training-batchnorm}. Formally, the adaptation in \textsc{VeRA} is expressed as:
\begin{gather}
    A^S \in \mathbb{R}^{m_{\text{max}} \times r}, \quad B^S \in \mathbb{R}^{r \times n_{\text{max}}}, \\
    \Lambda_t^d \in \mathbb{R}^{r \times r}, \quad \Lambda_t^b \in \mathbb{R}^{n \times n}, \\
    \Delta W_t = \frac{\alpha}{r} A^S[:m,:]\Lambda_t^d B^S[:,:n] \Lambda_t^b, \quad \Delta W_t \in \mathbb{R}^{m \times n},
\end{gather}
where $\Lambda_t^d$ and $\Lambda_t^b$ are diagonal matrices constructed at the $t$-th step from the trainable vectors $d_t \in \mathbb{R}^r$ and $b_t \in \mathbb{R}^n$. 

\textsc{\textbf{Tied-LoRA}}~\cite{renduchintala2023tied-lora} further investigates the performance of freezing different parts of shared low-rank matrices and scaling vectors upon the architecture of \textsc{VeRA}. This vector-based formulation shifts the optimization focus to the scaling vectors, enabling a substantial increase in rank while maintaining a trainable parameter count smaller than that of vanilla \textsc{LoRA}.

\textsc{\textbf{RaSA}}~\cite{he2025rasa} decomposes low-rank adaptation matrices into shared and module-specific (local) components. For a low-rank adapter of rank $r$, \textsc{RaSA} allocates $k$ ranks to be shared across all modules of the same type (e.g., query projection modules). Since these shared components have consistent shapes, no slicing operations are required during computation. The remaining $r - k$ ranks are kept specific to each module. In a model with $L$ layers, the effective rank of each \textsc{RaSA} adapter becomes $(r - k) + L \times k$, while the total number of trainable parameters remains identical to a \textsc{LoRA} adapter of rank $r$. Formally, the update $\Delta W_t$ computed by a \textsc{RaSA} adapter is given by:
\begin{equation}
\Delta W_t = 
\begin{bmatrix}
B_t^L & B_t^S
\end{bmatrix}
D_t^L 
\begin{bmatrix}
A_t^L \\
A_t^S
\end{bmatrix},
\end{equation}
where $A_t^L$ and $B_t^L$ are the local low-rank weights, and $D_t^L$ is a trainable diagonal scaling matrix. 

Similar to \textsc{RaSA}, \textsc{\textbf{BSLoRA}}~\cite{zhoubslora} decomposes low-rank adapters into three parts: \emph{inter-layer shared parts}, \emph{intra-layer shared parts}, and \emph{local parts}. This stems from an entropy-based analysis~\cite{lin2024mlp-can-be} on fine-tuned low-rank adapters, revealing high similarity of adapters within and between adjacent layers, indicating redundancy and sharing potential. Formally, the update $\Delta W_t$ of \textsc{BSLoRA} is:
\begin{equation}
\begin{split}
\Delta W_t = 2 \times (A_t^LB_t^L + \mathcal{T}(A_t^{S_1}B_t^{S_1}) + \mathcal{T}(A_t^{S_2}B_t^{S_2})),
\end{split}
\end{equation}
where $\mathcal{T}(\cdot)$ enables shape-flexible parameter sharing, and $2$ is a fixed scaling factor (adopted in the \href{https://github.com/yuhua-zhou/BSLoRA/blob/bf2f69c295e183578706d35a19c849ef8623e10b/peft/tuners/share_lora/layer.py\#L137}{official implementation}). Here, $A_t^{S_1}, B_t^{S_1}$ are shared within a layer, and $A_t^{S_2}, B_t^{S_2}$ are shared across layers. As slicing (Eq.~\eqref{slice operation}) requires shared weights to be initialized at the maximum module dimension, \textsc{BSLoRA} introduces two compact transformations $\mathcal{T}(AB)$ as:
\begin{gather}
    \mathcal{T}_g(AB) = G_{io} G_{id} AB G_{od} G_{ou}, \quad AB \in \mathbb{R}^{k \times d}, \\
    \mathcal{T}_k(AB) = (K_A \otimes A)(K_B \otimes B),
\end{gather}
where $G_{id} \in \mathbb{R}^{1\times k}$, $G_{od} \in \mathbb{R}^{d\times 1}$, $G_{iu} \in \mathbb{R}^{m\times 1}$, $G_{ou} \in \mathbb{R}^{1\times n}$ are gating matrices, and $K_A \in \mathbb{R}^{m/k \times 1}$, $K_B \in \mathbb{R}^{1 \times n/d}$ are Kronecker kernels. These allow shared weights of an arbitrary size $k \times d$ to be efficiently transformed to target dimensions $m \times n$, enabling flexible, efficient, and adaptive sharing.

\textsc{\textbf{RandLoRA}}~\cite{albert2025randlora} performs full-rank weight updates by decomposing a weight update $\Delta W_t\in \mathbb{R}^{m\times n}$ into a sum of products involving shared, fixed, randomly initialized low-rank bases and trainable scaling coefficients:
\begin{gather}
    n_b= \left\lceil \min(d_{s}, U) /r \right\rceil, \\
    \Delta W_t = \frac{2}{\sqrt{n_b}} \sum_{i=1}^{n_b} \Gamma_t^i A_t^S[:m,:] \Lambda_t^i B_t^{S_i}[:,:n],
\end{gather}
where \( d_{s} \) denotes the smaller dimension of the module with the largest output dimension among all target modules. (adopted in the \href{https://github.com/huggingface/peft/blob/337be05f03fd5c631154ba58afcc95c2c86529d8/src/peft/tuners/randlora/model.py\#L198}{PEFT implementation}), \(U\) is an additional hyperparameter introduced in LoRAFactory to balance computational efficiency of \textsc{RandLoRA}, and $\frac{2}{\sqrt{n_b}}$ is a scaling factor.   
Here, \( A_t^S \in \mathbb{R}^{\min(d,k) \times r} \) and \( B_t^{S_i} \in \mathbb{R}^{r \times \max(d,k)} \) are shared random basis matrices ($A_t^S$ is further shared across random bases), and \( \Gamma_t^i \in \mathbb{R}^{m\times m} \), \( \Lambda_t^i \in \mathbb{R}^{r\times r} \) are module-specific trainable scaling coefficients. 
For compatibility between adapted layers with distinct input and output dimensions and the fixed random bases, \textsc{RandLoRA} swaps \( A_t^S \) and \( B_t^{S_i} \) with \( {B_t^{S_i}}^\top \) and \( {A_t^S}^\top \) when the largest dimension of an adapted module is not its output dimension.

\textsc{\textbf{DenseLoRA}}~\cite{mu2025denselora} proposes a strategy that additionally refines the low-rank hidden states rather than only fine-tuning the low-rank weights. Specifically, \textsc{DenseLoRA} shares low-rank weights $A_t^S$ and $B_t^S$ across modules of the same type and introduces an intermediate module-specific
trainable matrix $C_t \in \mathbb{R}^{r \times r}$ to refine the hidden states. Formally, the adaptation form of \textsc{DenseLoRA} can be expressed as:
\begin{equation}
    \Delta W_t = \frac{\alpha}{r}A_t^SC_tB_t^S.
\end{equation}
By sharing low-rank matrices, DenseLoRA sharply reduces the trainable parameters, enabling high-rank configurations with smaller or comparable parameter counts compared with LoRA.

The aforementioned sharing-based methods share ranks across modules. In contrast, \textsc{\textbf{ProLoRA}}~\cite{wang2024prolora} shares ranks intra low-rank matrices. The computation of a \textsc{ProLoRA} adapter can be expressed as:
\begin{gather}
\begin{aligned}
    \Delta W_t &= A_tB_t = \\\bigl[ A_t^L \mid A_t^{S_1} \mid \cdots \mid A_t^{S_{P-1}} \bigr] 
             & \bigl[ B_t^L \mid B_t^{S_1} \mid \cdots \mid B_{S_t^{P-1}} \bigr]^\top,
\end{aligned}
\end{gather}
where $A_t^L, B_t^L$ are local (rank-specific) low-rank matrices, and $A_t^{S_i}, B_t^{S_i}$ are components obtained by applying a row-wise cyclic shift to a shared base matrix: $A_t^{S_i} = \text{Roll}(A_t^{S_0}, i\cdot\delta_A)$, $B_t^{S_i} = \text{Roll}(B_t^{S_0}, i\cdot\delta_B)$. Here, $\delta_A$ and $\delta_B$ are the \textbf{strides} that control the shift offset. This share and shift strategy allows parameters to be reused between ranks, enabling a larger effective rank.

\subsubsection{Rank Budgeting Methods}
\label{subsubsec:Rank Redistribution Methods}
As we mentioned before, \textsc{LoRA} neglects different modules contribute unequally to task-specific adaptation~\cite{zhang2023adalora, mao2024dora, ding2023sparse, rajabzadeh2024qdylora}. Allocating overmuch ranks to less critical modules may waste parameter budgets and potentially lead to overfitting, while assigning insufficient ranks to pivotal modules could constrain their ability to learn task-specific information. Consequently, the core challenge of such methods lies in intelligently and adaptively allocating the ranks of low-rank adapters with a predefined budget.

To facilitate masking ranks to a budget, \textsc{\textbf{AdaLoRA}}~\cite{zhang2023adalora} parameterizes a low-rank adapter as $\Delta W_t = A_t D_t B_t^\top$, factorized form analogous to truncated SVD, where $A_t \in \mathbb{R}^{m \times r}$ and $B_t \in \mathbb{R}^{n \times r}$ are matrices containing vectors simulating singular vectors, and $D_t \in \mathbb{R}^{r \times r}$ is a diagonal matrix containing values simulating singular values. To simulate the orthogonality of SVD during training, an auxiliary regularization term $\mathcal{L}_{\text{reg}}$ is added to the training loss with a hyperparameter \(\lambda\):
\begin{gather}
    \label{eq: orth}
    R_{\text{orth}}(A,B) = \|A^\top A - I\|_{\mathsf{F}}^2 + \|B^\top B - I\|_{\mathsf{F}}^2, \\
    \mathcal{L}_{\text{reg}} = \lambda \cdot \sum_{i=1}^{k}R_{\text{orth}}(A_t^i, B_t^i)
    \label{eq:regularization}
\end{gather}
where $k$ is the number of \textsc{AdaLoRA} adapters in the model.

\textsc{AdaLoRA} incorporates an importance scoring mechanism to mask less critical ranks. At each training step \(t\), the \(i\)-th diagonal value of \(D_t\) is masked to zero or retained after each backpropagation update, according to the importance score \(S_t^i\) of the \(i\)-th triplet of the adapter, comprising the \(i\)-th columns \(a_t^i \in \mathbb{R}^m\) and \(b_t^i \in \mathbb{R}^n\) of \(A_t\) and \(B_t\), and the \(i\)-th diagonal value \(d_t^i\) of \(D_t\), as follows:
\begin{equation}
    d_t^i \gets m_t^i \cdot d_t^i, \quad \text{where} \quad
    m_t^i =
    \begin{cases}
        1 & \text{if } S_t^i \geq \theta_t, \\
        0 & \text{otherwise}.
    \end{cases}
\end{equation}
Here, the threshold $\theta_t$ is set to the $b_t$-th largest value of importance scores of all triplets in the model, such that exactly $b_t$ singular values remain masked. The budget $b_t$, which controls the number of active singular values at each step $t$, follows a piecewise schedule across \(T\) total steps:
\begin{equation}
    b_t = 
    \begin{cases}
        b_0, & 0 \leq t < t_i, \\
        b_t^{\text{anneal}}, & t_i \leq t < T - t_f, \\
        b_T, & \text{otherwise},
    \end{cases}
    \label{eq:budget_scheduler}
\end{equation}
where $b_0$ and $b_T$ are the initial and final budgets, respectively. During the annealing phase, $b_t^{\text{anneal}}$ decreases cubically from $b_0$ to $b_T$ over the interval $[t_i, T - t_f)$, following:
\begin{equation}
    b_t^{\text{anneal}} = b_T + \left(b_0 - b_T\right) \left(1 - \frac{t - t_i}{T - t_i - t_f}\right)^3.
    \label{eq:annealing}
\end{equation}
One of the metrics for accurately estimating importance scores is a sensitivity measurement, defined below to capture the influence of parameter $w$ on the loss across update steps:
\begin{gather}
    I(w) = |w\cdot g|, \\
    \bar{I}_t(w) = \beta_1 \bar{I}_{t-1}(w) + (1-\beta_1)I_t(w), \\
    \bar{U}_t = \beta_2 \bar{U}_{t-1}(w) + (1 - \beta_2)|I(w) - \bar{I}_{t-1}(w)|, \\
    s_t(w) = \bar{I}_t(w) \cdot \bar{U}_t(w) ,
\end{gather}
where \(g\) is the gradient of \(w\), \(\beta_1\) and \(\beta_2\) are hyperparameters that are smaller than 1. The importance score of \textsc{AdaLoRA} is therefore defined as:
\begin{equation}
    S_t^i = s_t(d_t^i) + \frac{1}{m}\sum_{j=1}^m s_t(a_t^{ij}) + \frac{1}{n}\sum_{j=1}^n s_t(b_t^{ij})
\end{equation}
These designs enable \textsc{AdaLoRA} to adaptively allocate representational capacity during training, pruning less informative directions while preserving those critical for performance.

However, due to the masking mechanism, \textsc{AdaLoRA} initializes low-rank adapters with a uniform initial rank slightly larger than the final average rank (e.g., 1.5 times), leading to parameter redundancy, and the maximum rank of each adapter is bounded by the initial rank, limiting the model’s capacity to expand its representational budget. To address this issue, \textsc{\textbf{IncreLoRA}}~\cite{zhang2023increlora} adopts an incremental rank allocation strategy.
It first views the parameterization \(A_tD_tB_t^\top\) with the sum of the product of rank-one components \(a\), \(b\), and \(d\):
\begin{equation}
    \Delta W_t= A_tD_tB_t^\top = \sum_{i=1}^rd_t^i \cdot {a_t^i}  {b_t^i}^\top.
\end{equation}
During fine-tuning, additional rank-one components with randomly initialized \(a, b\) (\(d\) will be initialized with a small value) are allocated every $t_n$ steps to the top-$h$ most important modules at that interval. Both $h$ and $t_n$ are hyperparameters. A separate learning rate warmup and decay schedule is applied for newly added rank-one components. As a result, the ranks of modules with high importance are incrementally increased at every $t_n$ steps until the total rank budget is exhausted. The importance score used by \textsc{IncreLoRA} adopts the same smoothing strategy as \textsc{AdaLoRA}; the raw (unsmoothed) module-wise importance is computed by averaging all sensitivity scores in the corresponding update matrix. The orthogonality regularization (Eq.~\ref{eq:regularization}) is also applied by \textsc{IncreLoRA}.

\textsc{SaLoRA}~\cite{hu2023structure} extends \textsc{AdaLoRA} with a distinct masking strategy that formulates rank budgeting as an optimizable objective via $L_0$ regularization on simulated singular values. This is achieved through two techniques: (1) a differentiable surrogate $R_{L_0}$ approximating the non-differentiable $L_0$ norm, and (2) Lagrangian relaxation to embed the rank budget constraint $b$ into the loss for automatic budget control.

\textsc{SaLoRA} maps the diagonal entries of matrix $D$ to $[0,1]$ using a Hard-Concrete (HC) distribution with $u \sim \mathcal{U}(0,1)$:
\begin{gather}
    \tilde{d}_t^i = \sigma\!\left(\frac{\log\!\left(\frac{u}{1-u}\right) + \log(d_t^i)}{\tau}\right) \cdot (\zeta - \gamma) + \gamma, \\
    \Delta W_t = \sum_{i=1}^r \min\!\bigl(1, \max(0, \tilde{d}_t^i)\bigr) \cdot a_t^i {b_t^i}^\top,
\end{gather}
where $\sigma$ is the sigmoid function, $\tau$ is its temperature, and $\gamma < 0$, $\zeta > 1$ are HC hyperparameters that push most values outside $[0,1]$ toward $(-\infty, 0)$ or $(1, \infty)$. The surrogate $R_{L_0}$ has a closed-form expression based on the HC distribution:
\begin{equation}
    R_{L_0}(D) = \mathbb{P}(\tilde{d}_t^i > 0) = \sum_{i=1}^r \sigma\!\left(\log(d_i) - \tau \log\!\left(\frac{-\gamma}{\zeta}\right)\right).
\end{equation}

Given a target rank budget $b$, \textsc{\textbf{SaLoRA}} combines $R_{L_0}$ with orthogonality regularization $R_{\text{orth}}$ in Eq.~\ref{eq: orth} via Lagrangian relaxation, yielding the following regulation loss with hyperparameters $\lambda$ and $\beta$:
\begin{equation}
    \mathcal{L}_{\text{reg}} = \lambda \cdot \sum_{i=1}^k R_{\text{orth}}(A_t^i, B_t^i) + \beta \cdot \left( \frac{1}{r} \sum_{i=1}^k R_{L_0}(D_t^i) - b \right)^2.
\end{equation}

\textsc{\textbf{ALoRA}}~\cite{liu2024alora} introduces a rank reallocation strategy based on a train-evaluate-reallocate-retrain loop. At its core lies Ablation-based LoRA (\textsc{AB-LoRA}), an importance estimation method designed to guide rank reallocation. For a given rank component $r_i$, its importance score is computed as:
\begin{equation}
    \mathrm{IS}(r_i)=S(M)-S(M_{\setminus r_i})+S(M_{r_i}),
\end{equation}
where $S(M)$ denotes the performance of the fully fine-tuned model, $S(M_{\setminus r_i})$ is the performance after removing the component with rank $r_i$, and $S(M_{r_i})$ is the performance of the model when only rank $r_i$ is retained. 
Based on these scores, the least important ranks are removed from their respective modules and reallocated to more critical ones. The model is then further fine-tuned to adapt to the updated rank configuration.

The core idea of \textsc{\textbf{GoRA}}~\cite{he2025gora} is to perform a one-off gradient computation on a small subset of training data before training, jointly achieving adaptive rank allocation and gradient-driven weight initialization (detailed in Section~\ref{subsubsec:Gradient-based initialization}). By initializing low-rank matrices with allocated ranks, \textsc{GoRA} adaptively assigns more parameters to modules that have a greater impact on final performance, avoiding parameter redundancy and maintaining training stability. Specifically, \textsc{GoRA} first measures the advantage of each weight on \(k\) pretrained weights to be adapted $\{\widetilde{W}_i\}_{i=0}^k$ with corresponding gradients $\{G_i\}_{i=0}^k$ based on loss sensitivity:
\begin{gather}
    I_i=\text{avg}(|\widetilde{W}_i\odot G_i|), \quad\alpha_i=\frac{I_i}{\sum_{i=1}^kI_i}.
\end{gather}
The rank for the \(i\)-th low-rank adapter is determined by:
\begin{gather}
    P_{\mathrm{total}} = \sum_{i=1}^k (m_i +n_i)\cdot r_{\text{ref}} \\
    r_i=\mathrm{round}(\frac{P_{\mathrm{total}}\cdot\alpha_i}{\sqrt{m_i+n_i}}),\quad\mathrm{s.t.}\quad r_{\mathrm{min}}\leq r_i\leq r_{\mathrm{max}},
\end{gather}
Where \(\mathrm{round(\cdot)}\) denotes rounding to the nearest integer, and \(r_{\text{ref}}\) is a reference rank to control the parameter budget, and $r_{\min}, r_{\max}$ are predefined bounds of rank allocation.

Building upon the framework of \textsc{GoRA}, \textsc{\textbf{RaLoRA(-Pro)}}~\cite{ye2025gradient} further introduces an entropy-based effective rank estimator to measure the intrinsic dimensionality of gradient matrices $\{G_i\}_{i=0}^k$:
\begin{equation}
\operatorname{erank}(G_i)=\exp\left(-\sum_{i=1}^n p_i \log p_i\right), p_i=\frac{\sigma_i}{\sum_{j=1}^n\sigma_j},
\end{equation}
where $p_i$ denotes the normalized singular value distribution. The core insight of \textsc{RaLoRA(-Pro)} is revealing the substantial gap between the fixed small rank of the low-rank adapter (typically 8) and the gradient’s intrinsic dimensionality (GID), which can be up to 300. Building on this observation, using the block-diagonal structure in Eq.~\eqref{eq: melora}, RaLoRA aligns each low-rank adapter’s effective rank to the corresponding GID by adaptively increasing the number of diagonal blocks without increasing the total parameter count. RaLoRA-Pro further incorporates the parameter allocation strategy of \textsc{GoRA}, achieving dual adaptive alignment at both intra-layer and inter-layer levels with a manual parameter budget.

\textsc{\textbf{EVA}}~\cite{paischer2024eva} performs incremental SVD on downstream activation vectors and selects the top-$r$ right singular vectors to initialize the low-rank matrix $A$, thereby capturing the highest-variance directions in activation space and theoretically maximizing the initial gradient signal (in this section, we focus solely on \textsc{EVA}’s adaptive rank allocation; initialization is discussed in Section~\ref{subsubsec:Activation-based initialization}). Specifically, \textsc{EVA} computes the explained variance ratio for each singular vector:
\begin{equation}
    \xi_j^i=\frac{\sigma_j^{i^2}}{(M-1)||\boldsymbol{\sigma}^i||_1},
\end{equation}
and globally redistributes the rank budget by prioritizing directions with the largest explained variance, effectively reducing redundancy while preserving the most informative subspaces.
\subsection{Optimization Process Adjustment Based \textsc{LoRA} Variants}
\label{training_dynamics_variants}

\begin{figure}[t]
\begin{center}
\includegraphics[width=\columnwidth]{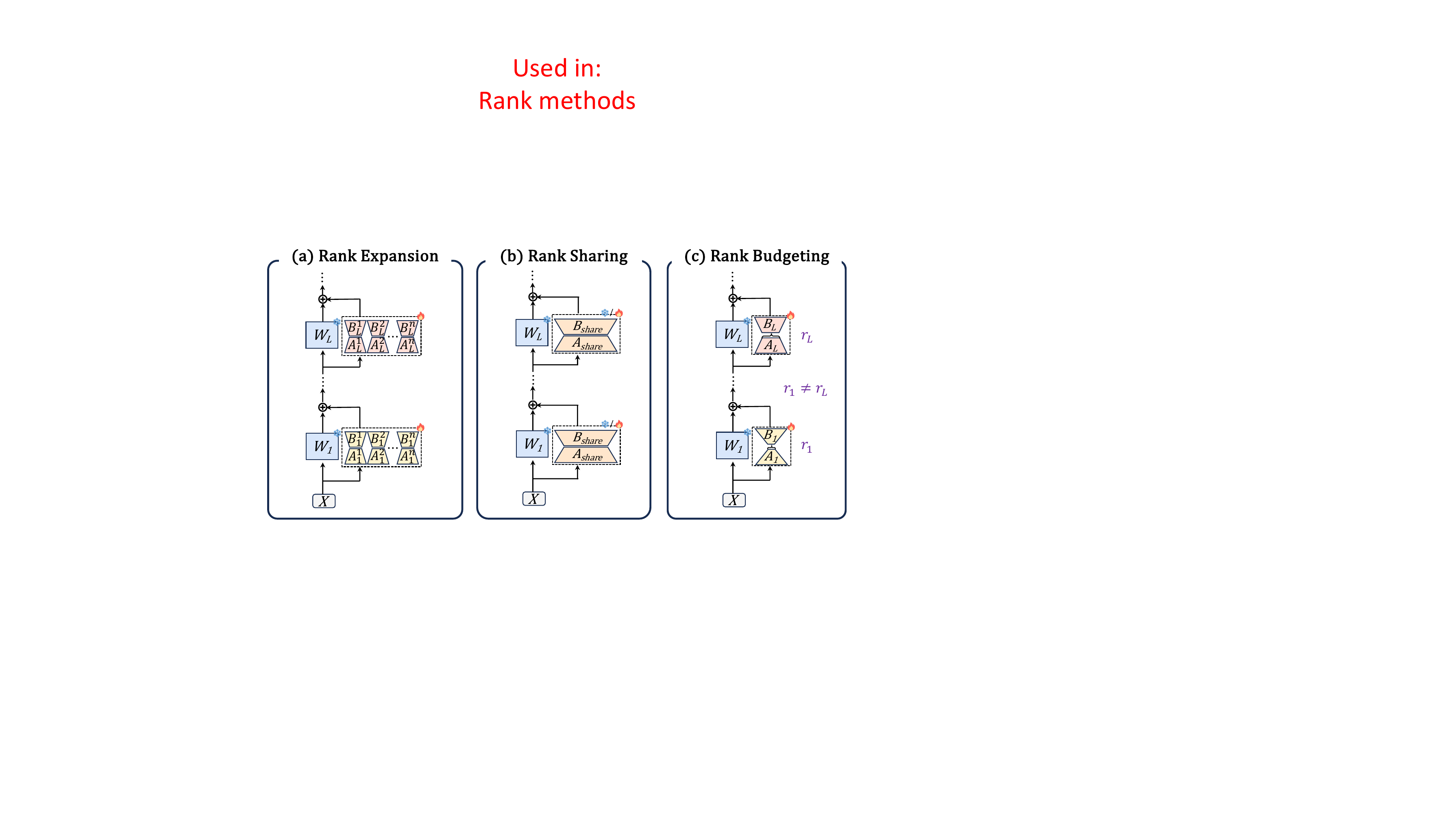}
\end{center}
\vspace{-3mm}
\caption{Illustration of optimization process adjustment based \textsc{LoRA} variants.}
\vspace{-5mm}
\label{fig: optimization_process}
\end{figure}

As shown in Figure~\ref{fig: optimization_process}, this section introduces \textsc{LoRA} variants that directly adjust the optimization process of \textsc{LoRA}, including:(a) Stability Enhancement Methods focus on regulating training dynamics to prevent collapse or instability. (b) Update Alignment Methods aim to bridge the gap between low-rank adaptation and full fine-tuning.

\subsubsection{Stability Enhancing Methods}
\label{subsubsec: stability_enhancement}
\textsc{\textbf{RsLoRA}}~\cite{kalajdzievski2023rslora} introduces an optimized scaling factor for \textsc{LoRA} to achieve a rank-stabilized optimization process, ensuring two key rank-stability properties: (1) \textbf{Forward Stability}: If the input $X \in \mathbb{R}^{bs \times m}$ to the low-rank adapter is i.i.d. with an m'th moment of $\Theta_r(1)$ per entry, then the m'th moment of the adapter’s outputs remains $\Theta_r(1)$ per entry. (2) \textbf{Backward Stability}: If the gradients of the loss with respect to the adapter outputs are $\Theta_r(1)$ per entry, then the gradients propagated back to the adapter’s inputs also maintain $\Theta_r(1)$ per entry.
Consider the scaling factor to be optimized $\gamma_r \in \mathbb{R}$ with $\gamma_r \xrightarrow{r \to \infty} 0$, which constrains the product of low-rank matrices as the rank $r$ increases. For any training step $t > 1$, the low-rank matrices evolve according to the gradient formula in Eq~\eqref{gradient} and the inductive derivation:
\begin{gather}
    \label{lora-induction-a}
    A_t = (I + \mathcal{O}_r(\gamma_r^2))A_0, \\
    \label{lora-induction-b}
    B_t = A_0^\top\left(-\eta\gamma_r \sum_{i=0}^{t-1}\nabla \widetilde{W}_i + \mathcal{O}_r(\gamma_r^2)\right), \\
    \label{lora-induction}
    \gamma_r A_tB_t = -\eta\gamma_r^2 \sum_{i=0}^{t-1}A_0 A_0^\top \nabla \widetilde{W}_i + A_0 A_0^\top\mathcal{O}_r(\gamma_r^3).
\end{gather}
Assuming the entries of $A_0$ are i.i.d. with mean $0$,  variance $\sigma_A$ ($\mathbb{E}_{A_0}[A_0A_0^\top] = r \sigma_A I$), the expectation of Eq.~\eqref{lora-induction} becomes:
\begin{equation}
    \mathbb{E}_{A_0}[\gamma_r A_tB_t] = -\gamma_r^2 r \sigma_A \eta \sum_{i=0}^{t-1} \nabla \widetilde{W}_i + \mathcal{O}_r(\gamma_r^3r),
\end{equation}
For an adapted linear model where $Y=X(\widetilde{W}+AB)$, the gradient $G$ and output $O$ of the adapter satisfy:
\begin{gather}
    G = -\gamma_r^2 r \sigma_A \eta \sum_{i=0}^{t-1} \nabla X_t Y_i^\top X_i + \mathcal{O}_r(\gamma_r^3r), \\
    \mathbb{E}_{X,A_0}[O] = -\gamma_r^2 r \sigma_A \eta \sum_{i=0}^{t-1} \nabla \mathbb{E}_X[X_tX_i^\top]Y_i+ \mathcal{O}_r(\gamma_r^3r),
\end{gather}
where $G \in \mathcal{O}_r(\gamma_r^2r)$ and $\mathbb{E}_{X,A_0}[O] \in \mathcal{O}_r(\gamma_r^2r)$. To maintain both forward and backward stability, we require $\mathcal{O}_r(\gamma_r^2r) = \mathcal{O}_r(1)$, implying $\gamma_r \in \mathcal{O}_r(1/\sqrt{r})$. Therefore, \textsc{RsLoRA} recommends setting the scaling factor from $\frac{\alpha}{r}$ to $\frac{\alpha}{\sqrt{r}}$. 


Hayou et al.~\cite{hayou2024lora+} analyze the optimization dynamics of models adapted via low-rank adapters in the limit as the model width \( m \to \infty \) increases and propose \textsc{\textbf{LoRA+}} which sets the learning rate of matrix \(B\) \(2^4\) times that of matrix \(A\). In the wide-network regime, one expects the change in model predictions at any training step \( t \) to remain stable—specifically, that the prediction increment \( \Delta f_t(x) = f_t(x) - f_{t-1}(x) \) scales as \( \Theta(1) \), meaning it neither vanishes nor diverges.

To investigate this behavior, Hayou et al. consider a simplified, analytically tractable model defined as \( f(x) = x^\top (\widetilde{w} + a b) \), where \( \widetilde{w} \in \mathbb{R}^m \) is a fixed pretrained weight vector, \( a \in \mathbb{R}^m \) and \( b \in \mathbb{R} \) are trainable rank-one components, and the input \( x \in \mathbb{R}^m \) satisfies \( \|x\| = \Theta(1) \). Following standard LoRA initialization, the variances of the initial parameters are \( \sigma_{a_0}^2 = \Theta(m^{-1}) \) and \( \sigma_{b_0}^2 = \Theta(1) \). In this setting, the prediction increment at step \( t \) is given by:
\begin{equation}
\label{lora+_update}
\begin{aligned}
    \Delta f_t(x)&=-\eta b_{t-1}^2 (f_{t-1}(x) - y) \|x\|_2^2 \\
    &\quad - \eta (a_{t-1}^\top x)^2 (f_{t-1}(x) - y) \\
    &\quad + \eta^2 (f_{t-1}(x) - y)^2 b_{t-1} (a_{t-1}^\top x) \|x\|_2^2,
\end{aligned}
\end{equation}
where \( y \) denotes the ground-truth label, the learning rate is \( \eta = \Theta(m^c) \) for some constant \( c \in \mathbb{R} \), and the loss function is \( \frac{1}{2}(f(x) - y)^2 \). For clarity, define the following terms:
\begin{equation}
\label{lora+_delta}
\begin{aligned}
    \delta_t^1 &= \eta b_{t-1}^2 (f_{t-1}(x) - y) \|x\|_2^2, \\
    \delta_t^2 &= \eta (a_{t-1}^\top x)^2 (f_{t-1}(x) - y), \\
    \delta_t^3 &= \eta^2 (f_{t-1}(x) - y)^2 b_{t-1} (a_{t-1}^\top x) \|x\|_2^2.
\end{aligned}
\end{equation}

The stable optimization dynamics requires \( \delta_t^1, \delta_t^2, \delta_t^3 \in \Theta(1) \), which further implies \( f_t(x) \in \Theta(1) \) throughout training. Notably, \( \delta_t^3 \in \Theta(1) \) is automatically satisfied if \( \delta_t^1, \delta_t^2 \in \Theta(1) \), since it is a higher-order term in \( \eta \).

Notation \( \gamma[\cdot] \) introduced such that \( \nu = \Theta(m^{\gamma[\nu]}) \) captures the asymptotic scaling of any quantity \( \nu \). The conditions for stable dynamics yield the following system of constraints:
\begin{equation}
\label{lora+_constraints}
    \begin{cases}
        c + 2\gamma[b_{t-1}] + 1 = 0 & \text{(for } \delta_t^1 = \Theta(1)\text{)}, \\
        c + 2\gamma[a_{t-1}^\top x] = 0 & \text{(for } \delta_t^2 = \Theta(1)\text{)}, \\
        \gamma[b_{t-1}] + \gamma[a_{t-1}^\top x] = 0 & \text{(for } f_t(x) = \Theta(1)\text{)}.
    \end{cases}
\end{equation}

Solving this system yields \( c = -\frac{1}{2} \), implying that the learning rate should scale as \( \eta \in \mathcal{O}(m^{-1/2}) \). However, due to the initialization \( \sigma_{b_0}^2 = \Theta(1) \) and \( a_0^\top x \in \mathcal{O}(1) \) (by the \emph{Central Limit Theorem}), one can inductively show that \( b_t \in \mathcal{O}(m^{-1/2}) \) and \( a_t^\top x \in \mathcal{O}(m^{-1/2}) \) for all \( t > 0 \), resulting in \( f_t(x) \in \mathcal{O}(m^{-1/2}) \). Consequently, the parameter updates for \( a_t \) and \( b_t \) are of order \( \mathcal{O}(m^{-1}) \) and \( \mathcal{O}(m^{-1/2}) \).

This analysis reveals that \( \delta_t^1 \) and \( \delta_t^2 \) cannot simultaneously be \( \Theta(1) \) under standard LoRA configurations with a shared learning rate. To resolve this, Hayou et al. propose decoupling the learning rates for \( a \) and \( b \), suggesting that \( \eta_b \in \mathcal{O}(1) \) (for \( b \)) and \( \eta_a \in \mathcal{O}(m^{-1}) \) (for \( a \)) can restore stable training dynamics.

Zhang et al.~\cite{zhang2024riemannian} investigate another solution for stable \(\Delta f_t(x)\) under \(m \to \infty\) increases, leveraging Riemannian preconditioning. Specifically, their approach—\textsc{Riemannian Preconditioned LoRA} (which we denote as \textsc{\textbf{RPLoRA}})—employs a Riemannian metric derived from a regularized Lagrangian framework. This metric is grounded in the geometric optimization principles for low-rank matrices with objectives and constraints introduced by Mishra and Sepulchre~\cite{mishra2016riemannian}. Following this, \textsc{RPLoRA} modifies the gradients of the low-rank adapter parameters according to the natural gradient flow on the manifold of fixed-rank matrices, effectively preconditioning the optimization dynamics to maintain stability:
\begin{equation}
    \nabla A^*_t = \nabla A_t (B_tB_t^\top)^{-1}, \quad \nabla B^*_t = (A^\top_t A_t)^{-1}\nabla B_t,
\end{equation}
where \(\nabla A^*\) and \(\nabla B^*\) are the modified gradients of \textsc{RPLoRA} for a low-rank adapter. Under the modified gradients, the prediction increment shown in Eq.~\eqref{lora+_update} can be rewritten as:
\begin{equation}
\label{rplora-update}
\begin{aligned}
    \Delta f_t(x) &= -\eta (f_{t-1}(x) - y) \|x\|_2^2 \\
    &\quad - \eta (a_{t-1}^\top x)^2 (f_{t-1}(x) - y)\left\|a_{t-1}\right\|^{-2}_2 \\
    &\quad + \eta^2 (f_{t-1}(x) - y)^2 b_{t-1}^{-1} \left\|a_{t-1}\right\|^{-2}_2(a_{t-1}^\top x) \|x\|_2^2.
\end{aligned}
\end{equation}
Similar to Eq.~\eqref{lora+_delta}, defining \(\delta_t^1,\delta_t^2,\delta_t^3\) with the three terms of Eq.~\eqref{rplora-update}, we can rewrite the constraints shown in Eq.~\eqref{lora+_constraints} as:
\begin{equation}
    \begin{cases}
        c+1 = 0 & \text{(for } \delta_t^1 = \Theta(1)\text{)}, \\
        c + 2\gamma[a_{t-1}^\top x] - \gamma[\left\|a_{t-1}\right\|_2^{2}] = 0 & \text{(for } \delta_t^2 = \Theta(1)\text{)}, \\
        \gamma[b_{t-1}] + \gamma[a_{t-1}^\top x] = 0 & \text{(for } f_t(x) = \Theta(1)\text{)},
    \end{cases}
\end{equation}
where we can drive \(c=-1\) and correspondingly \(\eta = m^{-1}\). Under \( \sigma_{b_0}^2 = \Theta(1) \) and \( a_0^\top x \in \mathcal{O}(1) \), one can recursively derive \(b_t, a_t^\top x, \delta_t^1, \delta_t^2,\delta_t^3 \in \mathcal{O}(1)\) for all \(t\). Hence, the stable training dynamics are achieved.

\subsubsection{Alignment Enhancing Methods}
\label{subsubsec: update_decomposition}
Liu et al.~\cite{liu2024dora} perform a weight decomposition analysis on the fine-tuning updates from both \textsc{LoRA} and full fine-tuning, revealing an interesting contrast: while \textsc{LoRA}'s updates demonstrate a positive correlation between magnitude and directional changes, full fine-tuning exhibits a slightly inverse relationship. This discrepancy motivates their proposed method, \textsc{\textbf{DoRA}}, which decouples magnitude and directional learning in \textsc{LoRA}, addressing the potential complexity of jointly optimizing both components and achieving an optimization pattern more closely aligning that of full fine-tuning. Formally, the adapted weight of \textsc{DoRA} can be expressed as:
\begin{equation}
\label{dora_weight}
    W_t = m_t\cdot\frac{\widetilde{W}+\gamma_rA_tB_t}{\left\|\widetilde{W}+ \gamma_rA_tB_t\right\|_F}, \quad \gamma_r = \frac{\alpha}{r},
\end{equation}
where $m_t$ is a learnable magnitude vector, initialized as the Frobenius norm on the input dimension of the pretrained weight. 

Similarly, \textsc{\textbf{DeLoRA}}~\cite{bini2025delora} introduces a strategy that decouples the directional and magnitude updates by combining \textsc{LoRA} with the idea of \textbf{ETHER}~\cite{bini2024ether}. Formally, the adaptation in \textsc{DeLoRA} can be expressed as:
\begin{equation}
    \Delta W_t = A_tD_tB_t = \frac{\lambda\left\|\widetilde{W}\right\|_2}{r} \sum_{i=1}^{r} \frac{a_t^i {b_t^i}^\top}{\|a_t^i\|_2 \|b_t^i\|_2},
\end{equation}
where $a_t^i$ and $b_t^i$ are the $i$-th rank-one components of the low-rank matrices, and $D_t$ is a diagonal matrix containing the scaling factors based on the norms of these components. Here, $\lambda$ is a trainable scalar that controls the upper bound on the norm of the update as:
\begin{equation}
    \left\|A_tD_tB_t\right\|_2 = \frac{\lambda\left\|\widetilde{W}\right\|_2}{r}\left\|\sum_{i=0}^{r}a_t^ib_t^i\right\|_2 \le  \lambda\left\|\widetilde{W}\right\|_2.
\end{equation}
The bounded adaptation prevents the adapted model from diverging from the pretrained model.

Hao et al.~\cite{hao2024flora} also utilize inductive derivation for $A_t, B_t$ to analyze the optimization dynamics of LoRA. Specifically, assume $\left\|\sum_{i=0}^{t}\nabla \widetilde{W}_i\right\|_F \leq L$ (constant $L$ is defined as an upper bound) for every training step $t$, which implies that the model stays within a finite Euclidean ball. In this case, by induction:
\begin{equation}
    A_t = (I + \eta\frac{\alpha}{r} f_A(t))A_0, \quad B_t = \eta\frac{\alpha}{r} A_0^\top f_B(t),
\end{equation}
where $f_A(t)$ and $f_B(t)$ are defined by induction as:
\begin{gather}
    f_A(t) = -\eta \frac{\alpha}{r}\sum_{i=0}^{t-1}\nabla \widetilde{W}f_B^\top(i), \\
    f_B(t) 
    = -\sum_{i=0}^{t-1}(I+ \eta\frac{\alpha}{r}f_A^\top(i))\nabla \widetilde{W}.
\end{gather}
The adapter's computation at step $t$ can be expressed as:
\begin{equation}
\label{flora-gammaab}
    \Delta W = \gamma_r A_tB_t = \eta\frac{\alpha^2}{r^2} A_0A_0^\top f_A(t)+ \eta^2\frac{\alpha^3}{r^3} f_B(t)A_0A_0^\top f_A(t).
\end{equation}
Hao et al. further establish the upper bound at every step:
\begin{equation}
\left\|f_A(t)\right\|_2 \leq \frac{\eta\gamma_rL^2(1-(\eta^2\gamma_r^2L^2)^t)}{1-\eta^2\gamma_r^2L^2},
\end{equation}
where $\gamma_r = \frac{\alpha}{r}$. This bound reveals that the second term in Eq.~\eqref{flora-gammaab} becomes negligible when $\eta\gamma_r \ll L$, since this condition ensures $\lim_{t \to \infty}\eta\gamma_r\left\|f_A(t)\right\| \ll 1$. Consequently:
\begin{equation}
\label{flora-gammaab-approx}
    \Delta W = \gamma_r A_tB_t \approx \gamma_rA_0\tilde{B_t} =: \eta\gamma_r^2 A_0A_0^\top \tilde{f}_B(t),
\end{equation}
where we can define $\tilde{f}_B(t) =:\tilde{f}_B(t-1) - \nabla \widetilde{W}_{t} = \sum_{i=0}^{t}\nabla \widetilde{W}_i$. Substituting this into Eq~\eqref{flora-gammaab-approx}, we can obtain the expression presented in Eq~\eqref{gradient_compression}, which implies that LoRA adapters function as gradient compressors under small learning rates.

Building upon this insight, Hao et al. propose \textsc{\textbf{FLoRA}}~\cite{hao2024flora}, which employs a random low-rank projection matrix $A^\top \in \mathbb{R}^{r \times m}$ to compress the gradients of a pretrained weight of size $m \times n$. The method efficiently computes and stores optimizer states using the compressed gradient, subsequently decompressing the optimizer's updates through $A$, enabling an intuitive update alignment. \textsc{GaLoRE}~\cite{zhao2024galore} proposes a similar framework where the projection matrix is obtained from the singular components of the gradient to retain gradient information as much as possible.

\textsc{\textbf{LoRA-Pro}}~\cite{wang2024lora-pro} enhances the alignment between \textsc{LoRA}'s optimization dynamics and full fine-tuning by explicitly minimizing the \textbf{step-wise discrepancy} between: the indirect updates to pretrained weights via \textsc{LoRA} (Eq.~\eqref{low-rank-gradient}) and the direct weight updates from full fine-tuning. \textsc{LoRA-Pro}'s objective can be viewed as an operational extension of \textsc{LoRA-GA}'s principle (Section~\ref{subsubsec:Gradient-based initialization}), generalizing the single-step gradient alignment to step-wise matching. \textsc{LoRA-Pro} incorporates \textsc{RsLoRA}'s scaling factor and optimizes:
\begin{equation}
\arg\min_{\nabla A_t^*,\nabla B_t^*} \left\| \frac{\alpha}{\sqrt{r}}(A_t\nabla B_t^* + \nabla A_t^*B_t) - \nabla \widetilde{W}_t \right\|_F^2,
\label{eq:lora_pro_objective}
\end{equation}
where $\nabla A^*_t,\nabla B_t^*$ are optimized gradients of $A_t,B_t$, $sA_t\nabla B_t^* + s\nabla A_tB_t^*$ represents the optimized indirect update. Denoting $\mathcal{D} = \left\| sA_t\nabla B_t + s\nabla A_tB_t - \nabla \widetilde{W}_t \right\|_F^2$, we derive the following optimality conditions:
\begin{gather}
\label{lora-pro-object-a}
\frac{\partial \mathcal{D}}{\partial \nabla A_t^*} = \frac{2\alpha}{\sqrt{r}}B_t^\top(sA_t\nabla B_t^* + s\nabla A_t^*B_t - \nabla \widetilde{W}_t) = 0, \\
\frac{\partial \mathcal{D}}{\partial \nabla B_t^*} = \frac{2\alpha}{\sqrt{r}}sA_t^\top (sA_t\nabla B_t^* + s\nabla A_t^*B_t - \nabla \widetilde{W}_t) = 0.
\end{gather}

Assuming $A_t,B_t$ maintain full rank during training such that $A_t^\top A_t$ and $B_t B_t^\top$ are invertible, we derive $\nabla B^*$:
\begin{equation}
\label{lora-pro-g-b-p}
\nabla B_t^* = \frac{\sqrt{r}}{\alpha}(A_t^\top A_t)^{-1}A_t^\top\nabla \widetilde{W}_t - (A_t^\top A_t)^{-1}A_t^\top\nabla A_t^*B_t.
\end{equation}
Substituting Eq.~\eqref{lora-pro-g-b-p} into Eq.~\eqref{lora-pro-object-a} and solving the resulting equation yields the expression for $\nabla A_t^*$:
\begin{equation}
\label{lora-pro-g-a}
\nabla A_t^* = \frac{\sqrt{r}}{\alpha}\nabla \widetilde{W}_tB_t^\top(B_tB_t^\top)^{-1} + A_tM_t,
\end{equation}
where $M_t \in \mathbb{R}^{r \times r}$ is an arbitrary matrix. Substituting Eq.~\eqref{lora-pro-g-a} into Eq.~\eqref{lora-pro-g-b-p} we have:
\begin{equation}
\label{lora-pro-g-b}
\nabla B_t^* = \frac{\sqrt{r}}{\alpha}(A_t^\top A_t)^{-1}A_t^\top\nabla \widetilde{W}_t[I - B_t^\top(B_tB_t^\top)^{-1}B_t] - B_tM_t,
\end{equation}
The final solutions to the objective of \textsc{LoRA-Pro} are shown in Eqs.~\eqref{lora-pro-g-a}-\eqref{lora-pro-g-b}. To utilize these solutions, LoRA-Pro alters the gradients of $A$ and $B$, effectively aligning the indirect updates from \textsc{LoRA} to the direct updates from full-finetuning (\textsc{LoRA-Pro} only utilizes the values and gradients of low-rank matrices to compute the aligned gradients since $\nabla \widetilde{W}_tB_t^\top=\nabla A_t$ and  $A_t^\top\nabla \widetilde{W}_t=\nabla B_t$).

To obtain the solution for the arbitrary matrix $M$, \textsc{LoRA-Pro} further consider the following optimization objective:
\begin{equation}
\arg\min_{M} \left\|\nabla A_t^* - \nabla A_t\right\|_F^2 + \left\|\nabla B_t^* - \nabla B_t\right\|_F^2,
\end{equation}
which can be optimized by solving the Sylvester equation:
\begin{equation}
M_tB_tB_t^\top + A_t^\top A_tM_t = - \frac{r}{\alpha^2}A_t^\top \nabla A_t (B_tB_t^\top)^{-1},
\end{equation}
which has a unique solution provided that $B_tB_t^\top$ and $-A_t^\top A_t$ do not have any shared eigenvalues.

Integrating \textsc{LoRA} with intermediate nonlinear functions is also a prevalent line of research~\cite{dong2025aurora,ji2024SineLoRA,li2024loran,liu2025loda}. Si et al.~\cite{si2024see} argue that the linear coupling mechanism in LoRA restricts its capacity to represent arbitrary rank-r matrices; therefore, introducing an intermediate \(r \times r\) matrix or nonlinearity function between \(A\) and \(B\) serves as a viable solution for a closer alignment with the learning capability of full fine-tuning. Dong et al.~\cite{dong2025aurora} assert that methodologies such as \textsc{\textbf{MoSLoRA}}~\cite{wu2024mixture} and \textsc{\textbf{FLoRA}} (Si et al.)~\cite{si2024flora}, which incorporate an intermediate \(r \times r\) matrix, preserve the linearity of LoRA and consequently restrict the exploration of broader parameter spaces.

\textsc{\textbf{LoDA(+)}}~\cite{liu2025loda} integrates LoRA with a multi-layer nonlinear activation structure to relax the low-rank linear constraint. Following the convention of PEFT methods, we denote the effective weight update induced by LoDA+ as $\Delta W$, defined implicitly through its action on the input:
\begin{equation}
    \frac{\alpha}{r}X \Delta W := \frac{\alpha}{r}(X AB + f_2\big(f_1(X AB)\big)),
\end{equation}
where $f_1(\cdot)$ and $f_2(\cdot)$ are parameterized nonlinear transformations (e.g., small linear layers with LeakyReLU activations). As illustrated in Figure 1 of the original paper, $f_1$ comprises two $r \times r$ matrices interleaved with three LeakyReLU functions, while $f_2$ consists of a single LeakyReLU function. $\Delta W$ in \textsc{LoDA+} does not correspond to a fixed low-rank matrix but rather an input-dependent mapping, and thus cannot be explicitly materialized as a standalone weight matrix. (The original paper did not specify the use of scaling for \textsc{LoDA+}; therefore, scaling is omitted here.)

Similar to \textsc{LoDA}+, \textsc{\textbf{Aurora}} enhances \textsc{LoRA} with an adaptive nonlinear layer (ANL) that combines both fixed and learnable nonlinear components. Formally, the effective weight update in Aurora can be expressed as:
\begin{gather}
\label{aurora_forward}
    f_{ANL}(M) = \mathrm{tanh}(\mathrm{tanh}(M)) + V_s \cdot \mathcal{S}(M),\\
    \frac{\alpha}{r}X \Delta W := \frac{\alpha}{r}f_{ANL}(XA)B,
\end{gather}
where the first term in the ANL function \(f_{ANL}(\cdot)\) represents a fixed nonlinear mapping implemented via the tanh activation function, and the second term introduces a learnable nonlinear mapping based on B-spline basis functions. Here, \(C \in \mathbb{R}^{r \times r}\) is an intermediate square matrix, \(\mathcal{S}(\cdot)\) is the spline basis function and \(V_s \in \mathbb{R}^{r}\) is the spline weight vector.  
\subsection{Initialization Adjustment Based \textsc{LoRA} Variants}
\label{initialization_variants}
\begin{figure}[t]
\begin{center}
\includegraphics[width=\columnwidth]{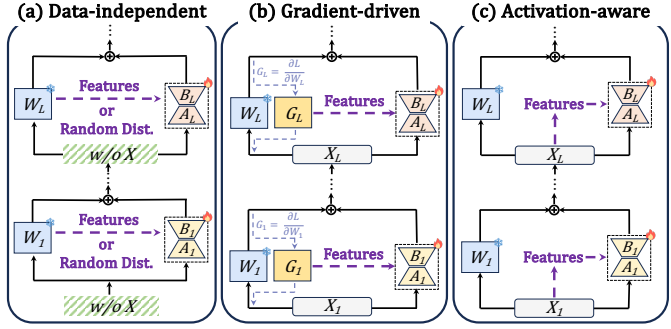}
\end{center}
\vspace{-3mm}
\caption{Illustration of initialization adjustment based \textsc{LoRA} variants.}
\vspace{-5mm}
\label{fig: init_methods}
\end{figure}
In the standard \textsc{LoRA} implementation, the gradients of $A$ and $B$ depend on each other's magnitudes (i.e., as indicated by Eq~\eqref{gradient}, initializing one matrix to zero causes the gradient of the other to vanish initially. For example, if $B_0 = 0$, then $\nabla A_0 = 0$, preventing updates to $A_t$ until $B_t \neq 0$.

This gradient suppression results in significantly slower convergence compared to full fine-tuning, particularly when using small learning rates. As discussed in Section~\ref{advantages_of_lora}, \textsc{LoRA} does not reduce the overall computational complexity of training relative to full fine-tuning. Consequently, the slower convergence can result in substantially more FLOPs to achieve comparable performance. As shown in Figure~\ref{fig: init_methods}, to address this limitation, while also aiming to improve performance, several advanced initialization strategies have been proposed.

\subsubsection{Data-independent Init Methods}
\label{subsubsec: random_init}
Typically, the initialization scheme of \textsc{LoRA} is defined as follows: (1) the weight matrix $A$ is initialized using either a Gaussian distribution (reported in the original paper) or a Kaiming uniform distribution (
    adopted in the \href{https://github.com/microsoft/LoRA/blob/c4593f060e6a368d7bb5af5273b8e42810cdef90/loralib/layers.py/\#L124}{official implementation} 
    and the \href{https://github.com/huggingface/peft/blob/47313792ddea21decbf0cad195cb880e7a487864/src/peft/tuners/lora/layer.py/\#L260}{PEFT library}), while (2) the weight matrix $B$ is initialized with zeros. Formally, when employing a Kaiming uniform distribution, the initialization can be expressed as:
\begin{equation}
\label{lora_init}
A_0 \sim \mathcal{U}\left(-\frac{1}{\sqrt{m}}, +\frac{1}{\sqrt{m}}\right), \quad B_0 = \mathbf{0}_{r\times n}.
\end{equation}
The original paper does not explore the potential differences between initializing matrix $A$ with zeros versus initializing matrix $B$ with zeros. Intuitively, one might assume these two initialization schemes exhibit similar performance. However, Hayou et al.~\cite{hayou2024impact-of-initialization} verify that under Kaiming init~\cite{he2016kaiming-init} or Lecun init~\cite{montavon2012lecun-init}, initializing matrix $B$ with zeros yields better performance and robustness to the learning rate.

Shiwei et al.~\cite{li2025beyond-zero-initialization} further explore a scheme where both matrices $A$ and $B$ are randomly initialized (referred to as \textsc{\textbf{NZLoRA}} in this paper). Given hyperparamters $\gamma_A$ and $\gamma_B$, \textsc{NZLoRA} can be formally expressed as:
\begin{equation}
A_0 \sim \mathcal{U}\left(-\frac{\gamma_A}{\sqrt{m}}, +\frac{\gamma_A}{\sqrt{m}}\right), \quad B_0 \sim \mathcal{U}\left(-\frac{\gamma_B}{\sqrt{m}}, +\frac{\gamma_B}{\sqrt{m}}\right).
\end{equation}
\textsc{NZLoRA} accelerates convergence by addressing the small gradient issue of vanilla \textsc{LoRA}, making it more robust to sub-optimal learning rates. A common challenge with non-zero initialization schemes is training instability. To mitigate this, existing methods typically adjust pretrained weights by subtracting the low-rank adapter's initial values—a process we term pretrained weights manipulation. However, this approach has a key limitation: since the pretrained weights must be modified again during inference, storing only the tuned low-rank adapters becomes infeasible. \textsc{NZLoRA} demonstrates that with carefully calibrated initialization variances (controlled by $\gamma_A$ and $\gamma_B$), pretrained weights manipulation can be safely omitted without compromising final fine-tuning performance.

The strategic initialization of low-rank adapters using pretrained weight statistics has become one of the dominant paradigms in data-independent initialization schemes. This methodology enables precise fine-tuning of targeted feature subspaces within pretrained weights.

\textsc{\textbf{PiSSA}}~\cite{meng2024pissa} laid the foundation for initialization from statistics of pretrained weights.
\textsc{PiSSA} initializes the adapter components using SVD of a pretrained weight matrix as:
\begin{gather}
\widetilde{W} = U_{\widetilde{W}} S_{\widetilde{W}} V_{\widetilde{W}}^\top, \\
A_0 = U_{\widetilde{W}}[:,:r]S_{\widetilde{W}}^{1/2}[:r,:r], \\
B_0 = S_{\widetilde{W}}^{1/2}[:r,:r]V_{\widetilde{W}}^\top [:r,:].
\end{gather}
Effectively capturing the most principal features of the original weight matrix according to the \textsc{Eckart-Young Theorem}~\cite{eckart1936Eckart, mirsky1960symmetric}. This inspired a series of subsequent works~\cite{azizi2024lamda, han2025dude, cao2024sorsa, guo2025nlora, lin2024nora}.

In contrast to this principal-component approach, \textsc{\textbf{MiLoRA}}~\cite{wang2024milora} exploits the minor components:
\begin{gather}
A_0 = U_{\widetilde{W}}{[:,-r:]} S_{\widetilde{W}}^{1/2}{[-r:,-r:]}, \\
B_0 = S_{\widetilde{W}}^{1/2}{[-r:,-r:]} V_{\widetilde{W}}^\top {[:,-r:]}.
\end{gather}
This preserves the primary knowledge in frozen weights while adaptively learning from the less dominant features.

\textsc{\textbf{OLoRA}}~\cite{buyukakyuz2024olora} uses QR decomposition for initialization as:
\begin{equation}
\widetilde{W} = QR, A_0 = Q[:,:r], B_0 = R[:r, :],
\end{equation}
where $Q \in \mathbb{R}^{m \times m}$ is orthonormal and $R \in \mathbb{R}^{m \times n}$ is upper triangular. This method achieves orthonormal initialization with computational efficiency.

Building upon these foundations, \textbf{SORSA}~\cite{cao2024sorsa} enhances orthonormal preservation through a regularization scheme. The method modifies \textsc{PiSSA}'s initialization while enforcing strict orthonormality constraints:
\begin{gather}
A_0 = U_{\widetilde{W}}{[:,:r]}, B_0 = V_{\widetilde{W}}^\top {[:r,:]}, D_0 = S_{\widetilde{W}}{[r:,r:]}, \\
\label{moslora-path}
W_t = \widetilde{W} + \Delta W_t = \widetilde{W} + A_tD_tB_t, \\
\mathcal{L}_\text{reg} := \left\|A_t^\top A_t-I\right\|^2_F + \left\|B_t B_t^\top-I\right\|^2_F,
\end{gather}
where $\mathcal{L}^{reg}$ represents the orthonormal regulation loss. This approach maintains the benefits of pretrained features while ensuring stable optimization through orthonormal constraints.

\subsubsection{Gradient-driven Init Methods}
\label{subsubsec:Gradient-based initialization}
As shown in Section~\ref{lora_vs_fft}, the optimization dynamics of \textsc{LoRA} adapters are closely tied to the corresponding gradients of the pretrained weights. Motivated by this insight, multiple gradient-driven methods are proposed to enhance performance. 


\textsc{\textbf{LoRA-GA}}~\cite{wang2024lora-ga} introduces a gradient-based initialization strategy for low-rank adapters by effectively leveraging pre-computed gradients. The central idea of \textsc{LoRA-GA} lies in its optimization objective, which explicitly minimizes the discrepancy between the weight updates at the initial training step induced by \textsc{LoRA} and those obtained through full fine-tuning with an arbitrary scaling factor $\zeta$. Formally, this objective can be expressed as the following minimization problem:
\begin{equation}
\arg\min_{A_0,B_0} \left\| \mathcal{P}(A_0,B_0,\nabla \widetilde{W}_0) - \zeta \nabla \widetilde{W}_0 \right\|_F^2,
\label{eq:lora_ga_objective}
\end{equation}
where the projection operator $\mathcal{P}(A_0,B_0,\nabla \widetilde{W}_0) \equiv A_0A_0^\top \nabla \widetilde{W}_0 + \nabla \widetilde{W}_0 B_0^\top B_0$ represents the approximate gradient of $\widetilde{W}$ as illustrated in Eq.~\eqref{low-rank-gradient}. Under the assumption of a single step of stochastic gradient descent (SGD) with a learning rate $\eta$, the objective in Eq.~\eqref{eq:lora_ga_objective} directly minimizes the difference between the updates of LoRA and full fine-tuning at the initial training step. 

Assuming the gradient matrix $\nabla \widetilde{W}$ is invertible and $2r < \text{min}(m,n)$, multiplying $A_0A_0^\top$ or $B_0^\top B_0$ by an invertible matrix does not alter its rank. Therefore, the maximum possible rank of $\mathcal{P}(A_0,B_0,\nabla \widetilde{W})$ is $2r$. In essence, \textsc{LoRA-GA} seeks to construct the optimal rank-$2r$ approximation of the gradient matrix $\nabla \widetilde{W}$ via $\mathcal{P}(A_0,B_0,\nabla \widetilde{W}_0)$.

The solution of Eq.~\eqref{eq:lora_ga_objective} can be derived from the truncated SVD of the pretrained weight gradient matrix as follows:
\begin{gather}
\label{lora-ga-s}
\nabla \widetilde{W}_0 = U_{\nabla {\widetilde{W}}_0}S_{\nabla {\widetilde{W}}_0}V_{\nabla {\widetilde{W}}_0}^\top, \\
\label{lora-ga-e}
A_0 = U_{\nabla {\widetilde{W}}_0}{[:,:r]}, \quad B_0=V^\top _{\nabla {\widetilde{W}}_0}{[r+1:2r,:]}.
\end{gather}

Following Eqs.~\eqref{lora-ga-s}--\eqref{lora-ga-e}, before the formal training phase, \textsc{LoRA-GA} efficiently computes and offloads the gradients of the pretrained weights layer by layer, resembling the fused gradient approach proposed in LOMO~\cite{lv2023fft-limited-resources} without performing optimization steps.

To further align with the scaling factor introduced by \textsc{RsLoRA}~\cite{kalajdzievski2023rslora} which we discussed in Section~\ref{training_dynamics_variants}, \textsc{LoRA-GA} incorporates the following scaling mechanism:
\begin{gather}
A_0 = \frac{\sqrt[4]{m}}{\sqrt{\gamma}} U_{\nabla \widetilde{W}_0}{[:,:r]}, \quad B_0 = \frac{\sqrt[4]{m}}{\sqrt{\gamma}} V^\top_{\nabla \widetilde{W}_0}{[r+1:2r,:]},
\end{gather}
where $ \gamma $ denotes a hyperparameter introduced by \textsc{LoRA-GA} to control the scaling.

\textsc{\textbf{LoRA-One}}~\cite{zhang2025loraone} identifies several purported misconceptions in \textsc{LoRA-GA} and proposes modifications to its SVD-based feature selection and scaling mechanisms. The authors contend that under \textsc{LoRA}'s standard zero-initialization scheme, as shown in Eq.~\eqref{lora_init}, the weight matrix \( B \) --- after the first training step---naturally resides in the top-\( r \) subspace of the right singular matrix of the gradient matrix. Furthermore, they argue that the subsequent training dynamics of \( B \) remain confined to this invariant subspace, while matrix \( A \) aligns with the top-\( r \) subspace of the left singular matrix of the gradient matrix under certain requirements.

Based on this premise, \textsc{LoRA-One} asserts that initializing \( B \) with \( V^\top_{\nabla \widetilde{W}_0}[r+1:2r,:] \) results in suboptimal alignment, trapping the optimization in an undesirable subspace. Instead, they claim \( B \) should align with \( V^\top_{\nabla \widetilde{W}_0}[:r,:] \). However, since this observation hinges on \textsc{LoRA}'s default zero-initialization scheme, its applicability to \textsc{LoRA-GA} --- which employs a different initialization strategy --- remains questionable.

Moreover, the indirect update to pretrained after the first gradient descent step of \textsc{LoRA-GA} is given by:
\begin{equation}
\begin{aligned}
\label{lora-ga-update}
&\frac{\alpha}{\sqrt{r}}A_1B_1 - \frac{\alpha}{\sqrt{r}}A_0B_0 = \frac{\alpha}{\sqrt{r}} \big[-\eta \nabla \widetilde{W}_0 B_0^{\top}B_0 \\
& - \eta A_0 A_0^{\top} \nabla \widetilde{W}_0 + \eta^2\nabla \widetilde{W} B_0^{\top}A_0^{\top}\nabla \widetilde{W}_0 \big].
\end{aligned}
\end{equation}
According to Eq.~\eqref{lora-ga-update}, the optimal $2r$-approximation of the initial gradient descent step's update direction can be achieved when the second-order $\eta^2$ term is negligible. This approximation, however, imposes an inherent constraint on the learning rate selection using \textsc{LoRA-GA}. 


Building upon these observations, \textsc{LoRA-One} introduces the following initialization:
\begin{gather}
-\nabla \widetilde{W}_0 = U_{\nabla \widetilde{W}_0}S_{\nabla \widetilde{W}_0}V_{\nabla \widetilde{W}_0}^\top, \quad S_{\nabla \widetilde{W}} \leftarrow S_{\nabla \widetilde{W}}/\sigma_1, \\
A_0 = \frac{1}{\sqrt{\gamma}} U_{\nabla \widetilde{W}_0}{[:,:r]}S_{\nabla \widetilde{W}_0}^{1/2}[:r,:r],\\
B_0 = \frac{1}{\sqrt{\gamma}} S_{\nabla \widetilde{W}_0}^{1/2}[:r,:r]V^\top_{\nabla \widetilde{W}_0}{[:r,:]},
\end{gather}
where $\gamma$ is a hyperparameter analogous to \textsc{LoRA-GA}'s scaling factor and $\sigma_1$ is the largest singular value of $-\nabla \widetilde{W}_0$. Remarkably, LoRA-One achieves recovery of the one-step gradient updates for pretrained weights—with negligible error—while eliminating the need for explicit weight manipulation discussed in Section~\ref{subsubsec: random_init}. \textsc{\textbf{LoRA-SB}}~\cite{ponkshe2024lora-sb} similarly initializes $A_0$ and $B_0$ using the top-$r$ left and right singular vectors respectively, while introducing $D \in \mathbb{R}^{r \times r}$ initialized with the corresponding singular values. During training, while maintaining the same forward pass formulation as Eq.~\eqref{moslora-path}, \textsc{LoRA-SB} keeps $A_0$ and $B_0$ frozen and only updates $D_t$.

\textsc{\textbf{GoRA}}~\cite{he2025gora} observes that the compression form shown in Eq.~\eqref{gradient_compression} is not the optimal solution given an initialized $A_0$. The best solution is given by:
\begin{equation}
A_0^{\dagger}=(A_0^\top A_0)^{-1}A_0^{\top}, \quad  A_0B_0 = -A_0A_0^{\dagger}\nabla \widetilde{W}_0 \approx -\nabla \widetilde{W}_0, 
\end{equation}
where $A^{\dagger}$ is the Moore-Penrose inverse of the matrix $A$. Furthermore, \textsc{GoRA} finds that the expected Frobenius norm of $\nabla \widetilde{W}_0$ is $\sqrt{mn}$, while that of $AB$ is $\sqrt{rn}$ under a zero-mean unit-variance distribution. Following these observations, \textsc{GoRA} initializes the low-rank weights by:
\begin{gather}
A_0 \sim \mathcal{U}\left(-\frac{1}{\sqrt{m}}, +\frac{1}{\sqrt{m}}\right),\quad B_0 = - \frac{\gamma \sqrt{m}}{\alpha}A_0^{\dagger}\nabla \widetilde{W}_0, \\
\frac{\alpha}{\sqrt{r}}A_0B_0 \approx - \frac{\gamma \alpha \sqrt{rmn}}{ \alpha\sqrt{rmn}} \nabla \widetilde{W}_0 \approx -\gamma \nabla \widetilde{W}_0,
\end{gather}
where $\gamma$ is a hyperparameter of \textsc{GoRA} that controls the scaling of initialization. With a proper $\gamma$, a lower initial loss and faster convergence speed can be achieved by \textsc{GoRA}.

\subsubsection{Activation-aware Init Methods}
\label{subsubsec:Activation-based initialization}
Let \( X \in \mathbb{R}^{bs \times m} \) denote the input activations of a pretrained weight matrix \( \widetilde{W} \in \mathbb{R}^{m \times n} \), where \( b \) is the batch size, \( s \) is the padded sequence length. The unnormalized covariance matrix \( C = X^\top X \) captures the second-order statistics of the inputs. To analyze how \( \widetilde{W} \) interacts with these input statistics, \textsc{CorDA} performs SVD on the matrix \( C\widetilde{W} = X^\top X \widetilde{W} \), which combines the data distribution \( C \) with the learned features \( \widetilde{W} \).
Formally, the decomposition can be expressed as follows:
\begin{gather}
C=X^\top X, \quad C\widetilde{W}=U_{C\widetilde{W}}S_{C\widetilde{W}}V_{C\widetilde{W}}^\top, \\
\widetilde{W} = C^{-1}C\widetilde{W} = (C^{-1}U_{C\widetilde{W}})S_{C\widetilde{W}}V_{C\widetilde{W}}^\top.
\end{gather}
This decomposition reveals task-relevant directions in the input space that are amplified or suppressed by \( W \).

Leveraging this decomposition, \textsc{\textbf{CorDA}}~\cite{yang2024corda} proposes two activation-based initialization schemes, namely \emph{knowledge-preserved adaptation} and \emph{instruction-previewed adaptation}. 

The key principle behind \emph{knowledge-preserved adaptation} is to retain the pretrained model's world knowledge as much as possible while adapting it to downstream tasks by altering minor directions. To operationalize this concept, \textsc{CorDA} employs the following methodology: First, it computes covariance matrices using question-answering datasets that are specifically selected for their relevance to the model's world knowledge representation.  Mathematically, after obtaining the covariance matrix, \textsc{CorDA} initializes the corresponding low-rank weights using the minor components of a weighted covariance matrix through the following transformation:
\begin{gather}
A_0 = (C^{-1}U_{C\widetilde{W}})[:,-r:]S_{C\widetilde{W}}^{-1/2}[-r:,-r:],\\
B_0 = S_{C\widetilde{W}}^{-1/2}[-r:,-r:]V_{C\widetilde{W}}^\top[-r:,:].
\end{gather}

In \emph{instruction-previewed adaptation}, the primary objective is to maximize alignment with the downstream task, prioritizing task-specific performance. For this purpose, \textsc{CorDA} computes the covariance matrices using a subset of the training dataset and initializes the low-rank weights as:
\begin{gather}
A_0 = (C^{-1}U_{C\widetilde{W}})[:,:r]S_{C\widetilde{W}}^{-1/2}[:r,:r],\\
B_0 = S_{C\widetilde{W}}^{-1/2}[:r,:r]V_{C\widetilde{W}}^\top[:r,:].
\end{gather}

As demonstrated in Section~\ref{initialization_variants}, \textsc{LoRA} faces the challenge of vanishing gradients during the initial training phases. To mitigate this issue, Paischer et al.~\cite{paischer2024eva} proposed \textbf{EVA} (\emph{Explained Variance Adaptation}), which utilizes 
principal components derived from the activation covariance matrix $X^\top X$ to 
properly initialize the weights of the matrix $A$. The primary objective of EVA is to maximize the expected gradient signal of the matrix $B$ during the initial training stages. Formally, this objective can be expressed as:
\begin{equation}
\max_{A_0A_0^\top=I} \mathbb{E} \left[\left\| \nabla B_0 \right\|_F^2\right] = \max_{A_0A_0^\top=I} \mathbb{E} \left[\left\| A_0^\top \nabla \widetilde{W} \right\|_F^2\right].
\label{eq:eva_objective}
\end{equation}

Consider the \textsc{LoRA} forward pass in a simple linear model where the input $x \in \mathbb{R}^{1 \times m}$ and output $\hat{y} \in \mathbb{R}^{1 \times n}$:

\begin{equation}
\hat{y} = x(\widetilde{W} + AB), \quad \nabla B_0 = A_0^\top x^\top \nabla \hat{y},
\end{equation}
where $\nabla \hat{y}$ represents the gradient of the predicted label $\hat{y}$ under the loss function $\mathcal{L}(\hat{y}, y)$. The expected squared Frobenius norm of the gradient of $B_0$ can then be derived as:
\begin{equation}
\begin{aligned}
\label{eva_nablab_fnorm}
\left\| \nabla B_0 \right\|_F^2 &= \mathrm{Tr}(\nabla B_0^\top \nabla B_0) = \mathrm{Tr}(\nabla \hat{y} \nabla \hat{y}^\top x A_0A_0^\top x^\top) \\
&= \underbrace{\nabla\hat{y} \nabla \hat{y}^\top}_{\mathrm{Scaler}}\cdot\mathrm{Tr}(A_0^\top x^\top x A_0).
\end{aligned}
\end{equation}

EVA makes the key assumption that the gradient of $\hat{y}$ is statistically independent of the input (i.e., $\nabla \hat{y} \perp x$), the gradient $\nabla \hat{y}$ depends solely on $\widetilde{W}$ since $A_0B_0=0$. Consequently, the expected covariance between the input and the gradient of $\hat{y}$ becomes:
\begin{equation}
\label{eva_input_output_independent}
\mathbb{E}\left[(x - \mathbb{E}[x])^\top(\nabla \hat{y} - \mathbb{E}[\nabla \hat{y}])\right] = \mathbf{0}_{m \times n}.
\end{equation}

This leads to EVA's fundamental conclusion that the expected initial gradient magnitude of $B_0$ is directly proportional to the trace of the activation matrix projected by $A_0$:
\begin{equation}
\mathbb{E}\left[\left\| \nabla B_0 \right\|_F^2\right] \propto \mathrm{Tr}(A^\top_0 \mathbb{E}\left[x^\top x\right] A_0).
\end{equation}
Therefore, the objective in Eq.~\ref{eq:eva_objective} can be rewritten as:
\begin{equation}
\max_{A_0A_0^\top=I} \mathbb{E} \left[\left\| \nabla B_0 \right\|_F^2\right] = \max_{A_0A_0^\top=I} \mathrm{Tr}(A^\top_0 \mathbb{E}\left[x^\top x\right] A_0). 
\end{equation}
This goal is equivalent to maximizing the variance of the down-projected activation $X_0A_0$ and maximizing the explained variance in a rank-$r$ approximation of the activation $X_0$. The solution can be derived from the truncated SVD:
\begin{gather}
\label{eva_init_s}
C = X^\top X = U_{C}S_{C}V_{C}^\top \\
A_0 = U_c[:,:r]S_C \in \mathrm{R}^{m \times r}, \quad B_0 = \mathbf{0}_{r \times n},
\label{eva_init_e}
\end{gather}
where $U_C$ contains the eigenvectors and $S_C$ contains the eigenvalues of $C$. For computational efficiency, EVA computes the covariance matrix $X^\top X$ using a subset of training data and employs incremental SVD~\cite{ross2008incremental} with truncation~\cite{halko2011finding} to minimize memory and time overheads during initialization. The final initialization of $A_0$ follows Eqs.~\eqref{eva_init_s}--\eqref{eva_init_e}.
\subsection{Mixture-of-Experts Integration Based \textsc{LoRA} Variants}
\label{moe_variants}

\begin{figure}[t]
\begin{center}
\includegraphics[width=\columnwidth]{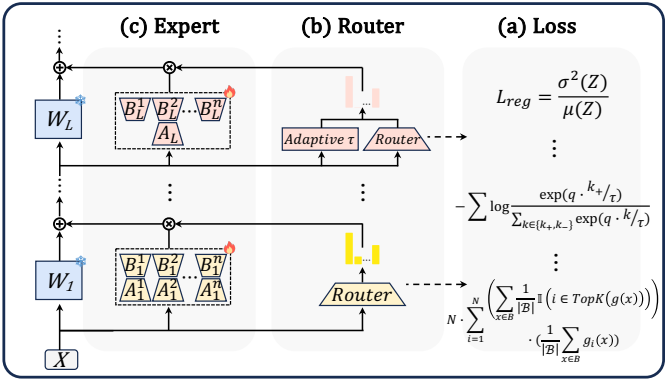}
\end{center}
\vspace{-3mm}
\caption{Illustration of mixture of experts Integration based \textsc{LoRA} variants.}

\vspace{-5mm}
\label{fig: moe_varients}
\end{figure}

By replacing standard LoRA layers with MoE modules composed of multiple LoRA experts, Mixture-of-Experts integration based \textsc{LoRA} variants aim to enhance model capacity and adaptability. Formally, the behavior of these variants is generally characterized by a routed combination of \textsc{LoRA} expert outputs. The adaptation form can be defined as:
\begin{equation}
\Delta W = \gamma_r \cdot \sum_{i=1}^N \omega_i(x) \cdot (B_i A_i),
\end{equation}
where $x$ is an input vector (hidden state of a token for LLMs), $N$ is the number of experts, and $\omega_i(x)$ is the routing weight for expert $i$ determined by a gating network $g(x)$. 
Here, the foundational router activates only the top $k$ experts based on the gating scores:
\begin{equation} 
\omega_i(x) = 
\begin{cases} 
g_i(x) & \text{if } i \in \text{TopK}(g(x)) \\ 
0 & \text{otherwise} 
\end{cases} 
\end{equation}

The optimization objective typically combines the primary task loss $L_{task}$ with a regularization loss $L_{reg}$ to regulate expert behavior.
In the standard setting, $L_{reg}$ typically serves as a load-balancing term to ensure even expert utilization. 
For a given batch of input tokens $\mathcal{B}$, this regularization loss is usually defined as the scaled dot-product between the expert selection frequency and the average gating probability:
\begin{equation}
L_{reg} = N \sum_{i=1}^N \left( \frac{1}{|\mathcal{B}|} \sum_{x \in \mathcal{B}} \mathbb{I}(i \in \text{TopK}(g(x))) \right) \cdot \left( \frac{1}{|\mathcal{B}|} \sum_{x \in \mathcal{B}} g_i(x) \right),
\end{equation}
where $N$ is the number of experts, $\mathbb{I}(\cdot)$ is the indicator function denoting whether expert $i$ is selected for token $x$, and the two summation terms represent the actual fraction of tokens routed to expert $i$ and the average predicted probability for expert $i$, respectively.

Recent variants have introduced innovations across three primary dimensions to this framework: modifications to the training objective (Loss), enhancements to the expert selection process (Router), and reconfigurations of the architectural design (Structure).

\subsubsection{Loss Modification Methods}
\label{subsubsec: loss_modification_methods}
While the standard regularization term $L_{reg}$ primarily focuses on load balancing to ensure equitable expert usage, it remains agnostic to the actual features learned by the experts. Consequently, this metric-driven constraint is insufficient for addressing the unique semantic challenges arising during fine-tuning, such as the tendency for experts to converge on identical representations or the overwriting of general world knowledge. To overcome these limitations, recent works have designed specialized auxiliary losses that go beyond simple routing statistics to actively shape expert specialization and diversity.

A key challenge in MoE training is random routing, where the gating network fails to develop strong preferences, causing different experts to converge on similar feature representations. This expert redundancy negates the capacity benefits of the MoE architecture. To address this, \textsc{\textbf{MoELoRA}}~\cite{luo2024moelora} incorporates a contrastive learning objective into the loss function. The motivation is to force experts to learn distinct features by maximizing the semantic distance between their outputs. Specifically, it treats the outputs processed by the same expert as positive pairs and those from different experts as negative pairs. The expert contrastive loss $L_{reg}$ is defined as:
\begin{equation}
L_{reg} = - \sum \log \frac{\exp(q \cdot k_+ / \tau)}{\sum_{k \in \{k_+, k_-\}} \exp(q \cdot k / \tau)},
\end{equation}
where $q$ is the query output, $k_+$ represents outputs from the same expert, $k_-$ represents outputs from other experts, and $\tau$ is the temperature. This auxiliary loss encourages high diversity among experts, ensuring that the expanded parameter space is effectively utilized for distinct feature processing.

While \textsc{MoELoRA} focuses on general expert diversity, other approaches leverage loss functions to enforce specific functional roles, particularly to mitigate "catastrophic forgetting." Standard fine-tuning often overwrites the model's pre-trained world knowledge while learning downstream tasks. \textsc{\textbf{LoRAMoE}}~\cite{dou2023loramoe} addresses this by introducing a localized balancing constraint to separate experts into two groups: those preserving world knowledge and those adapting to new tasks. It utilizes an importance matrix $Q$ and a coefficient matrix $I$ that rewards alignment between expert types and sample types:
\begin{equation} 
I_{n,m} = \begin{cases} 1 + \delta & \text{if } \text{Type}_e(n) = \text{Type}_s(m) \\ 1 - \delta & \text{otherwise} \end{cases} 
\end{equation}
The regularization loss is calculated based on the dispersion of the weighted importance matrix $Z = I \circ Q$:
\begin{equation}
L_{reg} = \frac{\sigma^2(Z)}{\mu(Z)}.
\end{equation}
By maximizing the variance of $Z$, \textsc{LoRAMoE} effectively disentangles task-specific adaptation from general knowledge retention, thereby solving the forgetting problem through guided expert specialization.

\subsubsection{Router Modification Methods}
\label{subsubsec: router_modification_methods}
The routing mechanism determines how inputs are distributed among experts. 
Standard approaches often rely on implicit, token-level routing, where different tokens within the same sequence might be sent to different experts based on latent features. This can lead to fragmented context and interference between heterogeneous tasks.
To address these limitations, \textsc{\textbf{MoA}}~\cite{feng2024mixture} diverges from the standard paradigm by adopting a sequence-level routing strategy guided by explicit domain metadata. Instead of relying on unsupervised token-wise learning, the model employs a regularization loss to penalize deviations from ground-truth domain labels. This supervision enforces precise, consistent data-to-expert assignments across all layer-wise routers, effectively mitigating task interference by prioritizing the sequence's domain identity over local token statistics.

Focusing on efficiency, \textsc{\textbf{AdaMoLE}}~\cite{liu2024adamole} challenges the static allocation of experts in Top-$k$ methods. Motivated by the observation that tokens vary in complexity, it introduces a dynamic, context-sensitive routing strategy. Instead of a fixed $k$, it employs an adaptive threshold $\tau(x)$ derived from the input features. An expert is activated only if its gating score exceeds this threshold:
\begin{gather}
\omega_i(x) = g_i(x) \cdot \mathbb{I}(g_i(x) > \tau(x)), \\
\text{where} \quad \tau(x) = \tau_{\text{max}} \cdot \sigma(W_\tau x + b_\tau).
\end{gather}
This mechanism allows the model to dynamically adjust the number of active experts based on the specific requirements of the input complexity.

\subsubsection{Expert Modification Methods}
\label{subsubsec: expert_modification_methods}
Beyond loss functions and routing, significant research focuses on structurally optimizing how LoRA experts are constructed, initialized, and arranged within the network.

\textsc{\textbf{MoLA}}~\cite{gao2024higher} is motivated by the understanding that different Transformer layers process features at varying levels of abstraction. Consequently, a uniform number of experts across all layers is suboptimal. \textsc{MoLA} structurally alters the MoE configuration by varying the number of experts $N_l$ specific to each layer $l$. By adopting architectures such as Diamond or Inverted-Triangle patterns for expert allocation, \textsc{MoLA} optimizes expert redundancy where it is most needed.

Another structural challenge is bridging the performance gap between LoRA-based methods and full fine-tuning. \textsc{\textbf{GOAT}}~\cite{fan2025make} addresses this by structurally aligning the initialization of experts with the singular value decomposition (SVD) of the pre-trained weights. Each LoRA expert is initialized using disjoint segments of the singular vectors $U$ and $V$:
\begin{equation}
B_i = \sqrt{\frac{1}{s}} U_i \Sigma_i^{1/2}, \quad A_i = \sqrt{\frac{1}{s}} \Sigma_i^{1/2} V_i^T.
\end{equation}
Combined with a theoretically derived scaling factor $s$, this structural initialization ensures that the optimization trajectory of the MoE-LoRA model closely mimics that of full-rank fine-tuning.

In the realm of structural efficiency, \textsc{\textbf{Hydra-LoRA}}~\cite{tian2024hydralora} addresses the parameter redundancy inherent in standard MoE designs where $(B_i, A_i)$ pairs are fully independent. Diverging from the symmetric expert structure, it proposes an asymmetric architecture consisting of a single shared $A$ and multiple $B$. The shared projection matrix $A$ captures general features across all inputs, while the set of distinct matrices $\{B_i\}$ serves as the experts. A dynamic router $g(x)$ computes input-sensitive weights to combine these heads, resulting in the update:
\begin{equation}
y = \widetilde{W} x + s \left( \sum_{i=1}^N \omega_i(x) B_i \right) A x.
\end{equation}
This design significantly reduces parameter count while maintaining the token-level adaptability of the router.

Conversely, \textsc{\textbf{MoSLoRA}}~\cite{wu2024mixture} modifies the internal structure of the LoRA module itself to introduce mixture capabilities without an external router.

\section{Overview of LoRAFactory}
\label{lorafactory}
\subsection{Core Implementations of LoRA in LoRAFactory}
LoRAFactory follows a modular inheritance hierarchy with \texttt{LinearWithLoRA} as the base class, enabling efficient implementation of \textsc{LoRA} variants through strategic method overriding. The \texttt{LinearWithLoRA} class extends \texttt{torch.nn.Linear} class and provides the foundations of LoRA's mechanism. The core forward computation of \texttt{LinearWithLoRA} is detailed below:
\begin{equation}
\label{fused_computation}
\begin{split}
&x_{\text{out}} = x_{\widetilde{W}} + x_{\Delta W} =\texttt{linear}(x_{\text{in}}, \widetilde{W}^\top) \\
&+ s \cdot \texttt{linear}\bigl(\texttt{linear}(\texttt{dropout}(x_{\text{in}}), A^\top), B^\top\bigr),
\end{split}
\end{equation}
where \(\texttt{linear}\) and \(\texttt{dropout}\) are functions provided by PyTorch. Key methods of \texttt{LinearWithLoRA} are:
\begin{itemize}
    \item \texttt{forward}: Orchestrates the forward pass as a linear layer. Conditionally applies low-rank adaptations (LoRA)---disabled either by the \texttt{DisableLoRA} context manager or when LoRA weights are inaccessible.
    \item \texttt{\_lora\_forward}: Computes the low-rank adaptation as shown in the fused forward pass of the low-rank adaptation part in Eq.~\eqref{fused_computation}.
    \item \texttt{init\_lora\_weights}: Initializes the low-rank weights.
    \item \texttt{compute\_lora\_weight}: Computes the effective LoRA weight \(\Delta W\).
\end{itemize}





A simple \texttt{LoRAConfig} data class is defined alongside, covering the following key configurations:
\begin{itemize}
    \item \texttt{in\_features: int}: The input dimension.
    \item \texttt{out\_features: int}: The output dimension.
    \item \texttt{bias: bool}: Whether a bias term is needed.
    \item \texttt{lora\_rank: int}: The rank of the low-rank adapter.
    \item \texttt{lora\_scaler: float}: The scaling coefficient of the low-rank adapter (defaultly used as \(\alpha\)).
    \item \texttt{lora\_dropout: float}: The dropout rate of the input of the low-rank adapter.
    \item \texttt{weight\_a\_init\_method: str}: The name of the initialization method for the matrix \(A\) (e.g., \texttt{kaiming}, representing the kaiming uniform distribution).
    \item \texttt{weight\_b\_init\_method: str}: The name of the initialization method for the matrix \(B\).
    \item \texttt{run\_lora\_in\_fp32: bool}: Whether run low-rank computation under the FP32 precision, while keeping the pretrained weight with original precision.
    \item \texttt{quant: bool}: Whether quantize the pretrained weight.
\end{itemize}











The class of \textsc{QLoRA} extends \texttt{LinearWithLoRA} by quantizing the layer's pretrained weights to lower precision (e.g., NF4). This implementation is highly flexible: it returns the output of \texttt{LinearWithLoRA} when \texttt{LoRAConfig.quant} is \texttt{False}; otherwise, it performs computation by de-quantizing the quantized weights. Hence, the \texttt{LinearWithQLoRA} is further inherited by classes of \texttt{LoRA} variant methods for easy quantization.

\subsection{Implementations of LoRA Variants in LoRAFactory}

Variants such as \textsc{DoRA} cannot directly use the forward pass of \texttt{LinearWithLoRA}; for example, \textsc{DoRA} requires merging the low-rank weights into pretrained weights and using the weights with altered magnitudes for forward computation, making the fused computation impossible. For example, the forward pass of \texttt{LinearWithDoRA} can be expressed as:
\begin{equation}
x_{\text{out}} = \texttt{linear}(x_{in}, \texttt{self.\_apply\_dora()}^\top),
\end{equation}
where \(\texttt{self.\_apply\_dora()}\) follows Eq.~\eqref{dora_weight}.

Variants such as \textsc{Aurora} necessitate forward passes of low-rank adaptation that differ from vanilla \textsc{LoRA}. For example, as shown in Eq.~\eqref{aurora_forward}, \textsc{Aurora} introduces a non-linear function \(f_{ANL}\) into the forward pass, the forward pass of \textsc{Aurora} can be correspondingly expressed as:
\begin{gather}
x_\text{ANL} = \texttt{self.\_ANL}(\texttt{linear}(\text{dropout}(x_{in}), A^\top)) \\
x_{\Delta W} = s \cdot \texttt{linear}
(x_\text{ANL}, B^\top),
\end{gather}
where \(\texttt{self.\_ANL()}\) is a method performing \(f_{ANL}\).

Variants such as \textsc{PiSSA}, relying on pretrained weights to initialize low-rank weights, are implemented by modifying the \texttt{init\_lora\_weights} method. In contrast, variants such as \textsc{LoRA-GA}, which utilize gradients or activations for initialization, require deactivating the \texttt{init\_lora\_weights} method and executing a variant-specific re-initialization function after computing and storing the gradients or activations.

Sharing-based variants share low-rank weights across modules; the low-rank weights of these variants cannot be directly initialized. For this reason, a variant-specific function \texttt{prepare\_shared\_lora\_weights} is required to identify all sets of modules that share low-rank weights and initialize the corresponding shared weights. After all sharing sets and shared weights are prepared, a variant-specific function \texttt{update\_shared\_weights\_to\_layer} is required to distribute the shared weights. 


\begin{figure*}[t]
\centering
\includegraphics[width=\textwidth]{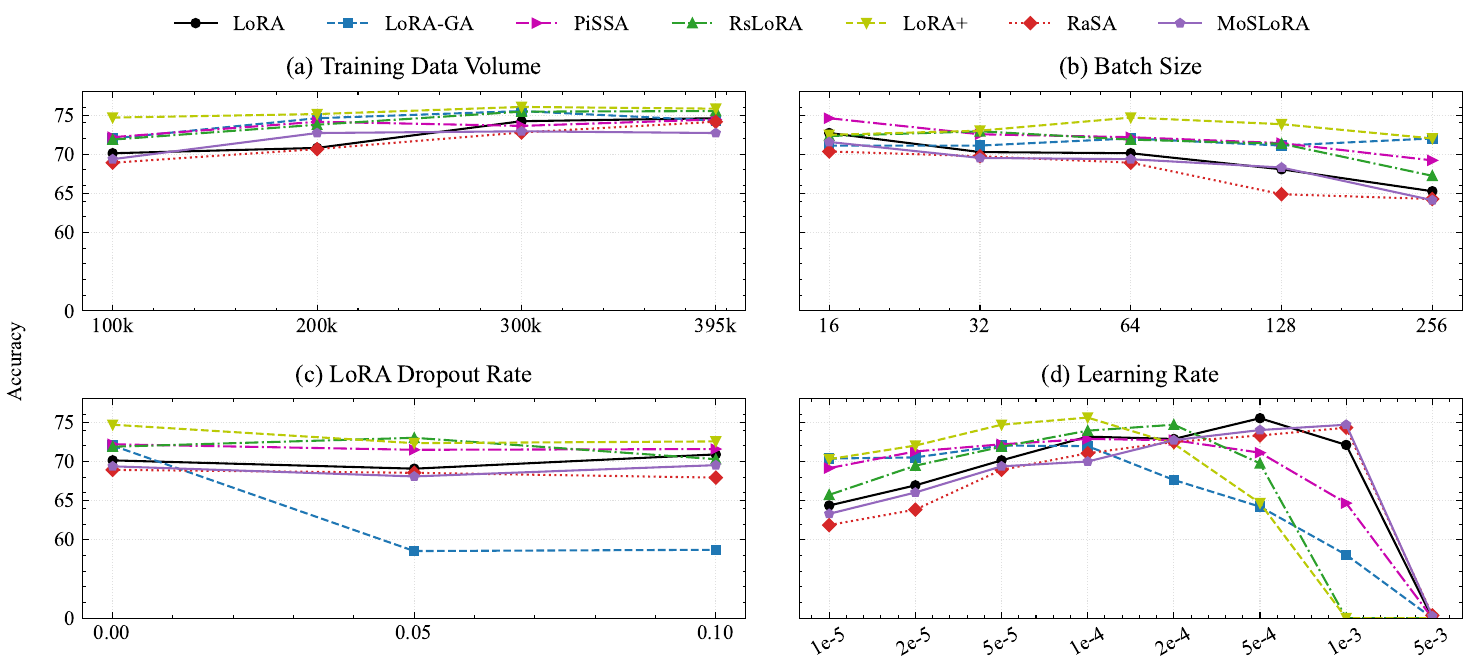}
\vspace{-7mm}
\caption{Performance of LoRA and selected variants with altering distinct hyperparameters.}
\label{fig:ablation_result}
\vspace{-5mm}
\end{figure*}

\subsection{Working Mechanism of LoRAFactory}

All LoRA-related hyperparameters within LoRAFactory are parsed via an argument parser and stored in a namespace variable, \texttt{args}. Following model initialization, both the model and \texttt{args} are passed to the \texttt{setup\_lora} function. This function identifies all targeted modules in the model designated for adaptation, and invokes the \texttt{switch\_to\_lora} function. The latter determines the targeted LoRA variant for adaptation and replaces all specified linear modules in the model with corresponding adaptation-class linear modules. Throughout this process, any exceptional cases are automatically managed by these functions. LoRAFactory is natively compatible with modern training strategies, including DeepSpeed ZeRO 3. The model with adapted modules can be trained using custom trainers, such as the Hugging Face Transformers Trainer, or it may employ the toolkits provided within the framework. 


\begin{figure*}[h]
\centering
\includegraphics[width=\textwidth]{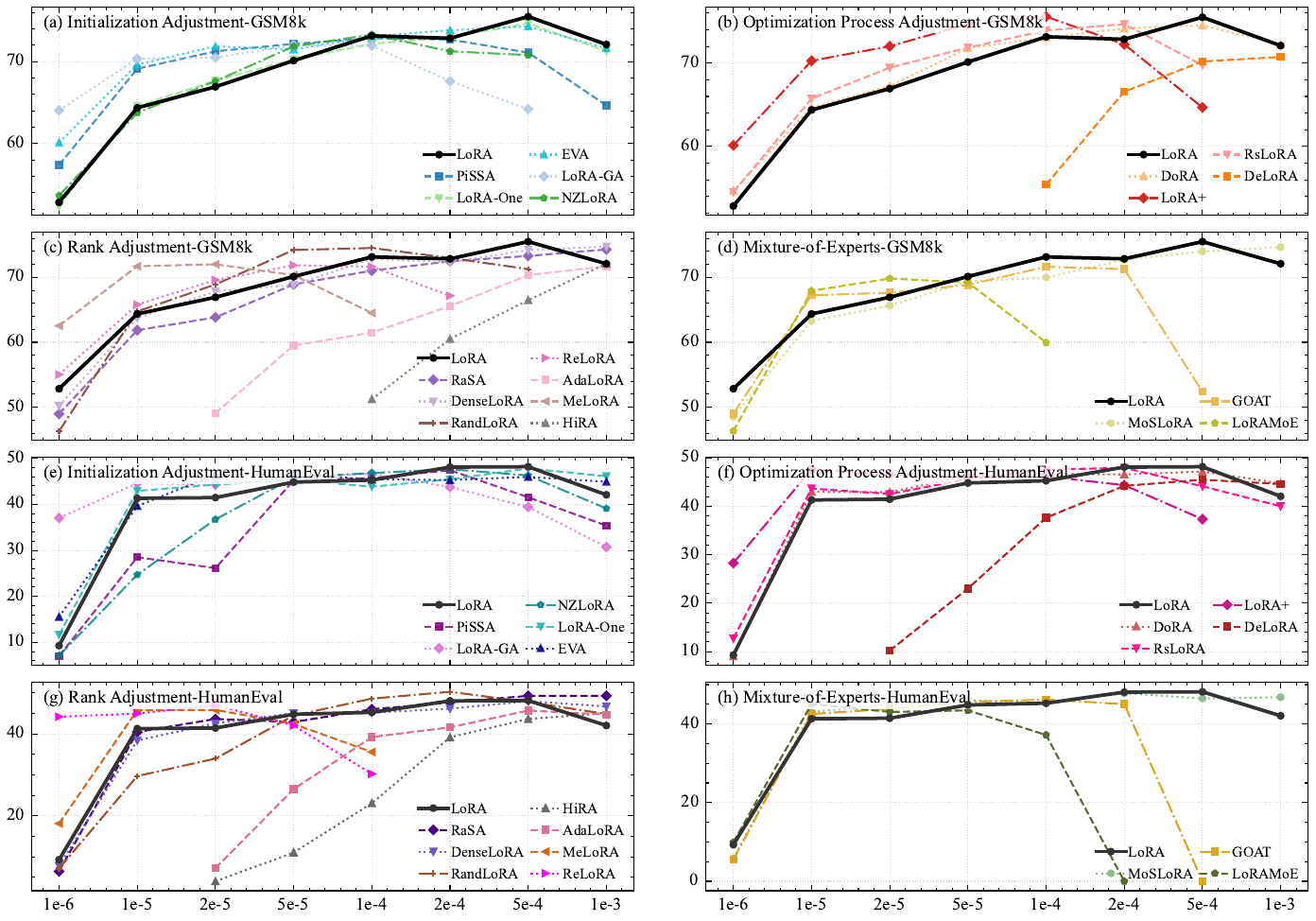}
\vspace{-5mm}
\caption{Performance comparison of LoRA and its variants on GSM8K (accuracy) and HumanEval (pass@1) across a range of learning rates.}
\vspace{-5mm}
\label{fig:nlg_result}
\end{figure*}

\begin{figure}[h!]
\centering
\includegraphics[width=\columnwidth]{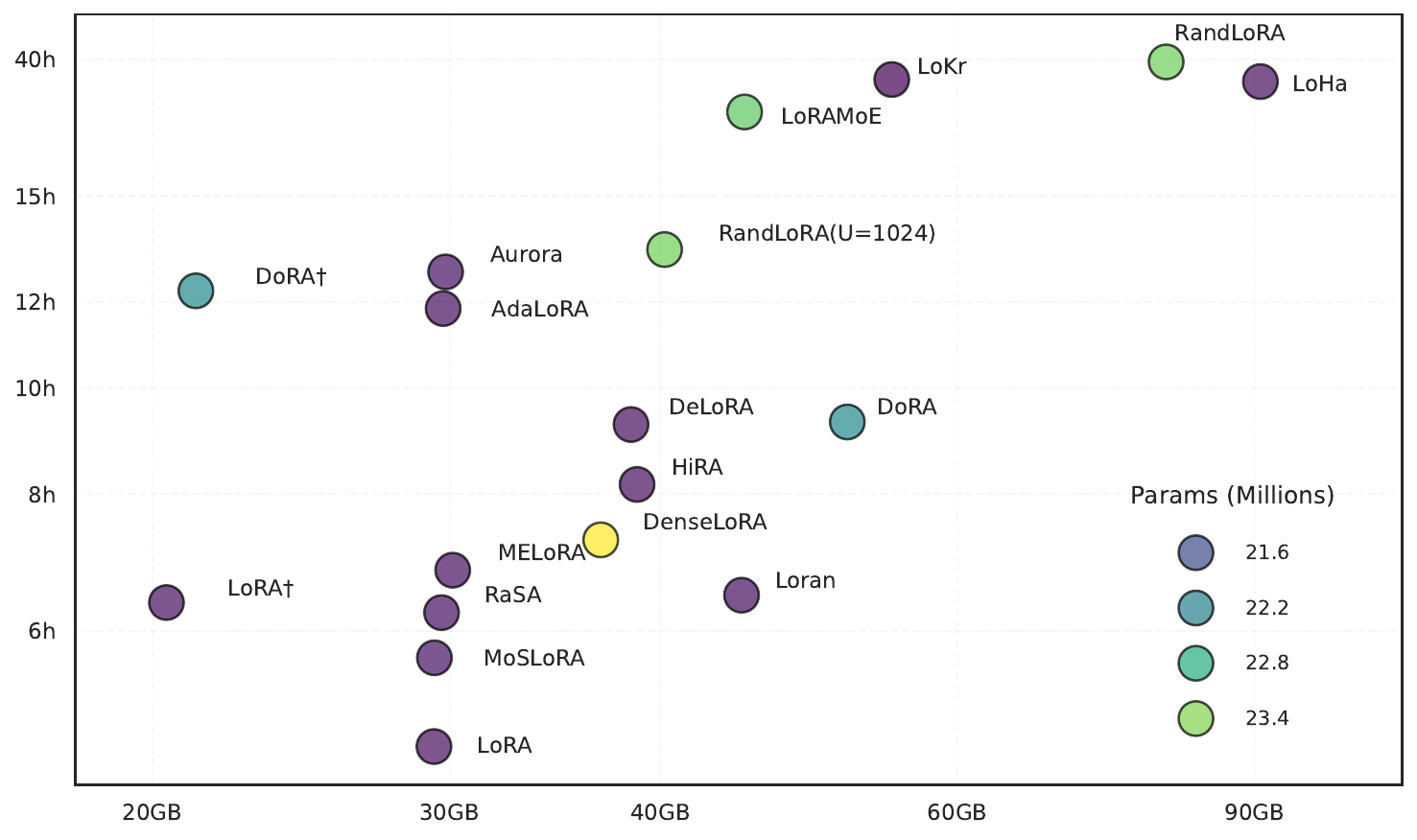}
\vspace{-3mm}
\caption{Computational and memory overhead of \textsc{LoRA} variants. \(\dagger\) dentes with activation checkpoint, and \(U\) is a hyperparameter of \textsc{RandLoRA}.}
\vspace{-5mm}
\label{fig:overhead}
\end{figure}

\section{Empirical Evaluation Using LoRAFactory}
\label{benchmark}
We present a systematic evaluation of 20 representative \textsc{LoRA} variants, which have been published in top-tier venues such as NeurIPS, ICLR, and ICML, within a unified framework implemented in our codebase, \textbf{LoRAFactory}. Our benchmark spans three domains: natural language understanding (NLU), natural language generation (NLG), and image classification (IC). A key challenge in comparing \textsc{LoRA} variants lies in their sensitivity to hyperparameter choices. While vanilla \textsc{LoRA} is known to be highly sensitive to the learning rate~\cite{biderman2024lora-learn-less, schulman2025lora-without-regret}, the sensitivity profiles of its variants remain largely uncharacterized. To address this, we conduct a comprehensive sensitivity analysis using Llama-3.1-8B-Base as a testbed, varying batch size, \textsc{LoRA} dropout rate, training data volume, and learning rate. Our results indicate that learning rate sensitivity is the most salient differentiator, with optimal ranges varying substantially across methods. Consequently, we fix all hyperparameters except the learning rate in the main experiments.

\noindent\textbf{Experimental Protocol.}  
For NLU, we fine-tune RoBERTa-Base~\cite{liu2019roberta} on all tasks of the GLUE benchmark~\cite{wang2018glue}. For NLG, we evaluate mathematical reasoning and code generation using Llama-3.1-8B-Base~\cite{dubey2024llama3}. For IC, we train CLIP-ViT-B/16~\cite{radford2021clip} on seven image classification datasets: Stanford-Cars~\cite{krause2013cars}, DTD~\cite{cimpoi2014dtd}, EuroSAT~\cite{helber2019eurosat}, GTSRB~\cite{houben2013gtsrb}, RESISC45~\cite{cheng2017resisc}, SUN397~\cite{xiao2010sun}, and SVHN~\cite{netzer2011svhn}. Given space limitations and consistent trends observed across domains, we present visualized NLG results in the main text, while full numerical results, including those of NLU and IC, are provided in Appendix C.

\noindent\textbf{Default Configuration.}  
All experiments use rank $r=8$, scaling coefficient $\alpha = 16$, and no \textsc{LoRA} dropout (except for variants like \textsc{DenseLoRA}, where larger $r$ and $\alpha$ are used to maintain comparable trainable parameter counts; see Appendix). We employ the AdamW optimizer~\cite{loshchilov2017adamw} with a cosine learning rate schedule. All runs use a fixed random seed; we observe that qualitative trends are robust to initialization.

\subsection{Computational and Memory Overhead Analysis}
Figure~\ref{fig:overhead} compares the training time and peak GPU memory usage of \textsc{LoRA} variants on Llama-3.1-8B-Base under a defaultly identical hardware and software conditions (single NVIDIA H200 GPU, BF16 precision, sequence length 1024, batch size 1, no activation checkpointing). Variants with negligible overhead (e.g., \textsc{LoRA+}) are omitted for clarity. Vanilla \textsc{LoRA} achieves the lowest overhead (4h 42m, 30,067 MB), serving as an efficiency baseline. \textsc{DoRA} incurs the highest memory cost (52,847 MB, +75\%) due to explicit materialization of low-rank matrices during forward propagation, a cost that can be mitigated via activation checkpointing. \textsc{LoRAMoE} is the slowest (36h 31m, +676.95\%) owing to its expert-routing mechanism. \textsc{AdaLoRA} and \textsc{RandLoRA} exhibit increased runtime due to dynamic rank allocation or full-rank computations. These results underscore a fundamental trade-off: architectural enhancements often come at the expense of computational and memory efficiency (\textbf{Finding 1}).

\subsection{Hyperparameter Sensitivity Analysis}
We fine-tune Llama-3.1-8B-Base on MetaMathQA~\cite{yu2023metamath} and evaluate on GSM8K~\cite{cobbe2021gsm}, using a base configuration of 100k samples, batch size 64, 0\% dropout, and learning rate 5e-5. In each ablation, only one hyperparameter is varied.

\subsubsection{Training Data Volume and Batch Size}
As shown in Figure~\ref{fig:ablation_result}(a), performance generally improves with data volume. Vanilla \textsc{LoRA} increases from 70.13 to 74.60 as data grows from 100k to 395k. However, variants like \textsc{LoRA-GA} and \textsc{MoSLoRA} saturate earlier. Performance remains roughly stable across batch sizes 16–64 (\textbf{Finding 2}), but degrades significantly at larger sizes due to fewer optimization steps. Vanilla \textsc{LoRA} drops from 72.71 to 65.28 at batch size 256, whereas \textsc{LoRA-GA} maintains robustness, attributable to its gradient-magnitude-enhanced initialization scheme, which reduces reliance on frequent updates.  Notably, inter-method performance gaps narrow with more update steps, with a smaller batch size or larger training data volume, suggesting that moderate-scale dataset volumes and update steps are more discriminative for evaluation (\textbf{Finding 3}).

\subsubsection{LoRA Dropout Rate}
Most variants are insensitive to dropout (\textbf{Finding 4}), but \textsc{LoRA-GA} suffers severe degradation (from 72.02 to 51.33). This stems from a mismatch between its initialization (computed without initialized low-rank weights) and training dynamics (with initialized low-rank weights and dropout), which alters input statistics. To ensure fair comparison, we disable dropout in all main experiments.

\subsubsection{Learning Rate}
Figure~\ref{fig:ablation_result}(d) reveals pronounced and method-specific learning rate sensitivity (\textbf{Finding 5}), with narrow and non-overlapping optimal ranges. This necessitates extensive learning rate sweeps to find out the near-optimal performance of each method we tested, as detailed next.

\subsection{Learning Rate Sweep Results on NLG}
\subsubsection{Task Settings}
We evaluate mathematical reasoning by fine-tuning on 100k samples from MetaMathQA~\cite{yu2023metamath} and testing on GSM8K~\cite{cobbe2021gsm}. For code generation, we train on 100k samples from CodeFeedback~\cite{zheng2024opencodeinterpreter} and evaluate on HumanEval~\cite{chen2021codex}. To mitigate variance from HumanEval’s small size (163 samples), we average over eight evaluation runs. We sweep eight learning rates (1e-6 to 1e-3) while fixing other settings with the default configurations.

\subsubsection{Results and Discussion}
As shown in Figure~\ref{fig:nlg_result}, performance typically rises with learning rate until an optimum, beyond which it declines. Several variants (e.g., \textsc{AdaLoRA}) converge slower than vanilla \textsc{LoRA}, likely due to additional regularization, for example, orthogonality constraints in \textsc{AdaLoRA} or coupling with pretrained weights (e.g., \textsc{HiRA}, which dampens gradient signals.

Notably, many variants outperform vanilla \textsc{LoRA} at low learning rates. On GSM8K at 1e-6, \textsc{LoRA-GA} (64.06) surpasses vanilla \textsc{LoRA} (52.82) by over 11 points. However, this advantage vanishes at higher rates: the best variant (\textsc{LoRA+}, 75.59 at 1e-4) exceeds \textsc{LoRA} (75.51 at 1e-4) by only 0.08 points. On HumanEval, only \textsc{RandLoRA} and \textsc{RaSA} marginally surpass \textsc{LoRA}. At 1e-6, most methods fail to achieve meaningful code generation performance, indicating that stronger update signals are essential for this task.

Surprisingly, \textsc{LoRA} exhibits higher performance ceilings than its most evaluated variants on both tasks, a trend also observed in NLU and IC experiments (\textbf{Key Finding}). This phenomenon arises from the \textit{small-gradient issue} in vanilla \textsc{LoRA} combined with improper hyperparameter configurations such as small learning rates or scaling factor, limited update steps, its parameter updates are not sufficient, hindering optimization. In contrast, variants like \textsc{LoRA-GA} produce initial gradients $\sim$100$\times$ larger, enabling effective learning in certain suboptimal hyperparameter regimes. However, when a proper set of hyperparameter configurations is adopted, compensating for vanilla \textsc{LoRA}’s inherent small gradients, thereby neutralizing the relative advantage of these variants.

Our findings suggest that prior studies, which often evaluate the performance using a fixed hyperparameter configuration, may have underestimated the performance of the most important baseline: \textsc{LoRA}. Performance gains frequently disappear under comprehensive hyperparameter sweeps, underscoring the necessity of broad hyperparameter exploration for fair and robust evaluation of \textsc{LoRA} methods.
\section{Conclusion}

In this work, we conduct a unified study of \textsc{LoRA} and its variants. We organize all methods into four categories, establishing a fine-grained and structured taxonomy based on their principal operational axes. Further, we unify them under a review framework of low-rank update dynamics, illuminating their connections. Empirically, we introduce LoRAFactory, a modular and extensible codebase that implements 50+ \textsc{LoRA} variants. Through extensive large-scale experiments, several key findings emerge. These results underscore the robust performance of the fundamental baseline, \textsc{LoRA}, and emphasize the critical role of hyperparameter tuning, specifically the calibration of the learning rate, in ensuring equitable benchmarking within \textsc{LoRA} research.
By releasing all code and configurations, we hope this work provides a foundation for more rigorous and transparent evaluation. 



\clearpage
\bibliographystyle{IEEEtran}
\bibliography{reference}
\clearpage
\appendix

\subsection{Related Works}

\subsubsection{Survey of PEFT and LoRA}
Despite considerable survey attention on PEFT, its most influential method, LoRA, and its proliferating variants receive superficial coverage. Existing surveys treat LoRA as a minor component within broader PEFT taxonomies, offering only cursory lists or brief summaries. For example, Han et al.~\cite{han2024han-peft} include a short section on “Reparameterized PEFT” without detailed analysis or mathematical formulations. Wang et al.~\cite{wang2024wang-peft} catalog over 10 LoRA-inspired methods but provide merely enumerative descriptions lacking systematic categorization. Xu et al.~\cite{xu2023xu-peft} list 11 LoRA variants with formulations but do not analyze their underlying mechanisms or comparative advantages. Mao et al.~\cite{mao2025mao-lora} and Yang et al.~\cite{yang2024yang-lora} conduct comprehensive surveys yet still offer lists with brief explanations rather than deeper insights. In summary, while acknowledging LoRA's popularity, these surveys lack a principled and in-depth examination. This gap motivates our work: a dedicated, systematic, and analytical survey tracing the evolution of LoRA variants, dissecting their innovations, and evaluating their trade-offs.

\subsubsection{Evaluation of LoRA and its Variants}
Evaluating LoRA and its variants is complex. The original LoRA paper~\cite{hu2022lora} benchmarks LoRA against PEFT methods including BitFit~\cite{zaken2021bitfit} and adapter tuning~\cite{houlsby2019adapter-tuning} on models like RoBERTa~\cite{liu2019roberta}, DeBERTa~\cite{he2020deberta}, GPT-2~\cite{radford2019gpt2}, and GPT-3-175B~\cite{brown2020gpt3}. Several follow-up studies~\cite{qiang2024bilora,zhang2023adalora,zi2023delta-lora} adhere to similar pipelines to demonstrate advantages over vanilla LoRA. However, the NLU evaluation pipeline for vanilla LoRA requires intensive hyperparameter grid searches, hindering large-scale comparisons. Its NLG pipeline uses outdated models like GPT-2/GPT-3 on tasks such as WikiSQL~\cite{zhong2017seq2sql} and MNLI~\cite{wang2018glue}, which are not representative of current frontier models like the Llama3 series~\cite{dubey2024llama3} or modern NLG scenarios. Recent LoRA variants are also evaluated on vision tasks alongside NLU and NLG. For NLU, the GLUE benchmark with models like RoBERTa, DeBERTa, and T5~\cite{raffel2020t5} is common, with RoBERTa being the most frequent choice. For NLG, LLMs such as the Llama series~\cite{dubey2024llama3,touvron2023llama-2,touvron2023llama} and Gemma series~\cite{team2024gemma,team2024gemma2,team2025gemma3} are evaluated on commonsense reasoning, chat, mathematical reasoning, and code generation. For vision, ViT~\cite{dosovitskiy2020vit} and CLIP-ViT are commonly tested on image classification tasks. Considering this, our comprehensive evaluation tests RoBERTa-Base on GLUE for NLU, Llama-3.1-8B-Base on mathematical reasoning and code generation for NLG, and CLIP-ViT-16/B on seven image classification tasks for vision performance.

\subsection{Notations}

This section delineates the notations utilized throughout this paper. Unless otherwise indicated, all notations conform to the definitions presented in Table~\ref{tab:notations}.

\begin{table*}[htbp]
\centering
\caption{List of Notations}
\label{tab:notations}
\begin{tabularx}{\textwidth}{l X}
\toprule
\textbf{Symbol} & \textbf{Description} \\
\midrule
$W \in \mathbb{R}^{m \times n}$ & Weight matrix of a linear layer. \\
$\widetilde{W} \in \mathbb{R}^{m \times n}$ & pretrained weight of $W$. \\
$\Delta W$ & Update applied to the weight $W$ during fine-tuning. \\
$A \in \mathbb{R}^{m \times r}$, $B \in \mathbb{R}^{r \times n}$ & Trainable low-rank matrices in the standard LoRA decomposition. \\
$r$ & Rank hyperparameter of LoRA and its most variants, with $r \ll \min(m, n)$. \\
$\alpha$ & Scaling hyperparameter of LoRA and its most variants. \\
$\gamma_r$ & Scaling factor of LoRA, \(\gamma_r \to 0\) as \(r \to \infty\). \\
$\eta$ & Learning rate used for parameter updates. \\
$\nabla \widetilde{W}$ & Gradient of the pretrained weight $\widetilde{W}$. \\
$\nabla A$, $\nabla B$ & Gradients of the low-rank matrices $A$ and $B$. \\
$W_t$ & Weight matrix of a linear layer at fine-tuning step $t$ ($W_t = \widetilde{W} + \Delta W_t$). \\
$A_t$, $B_t$ & Values of the low-rank matrices $A$ and $B$ at fine-tuning step $t$. \\
$A_0$, $B_0$ & Initial values of the low-rank matrices $A$ and $B$. \\
$\text{SVD}(M)$ & Singular Value Decomposition of matrix $M$. \\
$\text{Tr}(M)$ & Trace of matrix $M$. \\
$U, S, V$ & Matrices from the SVD of a matrix, i.e., $M = U S V^\top$. \\
$\odot$ & Hadamard (element-wise) product of two matrices. \\
$\otimes$ & Kronecker product of two matrices. \\
$\bigoplus$ & Block-diagonal matrix constructor. \\
$[M_1|M_2\dots|M_n]$ & Matrix concatenation operator. \\
$\mathcal{R}(M)$& Rank of matrix $M$. \\
$\|\cdot\|_F$ & Frobenius norm of a matrix. \\
$\mathcal{L}_{\text{task}}$ & Primary task-specific loss function (e.g., cross-entropy). \\
$\mathcal{L}_{\text{reg}}$ & Auxiliary regularization loss (e.g., for orthogonality in AdaLoRA). \\
LoRAFactory & A unified, modular codebase developed in this work for benchmarking LoRA variants. \\
LLMs & Large Language Models \\
NLU & Natural Language Understanding. \\
NLG & Natural Language Generation. \\
IC & Image Classification. \\
PEFT & Parameter-Efficient Fine-Tuning. \\
MoE & Mixture of Experts. \\
\bottomrule
\end{tabularx}
\end{table*}

\subsection{Additional Experimental Results}

Due to space constraints within the main body of the paper, this section presents supplementary experimental results of significance.

\subsubsection{Computational and Memory Overhead Analysis}
\label{sec:cost}

Table~\ref{tab: cost} presents the numerical training time and peak memory usage of some variants implemented in LoRAFactory (variants such as \textsc{LoRA+}, which do not affect the efficiency of \textsc{LoRA}, are not presented in this table). To ensure intrinsic efficiency is measured without masking effects from external optimizations, experiments were conducted on a single NVIDIA H200 GPU using the Llama-3.1-8B-Base model (BF16 mixed precision, sequence length 1024, batch size 1) without parallelism, CPU offloading, or activation checkpointing.

Vanilla \textsc{LoRA} serves as the foundational baseline, achieving the lowest memory footprint (30,067 MB) and the fastest training time (4h 42m). In contrast, \textsc{DoRA} exhibits the highest memory consumption (52,847 MB, $+75\%$), primarily because methods like \textsc{DoRA}, \textsc{HiRA}, and \textsc{LoHa} explicitly materialize low-rank matrices ($A$ and $B$) and their product during the forward pass. Vanilla LoRA avoids this by fusing computation of low-rank weights, a benefit that persists unless activation checkpointing is applied. As the router in each \textsc{LoRAMoE} module introduces additional trainable parameters, we compare the computational and memory overhead of \textsc{LoRAMoE} with two settings: 8 total experts and 2 activated experts per module; 6 total experts and 2 activated experts per module, each expert with a rank of \(1\). Therefore, the prior setting has an identical overall rank of \(8\) with \textsc{LoRA}, but it introduces significantly more trainable parameters. The latter setting has a comparable number of trainable parameters with \textsc{LoRA} but a smaller overall rank. Both settings of \textsc{LoRAMoE} require significantly longer training durations compared to \textsc{LoRA} due to the mixture-of-experts architecture, which introduces significant overhead through learnable token routers. While \textsc{MoSLoRA}, \textsc{RaSA}, and \textsc{MELoRA} maintain memory profiles similar to vanilla LoRA with moderate speed trade-offs, methods like \textsc{AdaLoRA} and \textsc{RandLoRA} suffer from prolonged training times due to dynamic rank allocation or high-rank computation strategies (The official implementation of \textsc{RandLoRA} adopts full-rank computation, which leads to high computational cost. We limit the upper bound of the dimension of random bases of \textsc{RandLoRA} to 1024, denoted as \textsc{RandLoRA\(_{U=1024}\)}). These results highlight the inherent tension between expressiveness and efficiency in LoRA extensions.

\begin{table*}[htbp]
\centering
\small
\caption{Computational and memory usage of \textsc{LoRA} variants. $\dagger$ denotes the use of activation checkpointing.}
\label{tab: cost}
\begin{tabular}{l *{3}{>{\centering\arraybackslash}p{2.2cm}}}
\toprule
\textbf{Method} & 
\textbf{\#Params} & 
\textbf{Time} & 
\textbf{Memory} \\
\midrule
\textsc{LoRA}~\cite{hu2022lora}         & 20.97M & 4h42min  & 30067MB \\
\textsc{LoRA$^\dagger$}~\cite{hu2022lora}         & 20.97M & 6h22min  & 20873MB \\
\textsc{DoRA}~\cite{liu2024dora}        & 22.35M & 9h19min  & 52847MB \\
\textsc{DoRA$^\dagger$}~\cite{liu2024dora}        & 22.35M & 12h17min & 21729MB \\
\textsc{DeLoRA}~\cite{bini2025delora}       & 20.97M & 9h16min  & 39343MB \\
\textsc{AdaLoRA}~\cite{zhang2023adalora}      & 20.97M & 12h10min & 30447MB \\
\textsc{HiRA}~\cite{huang2024hira}         & 20.97M & 8h10min  & 39669MB \\
\textsc{RaSA}~\cite{he2025rasa}         & 20.98M & 6h14min  & 30381MB \\
\textsc{DenseLoRA}~\cite{mu2025denselora}    & 23.99M & 7h16min  & 37751MB \\
\textsc{RandLoRA}~\cite{albert2025randlora}     & 23.30M & 39h31min & 81645MB \\
\textsc{RandLoRA\(_{U=1024}\)}~\cite{albert2025randlora}     & 23.30M & 13h24min & 41187MB \\
\textsc{MELoRA}~\cite{ren2024melora}& 20.97M & 6h49min  & 30847MB \\
\textsc{LoRAMoE$_{e=8}$}~\cite{dou2023loramoe}      & 30.93M & 36h31min & 46425MB \\
\textsc{LoRAMoE$_{e=6}$}~\cite{dou2023loramoe}      & 23.20M & 27h57min & 45937MB \\
\textsc{MoSLoRA}~\cite{wu2024mixture}      & 20.99M & 5h50min  & 30085MB \\
\textsc{Aurora}~\cite{dong2025aurora}       & 21.00M & 12h32min  & 30549MB \\
\textsc{LoHa}~\cite{hyeon2021fedpara}       & 20.97M & 35h28min  & 92851MB \\
\textsc{LoKr}~\cite{yeh2023lokr}       & 20.86M & 35h53min  & 56151MB \\
\textsc{Loran}~\cite{li2024loran}       & 20.97M & 6h28min  & 45747MB \\
\bottomrule
\end{tabular}
\end{table*}

\subsubsection{Learning Rate Sweep Results on NLU Tasks}

\textbf{Experimental settings.}
We fine-tune RoBERTa-base to evaluate LoRA and its variants on the full GLUE benchmark. We follow standard evaluation metrics for each GLUE sub-task: accuracy for SST-2, MNLI, MRPC, QNLI, and RTE; Pearson correlation for STS-B; F1 for QQP; and Matthews Correlation Coefficient for CoLA. We employ a linear learning rate decay schedule with a warm-up ratio of 0.03. Seven learning rates are tested for all methods, including: [1e-6, 1e-5, 5e-5, 1e-4, 5e-4, 1e-3, 5e-3]. 

All experimental settings remain consistent across runs. The batch size is 32, weight decay is disabled, and the maximum sequence length is 256 for all GLUE sub-tasks. Both the base model and all LoRA modules operate in FP32 precision without mixed-precision training. Each training run consists of 10 epochs, with the test performance recorded at the end of each epoch; the best test result across all 10 epochs is reported.

\textbf{Experimental results.}
\begin{figure*}[ht]
\centering
\includegraphics[width=\textwidth]{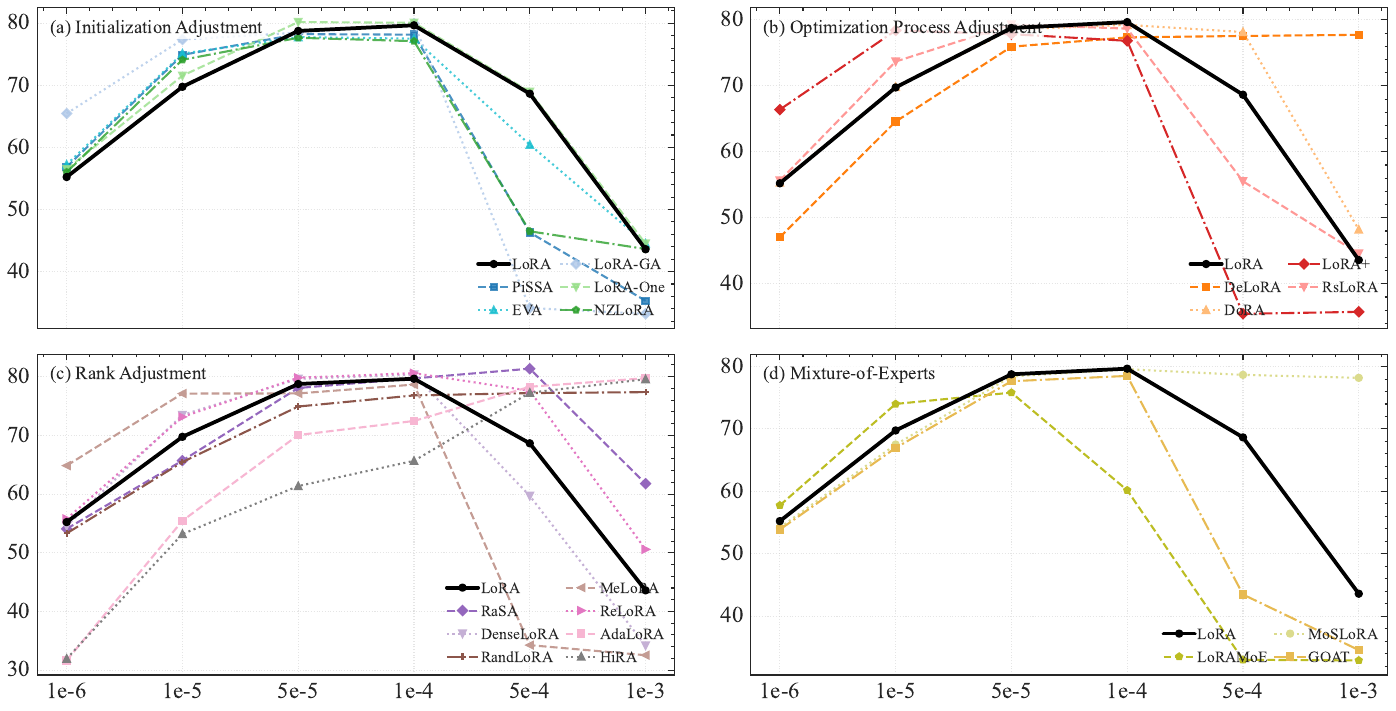}
\caption{Performance comparison of various LoRA variants on the GLUE benchmark across different learning rates. Results are grouped by method category as illustrated in Section \textrm{II}. All plots share the same y-axis (averaged numerical score) and x-axis (learning rate).}
\label{fig:glue_result}
\end{figure*}
Figure~\ref{fig:glue_result} shows the average evaluation results on the 9 subsets of the GLUE benchmark. For the detailed numerical results of each subset, please refer to Table~\ref{tab:glue 1e-6}-~\ref{tab:glue 5e-3}. 

The average performance of \textsc{LoRA} attains a relatively modest value of 55.23 at the lowest learning rate examined, with performance progressively increasing to 79.66 as the learning rate rises to 1e-4. Among all tested variants, only \textsc{RaSA} significantly exceeds LoRA in peak performance, achieving 81.37 at the learning rate of 5e-4.

All initialization-based \textsc{LoRA} variants significantly enhance gradient flow at small learning rates (e.g., 1e-5), leading to improved performance in such configurations. Similarly, \textsc{MELoRA} incorporates a block-diagonal structure, which also amplifies the gradient magnitude in \textsc{LoRA}. However, these enhancements come at a cost: the improved gradient properties hinder convergence at higher learning rates, preventing these methods from reaching or surpassing the peak performance achievable with standard \textsc{LoRA} under such settings. Simultaneously, \textsc{LoRA+} implements a learning rate decoupling strategy for the low-rank matrices \(A\) and \(B\) within low-rank adapters. For \textsc{LoRA+}, it is recommended that the learning rate for matrix \(B\) be set to 16 times that of matrix \(A\). This approach effectively applies a higher learning rate to matrix \(B\) compared to standard LoRA under an equivalent base learning rate, which governs matrix \(A\) and other trainable parameters. In practice, while these variants perform well with smaller learning rates, their effectiveness diminishes, occasionally sharply, when larger learning rates are employed.

In contrast, the auxiliary loss in \textsc{AdaLoRA}, the weight decomposition strategies in \textsc{DoRA} and \textsc{DeLoRA}, and the Hadamard product operation between low-rank and pretrained weights in \textsc{HiRA} significantly impede convergence at small learning rates. For instance, \textsc{HiRA} achieves only 53.19 at a learning rate of 1e-5, which is 16.6 points lower than \textsc{LoRA} under the same setting. As a result, these methods require learning rates that are about 10 to 1000 times larger than those used by LoRA to achieve comparable performance. It should be noted that \textsc{LoRA} itself typically employs learning rates 10 to 100 times higher than those commonly used in full fine-tuning. Notably, most of these methods do not explicitly state in their original papers that they require such elevated learning rates. This observation highlights the importance of carefully selecting learning rates when applying these methods in practice, as their optimal values may differ significantly from those used in standard fine-tuning approaches.

For Mixture-of-Experts integration based LoRA Variants, both \textsc{LoRAMoE} and \textsc{GoAT} fail to surpass the performance of LoRA on most learning rates we tested, while requiring substantially more training and inference time. One possible reason is that under similar trainable parameter budgets, the inherently sparse structure of MoE-based methods can limit their overall performance. As a non-traditional mixture-of-experts approach, \textsc{MoSLoRA} introduces a small intermediate matrix to the low-rank adapter, demonstrating greater stability with respect to learning rate selection than \textsc{LoRA}. Our experiments on NLG and IC also validate this observation.

\subsubsection{Learning Rate Sweep Results on IC Tasks}
\textbf{Experimental settings}
For image classification tasks, we fine-tune CLIP-ViT-16/B on seven benchmark datasets: Stanford-Cars~\cite{krause2013cars}, DTD~\cite{cimpoi2014dtd}, EuroSAT~\cite{helber2019eurosat}, GTSRB~\cite{houben2013gtsrb}, RESISC45~\cite{cheng2017resisc}, SUN397~\cite{xiao2010sun}, and SVHN~\cite{netzer2011svhn}. Each method’s classification accuracy is evaluated on the corresponding test set of each task. Experiments are repeated with ten distinct learning rates: [1e-6, 1e-5, 2e-5, 5e-5, 1e-4, 2e-4, 5e-4, 1e-3, 2e-3, 5e-3]. Other experimental settings, including the optimizer (and its configurations), learning rate scheduler (and its configurations), and batch size, remain identical to those used in our NLG experiments and are held constant across all runs.

\begin{figure*}[t]
\begin{center}
\includegraphics[width=\textwidth]{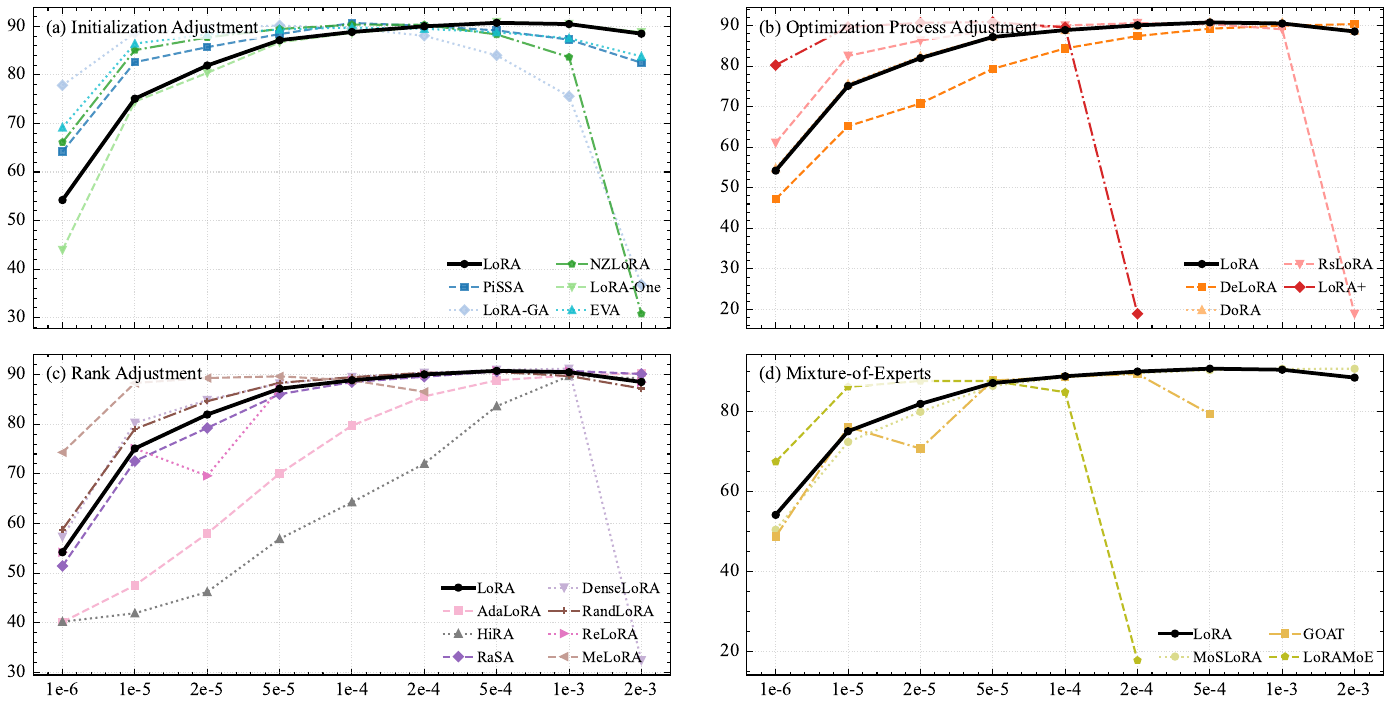}
\end{center}
\caption{Performance comparison of LoRA variants on seven image classification tasks across different learning rates. All plots share the same y-axis (accuracy) and x-axis (learning rate).}
\label{fig: cv_result}
\end{figure*}

\textbf{Experimental results}
Figure~\ref{fig: cv_result} shows the average evaluation results on the 7 subsets of the IC tasks, with \textsc{LoRA} and its variants exhibiting a gradual performance improvement as the learning rate increases, consistent with the trends observed in our NLU and NLG experiments; for the results of each subset, please refer to Table~\ref{tab:IC 1e-6}-~\ref{tab:IC 2e-3}. For \textsc{LoRA}, average performance rises steadily from 54.22 at a learning rate of 1e-6 to 90.75 at 5e-4. In contrast to our experiments on natural language understanding and generation, \textsc{LoRA} and its variants demonstrate greater robustness to variations in learning rates on the information classification tasks examined. Specifically, \textsc{LoRA} achieves a performance of 90.03 at a learning rate of 2e-4, with performance remaining approximately stable when the learning rate is slightly reduced to 1e-4 or increased to 5e-4. Most experimental results are similar to those of our experiments on NLU and NLG: variants with enhanced gradient flow (including \textsc{LoRA+}, which sets the learning rate of matrix $B$ 16 times the base learning rate) show advantages over vanilla \textsc{LoRA} at small learning rates, and the advantages diminish gradually as \textsc{LoRA} approaches its peak performance; \textsc{AdaLoRA}, \textsc{HiRA}, and \textsc{DeLoRA} show clear disadvantages over other methods at small learning rates for their constraints on the optimization process.

\subsubsection{Additional Experimental Settings}
For \textsc{LoRA} variants evaluated in our experiments, we adopt the following variant-specific hyperparameters (we adopt the recommended settings if possible):
\begin{itemize}
    \item \textsc{LoRA-GA}: The number of gradient estimation steps is set to 64; the hyperparameter \(\gamma\) is set to 16. 
    \item \textsc{NZLoRA}: Both hyperparameters \(\gamma_A\) and \(\gamma_B\) are set to 16.
    \item  \textsc{PiSSA}: The number of fast SVD decomposition iterations \textsc{PiSSA} is set to 64.
    \item \textsc{LoRA-One}: The number of gradient estimation steps is set to 64; the hyperparameter \(\gamma\) is set to 128.
    \item \textsc{EVA}: The number of the activation estimation steps is set to 64; the convergence threshold of the incremental SVD is set to 0.9.
    \item \textsc{DeLoRA}: The hyperparamter \(\lambda\) is set to 8. The initialization schemes for \(A\) and \(B\) are both the Kaiming uniform distribution.
    \item \textsc{LoRA+}: The ratio of learning rates used for \(B\) and \(A\) is set to 16.
    \item \textsc{MELoRA}: The number of diagonal blocks is set to 2, resulting in a overall rank of \(16\) for each adapter.
    \item \textsc{ReLoRA}: The low-rank weights are merged and reinitialized 3-5 times during each training process; the number of re-warmup steps after each \emph{merge-and-reinit} process is set to 10.
    \item \textsc{AdaLoRA}: The hyperparamter \(t_i\) is set to 100 while \(t_f\) is set to 900. The initial rank is set to 12, and the final effective rank is set to 8.
    \item \textsc{RandLoRA}: The rank of low-rank bases is set to 32, ensuring a comparable trainable parameter count to \textsc{LoRA}; the upper bound of the dimension of all low-rank bases is set to 1024, resulting in a maximum rank of 1024.
    \item \textsc{RaSA}: The shared rank \(k\) is set to 1, resulting in an overall rank of \(r-1 + L\) for each adapter.
    \item \textsc{DenseLoRA}: The rank of each adapter is set to \(24 \cdot r = 192\), where \(r\) is the rank used for vanilla \textsc{LoRA}. The hyperparameter \(\alpha\) is therefore set to \(48 \cdot r = 384\). This setting results in a comparable trainable parameter count to \textsc{LoRA} but with a much higher overall rank.
    \item \textsc{LoRAMoE} and \textsc{GOAT}: The number of experts is set to \(8\), each expert with a rank of \(1\), resulting in a total rank of \(8\); the number of activated experts for each token is set to \(2\).
    \item \textsc{HiRA}: No extra hyperparameters.
    \item \textsc{MoSLoRA}: No extra hyperparameters.
    \item \textsc{RsLoRA}: No extra hyperparameters.
    \item \textsc{DoRA}: No extra hyperparameters.
    
\end{itemize}

\subsubsection{Learning Rate Sweep Results on High-rank Settings}
To systematically investigate the high-rank performance of \textsc{LoRA} and its variants, we select one representative variant from each major category of LoRA extensions and evaluate their behavior under a high-rank setting. Specifically, we benchmark four variants: \textsc{LoRA-GA}, \textsc{RsLoRA}, \textsc{MELoRA}, and \textsc{LoRAMoE}, against vanilla \textsc{LoRA}. All experimental configurations are kept identical to those used in our main NLG experiments, except that the LoRA rank is uniformly set to 128 across all methods. As noted in Section \textrm{II}, it is common practice to set the scaling hyperparameter \(\alpha\) to twice the \textsc{LoRA} rank. To isolate and assess the impact of this convention, particularly in contrast to variants like \textsc{RsLoRA}, which inherently adjust scaling, we also evaluate vanilla \textsc{LoRA} with \(\alpha=256\) (i.e., 2 × 128) as a controlled baseline.

The experimental results under the high-rank setting are summarized in Table~\ref{tab: high-rank-12col}. On GSM8K, Vanilla \textsc{LoRA} achieves the best performance among all evaluated methods, reaching $76.57$ (an improvement of $+1.07$ over \textsc{LoRA} with $r=8$) with a learning rate of 1e-4 and $\alpha = 256$. Notably, \textsc{LoRA} with $\alpha = 16$ also attains a competitive accuracy of $76.12$ on GSM8K when using a relatively high learning rate of 5e-4. This suggests that, in our experimental setup, a larger learning rate can partially compensate for the effect of a small fixed $\alpha$. These trends differ from those reported in Figure S3 of Biderman et al.~\cite{biderman2024lora-learn-less}, which may be due to differences in learning-rate tuning or task selection across studies. Furthermore, the empirical findings from Kalajdzievski et al.~\cite{kalajdzievski2023rslora} indicate that the gradient norm of \textsc{LoRA} tends to collapse as the rank increases. Our results demonstrate that this issue can be mitigated through several strategies: applying the rank-stabilizing scaling of \textsc{RsLoRA}; setting $\alpha = 2r$ as empirical studies suggested; adopting \textsc{LoRA} variants with enhanced gradient flow such as \textsc{LoRA-GA}, or using a larger learning rate. On HumanEval, Vanilla \textsc{LoRA} ($\alpha=256$) and \textsc{LoRA-GA} both achieve peak performance of $51.30$, which is $3.28$ points higher than the peak performance of \textsc{LoRA} with $r=8$, at learning rates of 1e-4 and 5e-4, respectively. The substantial improvements on both GSM8k and HumanEval highlight the effectiveness of increasing \textsc{LoRA}'s rank when training settings are appropriately configured.

\begin{table*}[htbp]
\centering
\small
\caption{Performance of \textsc{LoRA} and its variants under a high-rank setting across learning rates on GSM8K and HumanEval. We use $\dagger$ and $\ddagger$ to denote LoRA with $\alpha=256$ and $\alpha=16$, respectively.}
\label{tab: high-rank-12col}
\resizebox{\textwidth}{!}{%
\begin{tabular}{l *{12}{>{\centering\arraybackslash}p{1.45cm}}}
\toprule
\textbf{Method} &
\multicolumn{6}{c}{\textbf{GSM8K}} &
\multicolumn{6}{c}{\textbf{HumanEval}} \\
\cmidrule(lr){2-7} \cmidrule(lr){8-13}
& \textbf{1e-5} & \textbf{2e-5} & \textbf{5e-5} & \textbf{1e-4} & \textbf{2e-4} & \textbf{5e-4}
& \textbf{1e-5} & \textbf{2e-5} & \textbf{5e-5} & \textbf{1e-4} & \textbf{2e-4} & \textbf{5e-4} \\
\midrule
LoRA$^\dagger$  & 72.02 & 72.40 & 74.98 & 76.57 & 74.30 & 1.29
                & 46.88 & 46.80 & 48.70 & 51.30 & 51.07 & 41.31 \\
LoRA$^\ddagger$ & 61.87 & 66.49 & 72.86 & 71.65 & 73.01 & 76.12
                & 40.02 & 41.84 & 43.90 & 46.11 & 47.09 & 48.63 \\
LoRA-GA         & 70.66 & 72.48 & 75.74 & 75.82 & 74.60 & 71.42
                & 41.16 & 46.49 & 51.30 & 50.99 & 51.07 & 48.48 \\
RsLoRA          & 68.34 & 71.72 & 74.91 & 75.66 & 74.75 & 1.52
                & 44.51 & 45.66 & 50.00 & 51.14 & 51.60 & 46.27 \\
MELoRA          & 71.42 & 72.02 & 74.60 & 72.63 & 69.90 & 57.32
                & 48.09 & 48.78 & 50.53 & 48.40 & 43.14 & 32.70 \\
\bottomrule
\end{tabular}}
\end{table*}

\begin{table*}[t]
\centering
\setlength{\tabcolsep}{6pt}
\renewcommand{\arraystretch}{1.08}
\caption{Performance comparison on the GLUE benchmark with a learning rate of 1e-6.}
\label{tab:glue 1e-6}

\begin{tabular*}{0.9\textwidth}{@{\extracolsep{\fill}}lcccccccccc@{}}
\toprule
Method & SST-2 & CoLA & MNLI & MRPC & QNLI & QQP & RTE & STS-B & WNLI & AVG \\
\midrule
LoRA
& 89.45 & 0.00 & 78.69 & 66.38 & 82.70 & 79.43 & 47.65 & 13.37 & 39.44 & 55.23 \\
\midrule
\rowcolor{gray!10}
\multicolumn{11}{c}{\textbf{Rank Adjustment Based LoRA Variants}} \\
HiRA
& 49.08 & 0.96 & 32.74 & 65.83 & 49.13 & 0.00 & 47.65 & 2.71 & 39.44 & 31.95 \\
RaSA
& 88.76 & 0.00 & 76.95 & 66.38 & 80.59 & 78.64 & 48.38 & 6.20 & 40.85 & 54.08 \\
AdaLoRA
& 49.08 & 0.00 & 32.77 & 65.71 & 48.92 & 0.00 & 47.65 & 2.64 & 39.44 & 31.66 \\
MELoRA
& 92.55 & 0.00 & 84.10 & 66.38 & 88.82 & 83.33 & 50.54 & 78.53 & 39.44 & 64.85 \\
DenseLoRA
& 91.51 & 0.00 & 81.29 & 66.38 & 86.02 & 81.30 & 48.74 & 0.00 & 36.62 & 54.58 \\
RandLoRA
& 84.29 & 0.00 & 71.93 & 66.38 & 80.08 & 78.10 & 47.65 & 11.85 & 39.44 & 53.30 \\
ReLoRA & 89.22 & 0.00 & 77.47 & 66.38 & 81.85 & 82.31 & 48.38 & 15.48 & 40.85 & 55.77 \\
\midrule
\rowcolor{gray!10}
\multicolumn{11}{c}{\textbf{Optimization Process Adjustment Based LoRA Variants}} \\
DoRA
& 89.56 & 0.00 & 78.70 & 66.38 & 82.93 & 79.48 & 47.65 & 13.88 & 39.44 & 55.34 \\
RsLoRA
& 91.97 & 0.00 & 81.83 & 66.38 & 86.14 & 81.24 & 49.10 & 6.17 & 38.03 & 55.65 \\
LoRA+
& 93.23 & 0.00 & 84.64 & 66.38 & 89.76 & 84.04 & 49.10 & 81.13 & 49.30 & 66.40 \\
DeLoRA
& 50.92 & 0.00 & 66.85 & 66.38 & 63.18 & 77.35 & 48.74 & 9.24 & 40.85 & 47.05 \\
\midrule
\rowcolor{gray!10}
\multicolumn{11}{c}{\textbf{Initialization Adjustment Based LoRA Variants}} \\
EVA
& 91.63 & 0.00 & 81.57 & 66.38 & 86.75 & 80.95 & 49.46 & 20.34 & 38.03 & 57.23 \\
PiSSA
& 90.94 & 0.00 & 82.01 & 66.38 & 85.68 & 81.48 & 49.10 & 18.05 & 38.03 & 56.85 \\
LoRAGA
& 92.55 & 0.00 & 84.51 & 66.38 & 89.97 & 83.61 & 53.79 & 80.34 & 38.03 & 65.46 \\
LoRA-One
& 89.11 & 0.00 & 78.18 & 66.38 & 83.29 & 78.60 & 47.29 & 9.15 & 56.34 & 56.48 \\
NZLoRA
& 90.94 & 0.00 & 82.04 & 66.38 & 86.56 & 81.27 & 49.46 & 9.38 & 38.03 & 56.01 \\
\midrule
\rowcolor{gray!10}
\multicolumn{11}{c}{\textbf{Mixture-of-Experts Integration Based Variants}} \\
GOAT
& 87.16 & 0.00 & 76.17 & 66.38 & 69.73 & 78.49 & 52.35 & 5.34 & 49.30 & 53.88 \\
MoSLoRA
& 88.42 & 0.00 & 77.32 & 66.38 & 81.48 & 78.62 & 48.38 & 6.26 & 40.85 & 54.19 \\
LoRAMoE
& 91.86 & 0.00 & 82.82 & 66.38 & 87.20 & 82.02 & 49.46 & 22.07 & 38.03 & 57.76 \\
\bottomrule
\end{tabular*}
\end{table*}

\begin{table*}[t]
\centering
\setlength{\tabcolsep}{6pt}
\renewcommand{\arraystretch}{1.08}
\caption{Performance comparison on the GLUE benchmark with a learning rate of 1e-5.}
\label{tab:glue 1e-5}

\begin{tabular*}{0.9\textwidth}{@{\extracolsep{\fill}}lcccccccccc@{}}
\toprule
Method & SST-2 & CoLA & MNLI & MRPC & QNLI & QQP & RTE & STS-B & WNLI & AVG \\
\midrule
LoRA
& 93.35 & 32.91 & 85.59 & 66.38 & 90.47 & 84.91 & 49.10 & 82.88 & 42.25 & 69.76 \\
\midrule
\rowcolor{gray!10}
\multicolumn{11}{c}{\textbf{Rank Adjustment Based LoRA Variants}} \\
HiRA
& 86.12 & 0.00 & 76.76 & 66.26 & 79.62 & 78.65 & 47.65 & 4.22 & 39.44 & 53.19 \\
RaSA
& 93.23 & 0.00 & 85.28 & 66.38 & 90.24 & 84.48 & 49.82 & 81.12 & 40.85 & 65.71 \\
AdaLoRA
& 90.37 & 0.00 & 82.29 & 66.38 & 85.97 & 81.31 & 47.65 & 5.69 & 39.44 & 55.46 \\
MELoRA
& 94.38 & 53.18 & 86.70 & 84.08 & 91.97 & 86.57 & 68.95 & 88.87 & 39.44 & 77.13 \\
DenseLoRA
& 94.04 & 50.99 & 86.15 & 66.38 & 91.70 & 85.59 & 49.46 & 86.41 & 50.70 & 73.49 \\
RandLoRA
& 92.66 & 0.00 & 84.88 & 66.38 & 89.12 & 84.03 & 51.99 & 80.68 & 39.44 & 65.46 \\
ReLoRA
& 93.46 & 52.74 & 85.09 & 66.38 & 89.89 & 87.23 & 49.46 & 83.39 & 50.70 & 73.15 \\
\midrule
\rowcolor{gray!10}
\multicolumn{11}{c}{\textbf{Optimization Process Adjustment Based LoRA Variants}} \\
DoRA
& 93.23 & 32.03 & 85.73 & 66.38 & 90.60 & 84.90 & 49.10 & 82.89 & 43.66 & 69.84 \\
RsLoRA
& 94.27 & 48.15 & 86.51 & 68.58 & 91.45 & 85.96 & 54.15 & 86.16 & 47.89 & 73.68 \\
LoRA+
& 94.38 & 56.06 & 87.31 & 84.81 & 92.44 & 87.58 & 71.12 & 89.53 & 42.25 & 78.39 \\
DeLoRA
& 91.86 & 0.00 & 84.66 & 66.38 & 89.08 & 83.22 & 49.10 & 76.11 & 40.85 & 64.58 \\
\midrule
\rowcolor{gray!10}
\multicolumn{11}{c}{\textbf{Initialization Adjustment Based LoRA Variants}} \\
EVA
& 94.38 & 51.11 & 86.06 & 75.96 & 91.83 & 85.57 & 61.01 & 86.38 & 43.66 & 75.11 \\
PiSSA
& 93.12 & 50.70 & 86.42 & 73.03 & 91.51 & 86.23 & 58.48 & 86.68 & 47.89 & 74.90 \\
LoRAGA
& 94.61 & 54.87 & 87.09 & 85.72 & 92.44 & 86.97 & 69.68 & 88.30 & 36.62 & 77.37 \\
LoRA-One
& 93.23 & 35.74 & 85.79 & 66.38 & 90.68 & 84.91 & 48.01 & 82.91 & 56.34 & 71.55 \\
NZLoRA
& 93.58 & 50.87 & 86.42 & 67.85 & 91.55 & 85.76 & 60.29 & 85.52 & 45.07 & 74.10 \\
\midrule
\rowcolor{gray!10}
\multicolumn{11}{c}{\textbf{Mixture-of-Experts Integration Based Variants}} \\
GOAT
& 93.12 & 0.00 & 85.64 & 66.38 & 90.41 & 85.33 & 51.26 & 81.28 & 49.30 & 66.97 \\
MoSLoRA
& 93.35 & 18.30 & 85.58 & 66.38 & 90.62 & 84.73 & 49.10 & 81.44 & 38.03 & 67.50 \\
LoRAMoE
& 93.35 & 48.07 & 85.63 & 69.49 & 92.02 & 86.06 & 66.06 & 88.75 & 36.62 & 74.01 \\
\bottomrule
\end{tabular*}
\end{table*}

\begin{table*}[t]
\centering
\setlength{\tabcolsep}{6pt}
\renewcommand{\arraystretch}{1.08}
\caption{Performance comparison on the GLUE benchmark with a learning rate of 5e-5.}
\label{tab:glue 5e-5}

\begin{tabular*}{0.9\textwidth}{@{\extracolsep{\fill}}lcccccccccc@{}}
\toprule
Method & SST-2 & CoLA & MNLI & MRPC & QNLI & QQP & RTE & STS-B & WNLI & AVG \\
\midrule
LoRA
& 94.38 & 55.25 & 87.67 & 85.60 & 92.35 & 87.37 & 67.51 & 89.37 & 49.30 & 78.76 \\
\midrule
\rowcolor{gray!10}
\multicolumn{11}{c}{\textbf{Rank Adjustment Based LoRA Variants}} \\
HiRA
& 92.09 & 0.00 & 84.36 & 66.38 & 88.93 & 83.58 & 48.38 & 47.79 & 40.85 & 61.37 \\
RaSA
& 93.92 & 54.32 & 87.76 & 84.38 & 92.44 & 87.15 & 66.79 & 89.48 & 46.48 & 78.08 \\
AdaLoRA
& 93.23 & 40.03 & 86.11 & 66.38 & 90.45 & 85.12 & 47.65 & 82.10 & 39.44 & 70.06 \\
MELoRA
& 93.46 & 56.25 & 85.89 & 84.93 & 91.19 & 86.67 & 75.81 & 90.57 & 29.58 & 77.15 \\
DenseLoRA
& 94.38 & 58.05 & 87.81 & 86.70 & 92.61 & 87.88 & 69.68 & 90.30 & 49.30 & 79.63 \\
RandLoRA
& 92.89 & 48.94 & 87.09 & 81.70 & 92.27 & 86.82 & 62.45 & 86.94 & 35.21 & 74.92 \\
ReLoRA
& 94.61 & 55.08 & 87.07 & 85.17 & 92.06 & 89.48 & 70.04 & 88.62 & 56.34 & 79.83 \\
\midrule
\rowcolor{gray!10}
\multicolumn{11}{c}{\textbf{Optimization Process Adjustment Based LoRA Variants}} \\
DoRA
& 94.38 & 52.44 & 87.73 & 85.36 & 92.39 & 87.37 & 68.23 & 89.42 & 49.30 & 78.51 \\
RsLoRA
& 94.61 & 54.98 & 87.67 & 85.72 & 92.73 & 88.07 & 72.92 & 90.39 & 46.48 & 79.29 \\
LoRA+
& 94.95 & 60.57 & 87.28 & 86.52 & 92.39 & 87.89 & 74.73 & 90.82 & 25.35 & 77.83 \\
DeLoRA
& 93.92 & 52.37 & 86.57 & 83.53 & 91.59 & 86.14 & 59.57 & 84.65 & 45.07 & 75.93 \\
\midrule
\rowcolor{gray!10}
\multicolumn{11}{c}{\textbf{Initialization Adjustment Based LoRA Variants}} \\
EVA
& 94.04 & 56.02 & 87.42 & 85.23 & 92.42 & 87.46 & 72.92 & 90.53 & 33.80 & 77.76 \\
PiSSA
& 93.92 & 57.29 & 87.38 & 86.21 & 92.08 & 87.75 & 70.40 & 89.73 & 39.44 & 78.24 \\
LoRAGA
& 93.35 & 59.07 & 87.34 & 86.33 & 92.73 & 87.27 & 76.17 & 90.29 & 33.80 & 78.48 \\
LoRA-One
& 94.15 & 55.79 & 87.77 & 84.62 & 92.71 & 87.35 & 70.76 & 89.28 & 59.15 & 80.18 \\
NZLoRA
& 94.27 & 56.58 & 87.07 & 85.85 & 92.40 & 87.74 & 74.01 & 89.95 & 30.99 & 77.65 \\
\midrule
\rowcolor{gray!10}
\multicolumn{11}{c}{\textbf{Mixture-of-Experts Integration Based Variants}} \\
GOAT
& 94.04 & 55.23 & 86.83 & 83.71 & 92.46 & 87.43 & 56.68 & 88.91 & 53.52 & 77.65 \\
MoSLoRA
& 94.27 & 53.82 & 87.47 & 85.17 & 92.20 & 87.24 & 70.04 & 89.12 & 46.48 & 78.42 \\
LoRAMoE
& 93.00 & 53.91 & 85.55 & 85.72 & 91.72 & 86.02 & 70.76 & 90.37 & 25.35 & 75.82 \\
\bottomrule
\end{tabular*}
\end{table*}

\begin{table*}[t]
\centering
\setlength{\tabcolsep}{6pt}
\renewcommand{\arraystretch}{1.08}
\caption{Performance comparison on the GLUE benchmark with a learning rate of 1e-4.}
\label{tab:glue 1e-4}

\begin{tabular*}{0.9\textwidth}{@{\extracolsep{\fill}}lcccccccccc@{}}
\toprule
Method & SST-2 & CoLA & MNLI & MRPC & QNLI & QQP & RTE & STS-B & WNLI & AVG \\
\midrule
LoRA
& 94.27 & 55.99 & 87.70 & 86.58 & 92.75 & 88.17 & 73.29 & 90.29 & 47.89 & 79.66 \\
\midrule
\rowcolor{gray!10}
\multicolumn{11}{c}{\textbf{Rank Adjustment Based LoRA Variants}} \\
HiRA
& 93.23 & 0.00 & 85.72 & 66.38 & 90.64 & 85.14 & 49.82 & 80.68 & 39.44 & 65.67 \\
RaSA
& 93.81 & 57.27 & 87.74 & 85.60 & 92.42 & 87.83 & 73.29 & 90.11 & 49.30 & 79.71 \\
AdaLoRA
& 93.81 & 49.69 & 87.27 & 66.38 & 92.14 & 86.68 & 49.82 & 86.94 & 39.44 & 72.46 \\
MELoRA
& 92.43 & 53.38 & 85.11 & 84.38 & 91.13 & 85.22 & 69.68 & 90.24 & 56.34 & 78.66 \\
DenseLoRA
& 93.81 & 59.57 & 87.47 & 87.00 & 92.78 & 88.64 & 75.45 & 90.59 & 47.89 & 80.36 \\
RandLoRA
& 93.81 & 51.71 & 87.35 & 85.23 & 92.46 & 87.70 & 68.95 & 88.83 & 35.21 & 76.81 \\
ReLoRA
& 94.72 & 58.35 & 87.67 & 87.19 & 92.69 & 87.50 & 71.12 & 89.88 & 56.34 & 80.61 \\
\midrule
\rowcolor{gray!10}
\multicolumn{11}{c}{\textbf{Optimization Process Adjustment Based LoRA Variants}} \\
DoRA
& 94.15 & 56.29 & 87.63 & 86.39 & 92.58 & 88.13 & 72.56 & 90.33 & 45.07 & 79.24 \\
RsLoRA
& 94.27 & 57.78 & 87.25 & 85.30 & 92.46 & 88.27 & 73.65 & 91.04 & 38.03 & 78.67 \\
LoRA+
& 93.58 & 61.82 & 86.56 & 86.58 & 92.40 & 86.01 & 52.71 & 90.87 & 40.85 & 76.82 \\
DeLoRA
& 94.38 & 52.42 & 87.11 & 85.30 & 92.29 & 86.95 & 66.79 & 87.13 & 43.66 & 77.34 \\
\midrule
\rowcolor{gray!10}
\multicolumn{11}{c}{\textbf{Initialization Adjustment Based LoRA Variants}} \\
EVA
& 94.15 & 58.81 & 87.51 & 85.48 & 92.29 & 88.18 & 75.09 & 90.75 & 25.35 & 77.51 \\
PiSSA
& 93.92 & 59.81 & 87.24 & 86.70 & 92.37 & 88.10 & 76.90 & 90.27 & 28.17 & 78.16 \\
LoRAGA
& 92.55 & 57.54 & 86.16 & 85.66 & 91.70 & 86.77 & 72.56 & 90.25 & 56.34 & 79.95 \\
LoRA-One
& 94.38 & 55.51 & 87.89 & 85.42 & 92.86 & 87.91 & 74.37 & 90.17 & 52.11 & 80.07 \\
NZLoRA
& 94.04 & 59.05 & 87.27 & 86.15 & 91.93 & 87.77 & 72.20 & 90.20 & 25.35 & 77.11 \\
\midrule
\rowcolor{gray!10}
\multicolumn{11}{c}{\textbf{Mixture-of-Experts Integration Based Variants}} \\
GOAT
& 94.04 & 53.64 & 86.99 & 85.72 & 92.10 & 87.25 & 69.31 & 89.56 & 47.89 & 78.50 \\
MoSLoRA
& 94.95 & 55.49 & 87.62 & 86.82 & 92.67 & 87.88 & 72.56 & 90.22 & 47.89 & 79.57 \\
LoRAMoE
& 91.97 & 56.25 & 32.74 & 84.26 & 90.24 & 0.00 & 71.84 & 90.39 & 23.94 & 60.18 \\
\bottomrule
\end{tabular*}
\end{table*}

\begin{table*}[t]
\centering
\setlength{\tabcolsep}{6pt}
\renewcommand{\arraystretch}{1.08}
\caption{Performance comparison on the GLUE benchmark with a learning rate of 5e-4.}
\label{tab:glue 5e-4}

\begin{tabular*}{0.9\textwidth}{@{\extracolsep{\fill}}lcccccccccc@{}}
\toprule
Method & SST-2 & CoLA & MNLI & MRPC & QNLI & QQP & RTE & STS-B & WNLI & AVG \\
\midrule
LoRA
& 93.23 & 61.58 & 87.12 & 86.15 & 92.23 & 0.00 & 78.34 & 91.00 & 28.17 & 68.65 \\
\midrule
\rowcolor{gray!10}
\multicolumn{11}{c}{\textbf{Rank Adjustment Based LoRA Variants}} \\
HiRA
& 94.04 & 55.47 & 87.15 & 84.81 & 92.58 & 87.79 & 64.98 & 87.77 & 40.85 & 77.27 \\
RaSA
& 94.04 & 60.32 & 87.11 & 86.94 & 92.42 & 88.50 & 75.81 & 90.88 & 56.34 & 81.37 \\
AdaLoRA
& 93.35 & 58.06 & 87.25 & 87.19 & 92.76 & 88.26 & 70.76 & 90.31 & 36.62 & 78.29 \\
MELoRA
& 50.92 & 0.00 & 32.74 & 66.38 & 50.63 & 0.00 & 52.71 & 0.00 & 56.34 & 34.26 \\
DenseLoRA
& 93.35 & 60.82 & 31.82 & 87.37 & 92.08 & 0.00 & 52.71 & 91.01 & 28.17 & 59.70 \\
RandLoRA
& 94.15 & 57.53 & 86.82 & 87.00 & 92.35 & 88.15 & 75.81 & 90.53 & 22.54 & 77.21 \\
ReLoRA 
& 93.58 & 55.48 & 84.49 & 86.09 & 91.83 & 86.83 & 52.71 & 91.05 & 56.34 & 77.60 \\
\midrule
\rowcolor{gray!10}
\multicolumn{11}{c}{\textbf{Optimization Process Adjustment Based LoRA Variants}} \\
DoRA
& 93.58 & 62.35 & 86.56 & 87.25 & 92.42 & 85.52 & 76.53 & 90.94 & 28.17 & 78.15 \\
RsLoRA
& 50.92 & 60.58 & 32.74 & 84.81 & 50.63 & 0.00 & 72.92 & 90.85 & 56.34 & 55.53 \\
LoRA+
& 50.92 & 0.00 & 32.74 & 66.38 & 49.37 & 0.00 & 52.71 & 10.57 & 56.34 & 35.45 \\
DeLoRA
& 94.27 & 59.05 & 87.50 & 86.64 & 92.42 & 87.97 & 72.20 & 89.81 & 28.17 & 77.56 \\
\midrule
\rowcolor{gray!10}
\multicolumn{11}{c}{\textbf{Initialization Adjustment Based LoRA Variants}} \\
EVA
& 92.66 & 59.05 & 86.81 & 84.81 & 91.63 & 0.00 & 52.71 & 90.72 & 56.34 & 68.30 \\
PiSSA
& 50.92 & 0.00 & 32.74 & 83.89 & 50.63 & 0.00 & 52.71 & 89.49 & 56.34 & 46.30 \\
LoRAGA
& 50.92 & 0.00 & 31.82 & 66.38 & 49.37 & 0.00 & 47.29 & 5.15 & 56.34 & 34.14 \\
LoRA-One
& 93.69 & 62.07 & 86.95 & 87.37 & 92.01 & 0.00 & 78.34 & 90.65 & 29.58 & 68.96 \\
NZLoRA
& 50.92 & 0.00 & 32.74 & 85.42 & 50.63 & 0.00 & 52.71 & 89.66 & 56.34 & 46.49 \\
\midrule
\rowcolor{gray!10}
\multicolumn{11}{c}{\textbf{Mixture-of-Experts Integration Based Variants}} \\
GOAT
& 50.92 & 0.00 & 32.74 & 85.17 & 49.37 & 0.00 & 52.71 & 89.29 & 30.99 & 43.47 \\
MoSLoRA
& 93.23 & 59.56 & 87.00 & 85.36 & 92.12 & 88.29 & 76.53 & 90.72 & 35.21 & 78.67 \\
LoRAMoE
& 50.92 & 0.00 & 32.74 & 66.38 & 49.37 & 0.00 & 52.71 & 1.29 & 43.66 & 33.01 \\
\bottomrule
\end{tabular*}
\end{table*}

\begin{table*}[t]
\centering
\setlength{\tabcolsep}{6pt}
\renewcommand{\arraystretch}{1.08}
\caption{Performance comparison on the GLUE benchmark with a learning rate of 1e-3.}
\label{tab:glue 1e-3}

\begin{tabular*}{0.9\textwidth}{@{\extracolsep{\fill}}lcccccccccc@{}}
\toprule
Method & SST-2 & CoLA & MNLI & MRPC & QNLI & QQP & RTE & STS-B & WNLI & AVG \\
\midrule
LoRA
& 50.92 & 0.00 & 32.74 & 86.33 & 49.37 & 0.00 & 52.71 & 89.35 & 30.99 & 43.60 \\
\midrule
\rowcolor{gray!10}
\multicolumn{11}{c}{\textbf{Rank Adjustment Based LoRA Variants}} \\
HiRA
& 93.69 & 57.27 & 86.89 & 86.94 & 92.23 & 88.52 & 72.92 & 89.76 & 47.89 & 79.57 \\
RaSA
& 93.00 & 0.00 & 87.27 & 86.46 & 92.25 & 0.00 & 77.98 & 90.90 & 28.17 & 61.78 \\
AdaLoRA
& 93.81 & 61.57 & 87.33 & 86.76 & 92.12 & 88.27 & 76.53 & 90.53 & 40.85 & 79.75 \\
MELoRA
& 50.92 & 0.00 & 31.82 & 66.38 & 49.37 & 0.00 & 52.71 & 0.00 & 43.66 & 32.55 \\
DenseLoRA
& 50.92 & 0.00 & 32.74 & 66.38 & 50.63 & 0.00 & 52.71 & 0.00 & 56.34 & 34.18 \\
RandLoRA
& 93.58 & 56.50 & 86.59 & 87.37 & 91.19 & 87.13 & 77.98 & 90.65 & 25.35 & 77.37 \\
ReLoRA 
& 50.92 & 15.54 & 35.45 & 66.38 & 50.63 & 38.72 & 52.71 & 88.51 & 56.34 & 50.58 \\
\midrule
\rowcolor{gray!10}
\multicolumn{11}{c}{\textbf{Optimization Process Adjustment Based LoRA Variants}} \\
DoRA
& 50.92 & 58.83 & 32.74 & 66.38 & 50.63 & 0.00 & 52.71 & 89.39 & 32.39 & 48.22 \\
RsLoRA
& 50.92 & 0.00 & 35.45 & 66.38 & 49.37 & 0.00 & 52.71 & 89.94 & 56.34 & 44.57 \\
LoRA+
& 50.92 & 0.00 & 32.74 & 66.38 & 49.37 & 0.00 & 52.71 & 13.26 & 56.34 & 35.75 \\
DeLoRA
& 93.81 & 61.09 & 87.72 & 86.58 & 92.50 & 88.23 & 68.95 & 89.74 & 30.99 & 77.73 \\
\midrule
\rowcolor{gray!10}
\multicolumn{11}{c}{\textbf{Initialization Adjustment Based LoRA Variants}} \\
EVA
& 50.92 & 0.00 & 35.45 & 66.38 & 49.37 & 0.00 & 52.71 & 89.70 & 56.34 & 44.54 \\
PiSSA
& 50.92 & 0.00 & 32.74 & 66.38 & 50.63 & 0.00 & 52.71 & 7.96 & 56.34 & 35.30 \\
LoRAGA
& 50.92 & 0.00 & 31.82 & 66.38 & 49.37 & 0.00 & 47.29 & 0.00 & 56.34 & 33.20 \\
LoRA-One
& 50.92 & 0.00 & 32.74 & 85.91 & 49.37 & 0.00 & 52.71 & 90.79 & 38.03 & 44.50 \\
NZLoRA
& 50.92 & 0.00 & 32.74 & 66.38 & 50.63 & 0.00 & 47.29 & 88.46 & 56.34 & 43.64 \\
\midrule
\rowcolor{gray!10}
\multicolumn{11}{c}{\textbf{Mixture-of-Experts Integration Based Variants}} \\
GOAT
& 50.92 & 0.00 & 32.74 & 66.38 & 49.37 & 0.00 & 52.71 & 2.65 & 56.34 & 34.57 \\
MoSLoRA
& 94.04 & 60.07 & 87.32 & 86.39 & 92.33 & 87.76 & 52.71 & 89.55 & 53.52 & 78.19 \\
LoRAMoE
& 50.92 & 0.00 & 32.74 & 66.38 & 49.37 & 0.00 & 52.71 & 0.40 & 43.66 & 32.91 \\
\bottomrule
\end{tabular*}
\end{table*}

\begin{table*}[t]
\centering
\setlength{\tabcolsep}{6pt}
\renewcommand{\arraystretch}{1.08}
\caption{Performance comparison on the GLUE benchmark with a learning rate of 5e-3.}
\label{tab:glue 5e-3}

\begin{tabular*}{0.9\textwidth}{@{\extracolsep{\fill}}lcccccccccc@{}}
\toprule
Method & SST-2 & CoLA & MNLI & MRPC & QNLI & QQP & RTE & STS-B & WNLI & AVG \\
\midrule
LoRA
& 50.92 & 0.00 & 32.74 & 66.38 & 49.37 & 0.00 & 52.71 & 0.00 & 56.34 & 34.18 \\
\midrule
\rowcolor{gray!10}
\multicolumn{11}{c}{\textbf{Rank Adjustment Based LoRA Variants}} \\
HiRA
& 90.37 & 59.12 & 32.74 & 86.82 & 49.37 & 0.00 & 79.78 & 90.69 & 23.94 & 56.98 \\
RaSA
& 50.92 & 0.00 & 32.74 & 66.38 & 50.63 & 0.00 & 52.71 & 0.00 & 43.66 & 32.73 \\
AdaLoRA
& 50.92 & 60.08 & 32.74 & 86.03 & 50.63 & 0.00 & 52.71 & 0.00 & 54.93 & 42.99 \\
MELoRA
& 50.92 & 0.00 & 32.74 & 66.38 & 49.37 & 0.00 & 52.71 & 0.10 & 56.34 & 34.28 \\
DenseLoRA
& 50.92 & 0.00 & 31.82 & 66.38 & 49.37 & 0.00 & 52.71 & 0.23 & 56.34 & 34.20 \\
RandLoRA
& 50.92 & 0.00 & 31.82 & 66.38 & 50.63 & 0.00 & 52.71 & 0.00 & 56.34 & 33.21 \\
ReLoRA
& 50.92 & 62.40 & 35.45 & 66.38 & 50.63 & 38.72 & 52.71 & 2.52 & 56.34 & 46.23 \\
\midrule
\rowcolor{gray!10}
\multicolumn{11}{c}{\textbf{Optimization Process Adjustment Based LoRA Variants}} \\
DoRA
& 50.92 & 0.00 & 32.74 & 66.38 & 49.37 & 0.00 & 52.71 & 0.00 & 43.66 & 32.82 \\
RsLoRA
& 50.92 & 0.00 & 32.74 & 66.38 & 49.37 & 0.00 & 52.71 & 0.34 & 43.66 & 32.90 \\
LoRA+
& 50.92 & 0.00 & 32.74 & 66.38 & 49.37 & 0.00 & 52.71 & 1.23 & 43.66 & 33.00 \\
DeLoRA
& 94.15 & 0.00 & 87.07 & 66.38 & 92.63 & 88.24 & 47.29 & 0.00 & 56.34 & 58.71 \\
\midrule
\rowcolor{gray!10}
\multicolumn{11}{c}{\textbf{Initialization Adjustment Based LoRA Variants}} \\
EVA
& 50.92 & 0.00 & 32.74 & 66.38 & 49.37 & 0.00 & 52.71 & 0.00 & 56.34 & 34.18 \\
PiSSA
& 50.92 & 0.00 & 32.74 & 66.38 & 49.37 & 0.00 & 52.71 & 0.00 & 43.66 & 32.62 \\
LoRAGA
& 50.92 & 0.00 & 32.74 & 66.38 & 49.37 & 0.00 & 52.71 & 2.97 & 43.66 & 33.19 \\
LoRA-One
& 50.92 & 0.00 & 35.45 & 66.38 & 49.37 & 0.00 & 52.71 & 2.89 & 43.66 & 33.49 \\
NZLoRA
& 50.92 & 0.00 & 32.74 & 66.38 & 49.37 & 0.00 & 52.71 & 0.98 & 43.66 & 32.97 \\
\midrule
\rowcolor{gray!10}
\multicolumn{11}{c}{\textbf{Mixture-of-Experts Integration Based Variants}} \\
GOAT
& 50.92 & 0.00 & 35.45 & 66.38 & 49.37 & 0.00 & 52.71 & 1.60 & 43.66 & 33.34 \\
MoSLoRA
& 50.92 & 0.00 & 32.74 & 66.38 & 49.37 & 0.00 & 52.71 & 0.00 & 56.34 & 34.27 \\
LoRAMoE
& 50.92 & 0.00 & 32.74 & 66.38 & 50.63 & 0.00 & 52.71 & 0.80 & 56.34 & 34.50 \\
\bottomrule
\end{tabular*}
\end{table*}

\begin{table*}[t]
\centering
\footnotesize
\setlength{\tabcolsep}{5pt}
\renewcommand{\arraystretch}{1.1}
\caption{Performance on the seven image classification tasks with a learning rate of 1e-6.}
\label{tab:IC 1e-6}

\begin{tabular*}{0.9\textwidth}{@{\extracolsep{\fill}}lcccccccc@{}}
\toprule
Method & Cars & DTD & EuroSAT & GTSRB & RESISIC45 & SUN397 & SVHN & AVG \\
\midrule
LoRA & 65.58 & 47.39 & 73.69 & 44.00 & 65.16 & 65.30 & 62.01 & 54.22 \\
\midrule
\rowcolor{gray!10}
\multicolumn{9}{c}{\textbf{Rank Adjustment Based LoRA Variants}} \\
HiRA & 63.96 & 42.87 & 40.74 & 69.00 & 56.71 & 62.76 & 13.91 & 40.23 \\
RaSA & 65.25 & 46.54 & 66.72 & 36.00 & 63.32 & 64.87 & 53.01 & 51.44 \\
AdaLoRA & 63.89 & 42.87 & 40.44 & 69.00 & 56.63 & 62.75 & 13.75 & 40.15 \\
MELoRA & 71.60 & 63.30 & 96.93 & 39.59 & 86.27 & 69.90 & 92.75 & 74.33 \\
DenseLoRA & 66.24 & 50.00 & 79.13 & 69.00 & 67.52 & 66.65 & 71.05 & 57.33 \\
RandLoRA & 66.70 & 49.95 & 84.31 & 1.18 & 69.03 & 65.79 & 74.47 & 58.78 \\
ReLoRA & 65.44 & 46.60 & 70.44 & 0.39 & 63.98 & 64.91 & 55.38 & 52.45 \\
\midrule
\rowcolor{gray!10}
\multicolumn{9}{c}{\textbf{Optimization Process Adjustment Based LoRA Variants}} \\
DoRA & 65.64 & 47.93 & 74.94 & 48.00 & 65.51 & 65.46 & 63.82 & 54.83 \\
RsLoRA & 67.35 & 52.13 & 87.39 & 3.06 & 70.87 & 67.30 & 79.54 & 61.09 \\
LoRA+ & 74.31 & 70.64 & 97.39 & 63.65 & 89.30 & 71.83 & 94.48 & 80.23 \\
DeLoRA & 64.92 & 44.79 & 59.07 & 37.00 & 60.94 & 63.73 & 36.37 & 47.17 \\
\midrule
\rowcolor{gray!10}
\multicolumn{9}{c}{\textbf{Initialization Adjustment Based LoRA Variants}} \\
EVA & 69.53 & 58.35 & 94.89 & 22.64 & 81.17 & 68.71 & 89.02 & 69.19 \\
PiSSA & 67.83 & 53.94 & 92.19 & 8.84 & 75.33 & 67.89 & 83.42 & 64.21 \\
LoRA-GA & 73.50 & 68.03 & 97.26 & 53.33 & 88.35 & 70.83 & 93.73 & 77.86 \\
LoRA-One & 18.70 & 46.65 & 61.65 & 2.47 & 51.97 & 62.66 & 63.39 & 43.93 \\
NZLoRA & 68.69 & 55.59 & 94.20 & 12.73 & 78.22 & 67.90 & 85.57 & 66.13 \\
\midrule
\rowcolor{gray!10}
\multicolumn{9}{c}{\textbf{Mixture-of-Experts Integration Based Variants}} \\
GOAT & 63.11 & 44.04 & 60.04 & 44.00 & 59.46 & 61.94 & 52.57 & 48.80 \\
MoSLoRA & 65.27 & 45.80 & 64.28 & 36.00 & 62.59 & 64.62 & 50.50 & 50.49 \\
LoRAMoE & 69.37 & 55.21 & 94.93 & 16.94 & 79.05 & 67.89 & 89.07 & 67.49 \\
\bottomrule
\end{tabular*}
\end{table*}

\begin{table*}[t]
\centering
\footnotesize
\setlength{\tabcolsep}{5pt}
\renewcommand{\arraystretch}{1.1}
\caption{Performance on the Seven image classification tasks with a learning rate of 1e-5.}
\label{tab:IC 1e-5}

\begin{tabular*}{0.9\textwidth}{@{\extracolsep{\fill}}lcccccccc@{}}
\toprule
Method & Cars & DTD & EuroSAT & GTSRB & RESISIC45 & SUN397 & SVHN & AVG \\
\midrule
LoRA & 71.79 & 66.28 & 96.93 & 40.40 & 86.37 & 70.59 & 93.38 & 75.11 \\
\midrule
\rowcolor{gray!10}
\multicolumn{9}{c}{\textbf{Rank Adjustment Based LoRA Variants}} \\
HiRA & 64.35 & 43.35 & 44.81 & 58.00 & 57.89 & 63.04 & 19.40 & 41.92 \\
RaSA & 70.73 & 63.03 & 96.22 & 29.46 & 84.56 & 69.89 & 92.29 & 72.31 \\
AdaLoRA & 64.64 & 45.32 & 50.83 & 44.00 & 60.33 & 63.88 & 47.41 & 47.55 \\
MELoRA & 80.74 & 75.05 & 98.35 & 96.85 & 94.75 & 76.16 & 96.29 & 88.31 \\
DenseLoRA & 74.38 & 71.76 & 97.28 & 63.28 & 89.25 & 71.82 & 94.12 & 80.27 \\
RandLoRA & 74.06 & 67.82 & 97.61 & 60.03 & 88.87 & 70.84 & 94.01 & 79.03 \\
ReLoRA & 70.39 & 62.34 & 96.13 & 29.06 & 84.41 & 69.71 & 92.38 & 72.06 \\
\midrule
\rowcolor{gray!10}
\multicolumn{9}{c}{\textbf{Optimization Process Adjustment Based LoRA Variants}} \\
DoRA & 72.18 & 66.44 & 97.06 & 42.59 & 86.81 & 70.73 & 93.48 & 75.61 \\
RsLoRA & 75.60 & 71.70 & 97.80 & 74.34 & 90.59 & 72.59 & 94.89 & 82.50 \\
LoRA+ & 83.32 & 77.66 & 98.50 & 98.17 & 95.22 & 77.56 & 96.91 & 89.62 \\
DeLoRA & 68.09 & 54.89 & 93.37 & 11.46 & 76.54 & 67.37 & 84.50 & 65.17 \\
\midrule
\rowcolor{gray!10}
\multicolumn{9}{c}{\textbf{Initialization Adjustment Based LoRA Variants}} \\
EVA & 78.51 & 74.79 & 98.06 & 91.71 & 92.56 & 73.97 & 95.72 & 86.47 \\
PiSSA & 75.36 & 69.52 & 97.83 & 76.98 & 91.03 & 72.72 & 95.06 & 82.64 \\
LoRA-GA & 82.33 & 75.53 & 98.52 & 96.96 & 94.70 & 76.01 & 96.49 & 88.65 \\
LoRA-One & 70.50 & 67.39 & 96.11 & 39.07 & 85.84 & 70.28 & 92.41 & 74.51 \\
NZLoRA & 77.70 & 72.82 & 98.26 & 85.82 & 92.25 & 73.36 & 95.55 & 85.11 \\
\midrule
\rowcolor{gray!10}
\multicolumn{9}{c}{\textbf{Mixture-of-Experts Integration Based Variants}} \\
GOAT & 71.40 & 66.65 & 96.48 & 48.87 & 87.49 & 69.36 & 92.91 & 76.17 \\
MoSLoRA & 70.69 & 63.09 & 96.09 & 30.37 & 84.48 & 69.92 & 92.36 & 72.43 \\
LoRAMoE & 78.72 & 72.02 & 98.15 & 91.92 & 92.70 & 74.26 & 95.62 & 86.20 \\
\bottomrule
\end{tabular*}
\end{table*}

\begin{table*}[t]
\centering
\footnotesize
\setlength{\tabcolsep}{5pt}
\renewcommand{\arraystretch}{1.1}
\caption{Performance on the seven image classification tasks with a learning rate of 2e-5.}
\label{tab:IC 2e-5}

\begin{tabular*}{0.9\textwidth}{@{\extracolsep{\fill}}lcccccccc@{}}
\toprule
Method & Cars & DTD & EuroSAT & GTSRB & RESISIC45 & SUN397 & SVHN & AVG \\
\midrule
LoRA & 75.29 & 71.44 & 97.63 & 71.57 & 90.41 & 72.43 & 94.90 & 81.95 \\
\midrule
\rowcolor{gray!10}
\multicolumn{9}{c}{\textbf{Rank Adjustment Based LoRA Variants}} \\
HiRA & 64.83 & 44.68 & 53.80 & 40.00 & 60.19 & 63.69 & 35.93 & 46.22 \\
RaSA & 73.91 & 69.31 & 97.33 & 59.29 & 88.97 & 71.71 & 94.26 & 79.25 \\
AdaLoRA & 66.14 & 51.06 & 77.07 & 50.00 & 67.76 & 67.46 & 75.92 & 57.99 \\
MELoRA & 82.35 & 76.44 & 98.57 & 98.25 & 95.59 & 77.12 & 96.83 & 89.31 \\
DenseLoRA & 77.84 & 73.56 & 98.04 & 83.28 & 91.81 & 74.21 & 95.39 & 84.88 \\
RandLoRA & 77.50 & 73.62 & 98.20 & 83.22 & 91.75 & 72.79 & 95.31 & 84.63 \\
ReLoRA & 73.45 & 69.41 & 97.33 & 59.98 & 88.49 & 71.26 & 94.19 & 79.16 \\
\midrule
\rowcolor{gray!10}
\multicolumn{9}{c}{\textbf{Optimization Process Adjustment Based LoRA Variants}} \\
DoRA & 75.59 & 71.81 & 97.80 & 73.37 & 90.56 & 72.61 & 94.98 & 82.39 \\
RsLoRA & 78.97 & 74.26 & 98.13 & 89.09 & 92.79 & 74.92 & 95.91 & 86.30 \\
LoRA+ & 85.95 & 79.68 & 98.76 & 98.77 & 95.71 & 78.45 & 97.28 & 90.66 \\
DeLoRA & 69.73 & 60.16 & 95.83 & 27.21 & 83.00 & 68.54 & 90.98 & 70.78 \\
\midrule
\rowcolor{gray!10}
\multicolumn{9}{c}{\textbf{Initialization Adjustment Based LoRA Variants}} \\
EVA & 81.05 & 76.17 & 98.37 & 96.75 & 94.30 & 75.43 & 96.34 & 88.34 \\
PiSSA & 77.95 & 71.17 & 98.30 & 89.37 & 92.90 & 74.52 & 95.91 & 85.73 \\
LoRA-GA & 84.17 & 77.13 & 98.52 & 98.34 & 95.63 & 76.89 & 96.85 & 89.65 \\
LoRA-One & 73.39 & 72.77 & 97.26 & 63.67 & 89.56 & 72.01 & 94.29 & 80.42 \\
NZLoRA & 80.13 & 74.47 & 98.41 & 95.42 & 94.11 & 75.48 & 96.17 & 87.74 \\
\midrule
\rowcolor{gray!10}
\multicolumn{9}{c}{\textbf{Mixture-of-Experts Integration Based Variants}} \\
GOAT & 76.48 & 72.34 & 97.54 & 78.95 & 91.30 & 72.21 & 6.70 & 70.79 \\
MoSLoRA & 74.05 & 70.05 & 97.19 & 63.17 & 89.27 & 71.81 & 94.25 & 79.97 \\
LoRAMoE & 79.42 & 74.26 & 98.20 & 95.86 & 94.29 & 75.75 & 96.01 & 87.68 \\
\bottomrule
\end{tabular*}
\end{table*}

\begin{table*}[t]
\centering
\footnotesize
\setlength{\tabcolsep}{5pt}
\renewcommand{\arraystretch}{1.1}
\caption{Performance on the image classification tasks with a learning rate of 5e-5.}
\label{tab:IC 5e-5}

\begin{tabular*}{0.9\textwidth}{@{\extracolsep{\fill}}lcccccccc@{}}
\toprule
Method & Cars & DTD & EuroSAT & GTSRB & RESISIC45 & SUN397 & SVHN & AVG \\
\midrule
LoRA & 80.08 & 74.52 & 98.07 & 92.15 & 93.49 & 75.68 & 96.15 & 87.16 \\
\midrule
\rowcolor{gray!10}
\multicolumn{9}{c}{\textbf{Rank Adjustment Based LoRA Variants}} \\
HiRA & 65.89 & 50.64 & 77.04 & 53.00 & 67.22 & 66.16 & 70.83 & 56.90 \\
RaSA & 78.72 & 74.63 & 98.09 & 87.57 & 92.68 & 74.82 & 95.83 & 86.05 \\
AdaLoRA & 69.39 & 64.47 & 95.00 & 17.81 & 82.87 & 69.22 & 91.61 & 70.05 \\
MELoRA & 83.16 & 77.13 & 98.80 & 98.71 & 96.11 & 76.52 & 97.15 & 89.65 \\
DenseLoRA & 81.63 & 74.79 & 98.41 & 95.61 & 94.19 & 76.57 & 96.32 & 88.22 \\
RandLoRA & 81.61 & 75.74 & 98.54 & 95.99 & 94.48 & 75.71 & 96.40 & 88.35 \\
ReLoRA & 78.10 & 74.68 & 98.04 & 86.92 & 92.19 & 73.98 & 95.63 & 85.65 \\
\midrule
\rowcolor{gray!10}
\multicolumn{9}{c}{\textbf{Optimization Process Adjustment Based LoRA Variants}} \\
DoRA & 80.25 & 74.68 & 98.11 & 92.80 & 93.48 & 75.87 & 96.19 & 87.34 \\
RsLoRA & 82.08 & 75.74 & 98.35 & 97.11 & 94.81 & 77.10 & 96.62 & 88.83 \\
LoRA+ & 87.30 & 79.04 & 98.70 & 98.90 & 96.41 & 78.11 & 97.36 & 90.83 \\
DeLoRA & 73.24 & 68.51 & 97.15 & 62.95 & 88.73 & 70.76 & 93.85 & 79.31 \\
\midrule
\rowcolor{gray!10}
\multicolumn{9}{c}{\textbf{Initialization Adjustment Based LoRA Variants}} \\
EVA & 82.42 & 77.55 & 98.44 & 98.59 & 95.37 & 76.50 & 96.77 & 89.38 \\
PiSSA & 81.22 & 74.20 & 98.57 & 96.92 & 94.84 & 76.57 & 96.63 & 88.42 \\
LoRA-GA & 85.93 & 77.87 & 98.61 & 98.96 & 95.81 & 77.28 & 97.02 & 90.21 \\
LoRA-One & 78.46 & 75.43 & 98.11 & 91.14 & 92.89 & 75.11 & 95.89 & 86.72 \\
NZLoRA & 82.55 & 76.60 & 98.61 & 98.60 & 95.71 & 77.39 & 96.87 & 89.48 \\
\midrule
\rowcolor{gray!10}
\multicolumn{9}{c}{\textbf{Mixture-of-Experts Integration Based Variants}} \\
GOAT & 81.06 & 75.37 & 98.06 & 94.53 & 93.97 & 75.59 & 96.00 & 87.80 \\
MoSLoRA & 79.06 & 74.36 & 98.04 & 89.83 & 92.79 & 75.22 & 95.85 & 86.45 \\
LoRAMoE & 80.11 & 73.30 & 98.43 & 95.50 & 94.63 & 75.94 & 95.68 & 87.66 \\
\bottomrule
\end{tabular*}
\end{table*}

\begin{table*}[t]
\centering
\footnotesize
\setlength{\tabcolsep}{5pt}
\renewcommand{\arraystretch}{1.1}
\caption{Performance on seven image classification tasks with a learning rate of 1e-4.}
\label{tab:IC 1e-4}

\begin{tabular*}{0.9\textwidth}{@{\extracolsep{\fill}}lcccccccc@{}}
\toprule
Method & Cars & DTD & EuroSAT & GTSRB & RESISIC45 & SUN397 & SVHN & AVG \\
\midrule
LoRA & 82.22 & 75.85 & 98.41 & 97.08 & 94.67 & 77.18 & 96.58 & 88.86 \\
\midrule
\rowcolor{gray!10}
\multicolumn{9}{c}{\textbf{Rank Adjustment Based LoRA Variants}} \\
HiRA & 68.08 & 56.06 & 89.80 & 5.91 & 76.21 & 68.12 & 85.80 & 64.28 \\
RaSA & 81.52 & 76.01 & 98.26 & 96.36 & 94.11 & 76.79 & 96.36 & 88.49 \\
AdaLoRA & 73.29 & 74.31 & 97.06 & 58.87 & 88.51 & 71.76 & 94.21 & 79.72 \\
MELoRA & 82.66 & 74.84 & 98.57 & 98.68 & 95.59 & 74.59 & 96.98 & 88.84 \\
DenseLoRA & 83.80 & 77.18 & 98.43 & 97.87 & 95.22 & 77.37 & 96.75 & 89.52 \\
RandLoRA & 83.04 & 76.81 & 98.65 & 98.37 & 95.51 & 77.08 & 96.85 & 89.47 \\
ReLoRA & 81.18 & 76.44 & 98.30 & 95.56 & 93.92 & 76.04 & 96.31 & 88.25 \\
\midrule
\rowcolor{gray!10}
\multicolumn{9}{c}{\textbf{Optimization Process Adjustment Based LoRA Variants}} \\
DoRA & 82.20 & 76.01 & 98.41 & 97.22 & 94.70 & 77.27 & 96.64 & 88.92 \\
RsLoRA & 83.78 & 78.40 & 98.63 & 98.85 & 95.51 & 77.85 & 96.98 & 90.00 \\
LoRA+ & 87.40 & 72.71 & 98.70 & 98.84 & 95.54 & 76.61 & 96.91 & 89.53 \\
DeLoRA & 76.53 & 72.77 & 97.94 & 83.49 & 91.68 & 72.78 & 95.22 & 84.34 \\
\midrule
\rowcolor{gray!10}
\multicolumn{9}{c}{\textbf{Initialization Adjustment Based LoRA Variants}} \\
EVA & 83.99 & 77.45 & 98.33 & 98.74 & 95.79 & 77.07 & 97.07 & 89.78 \\
PiSSA & 83.34 & 77.66 & 98.67 & 98.27 & 95.63 & 77.10 & 97.00 & 89.67 \\
LoRA-GA & 85.04 & 75.74 & 98.61 & 98.80 & 95.57 & 76.49 & 96.94 & 89.60 \\
LoRA-One & 81.30 & 76.38 & 98.43 & 97.01 & 94.38 & 76.82 & 96.41 & 88.68 \\
NZLoRA & 84.77 & 78.72 & 98.72 & 99.07 & 96.00 & 78.20 & 97.09 & 90.37 \\
\midrule
\rowcolor{gray!10}
\multicolumn{9}{c}{\textbf{Mixture-of-Experts Integration Based Variants}} \\
GOAT & 82.43 & 75.59 & 98.26 & 96.53 & 95.03 & 76.82 & 96.25 & 88.70 \\
MoSLoRA & 81.81 & 76.01 & 98.35 & 96.56 & 94.32 & 76.75 & 96.39 & 88.60 \\
LoRAMoE & 77.85 & 72.23 & 97.26 & 89.53 & 92.06 & 73.31 & 91.88 & 84.87 \\
\bottomrule
\end{tabular*}
\end{table*}

\begin{table*}[t]
\centering
\footnotesize
\setlength{\tabcolsep}{5pt}
\renewcommand{\arraystretch}{1.1}
\caption{Performance on the seven image classification tasks with a learning rate of 2e-4.}
\label{tab:IC 2e-4}

\begin{tabular*}{0.9\textwidth}{@{\extracolsep{\fill}}lcccccccc@{}}
\toprule
Method & Cars & DTD & EuroSAT & GTSRB & RESISIC45 & SUN397 & SVHN & AVG \\
\midrule
LoRA & 84.07 & 78.62 & 98.57 & 98.50 & 95.52 & 77.90 & 97.02 & 90.03 \\
\midrule
\rowcolor{gray!10}
\multicolumn{9}{c}{\textbf{Rank Adjustment Based LoRA Variants}} \\
HiRA & 70.87 & 66.06 & 95.69 & 24.93 & 84.62 & 69.76 & 92.49 & 72.06 \\
RaSA & 83.17 & 77.50 & 98.50 & 98.11 & 95.29 & 77.63 & 96.73 & 89.56 \\
AdaLoRA & 78.12 & 75.05 & 97.65 & 86.07 & 91.94 & 74.75 & 95.72 & 85.61 \\
MELoRA & 79.34 & 70.00 & 98.37 & 98.14 & 94.14 & 69.75 & 96.12 & 86.55 \\
DenseLoRA & 85.75 & 78.40 & 98.70 & 98.45 & 95.90 & 78.35 & 97.08 & 90.38 \\
RandLoRA & 85.24 & 78.19 & 98.61 & 99.15 & 96.25 & 78.03 & 97.05 & 90.36 \\
ReLoRA & 82.84 & 78.46 & 98.44 & 97.80 & 94.98 & 77.06 & 96.78 & 89.48 \\
\midrule
\rowcolor{gray!10}
\multicolumn{9}{c}{\textbf{Optimization Process Adjustment Based LoRA Variants}} \\
DoRA & 84.12 & 77.77 & 98.65 & 98.76 & 95.59 & 77.88 & 97.06 & 89.98 \\
RsLoRA & 85.92 & 78.88 & 98.63 & 99.19 & 96.14 & 78.02 & 97.30 & 90.58 \\
LoRA+ & 1.12 & 4.84 & 70.54 & 23.14 & 10.21 & 3.15 & 19.59 & 18.94 \\
DeLoRA & 79.69 & 74.47 & 98.37 & 94.96 & 93.52 & 74.80 & 95.95 & 87.39 \\
\midrule
\rowcolor{gray!10}
\multicolumn{9}{c}{\textbf{Initialization Adjustment Based LoRA Variants}} \\
EVA & 83.92 & 76.22 & 98.56 & 98.53 & 95.81 & 76.63 & 96.98 & 89.52 \\
PiSSA & 85.09 & 78.30 & 98.83 & 98.69 & 95.73 & 77.22 & 97.07 & 90.13 \\
LoRA-GA & 82.74 & 71.44 & 98.43 & 98.72 & 94.81 & 73.43 & 96.96 & 88.08 \\
LoRA-One & 83.11 & 78.83 & 98.61 & 98.48 & 95.29 & 77.64 & 96.90 & 89.84 \\
NZLoRA & 85.66 & 77.98 & 98.80 & 98.82 & 96.22 & 77.52 & 97.36 & 90.34 \\
\midrule
\rowcolor{gray!10}
\multicolumn{9}{c}{\textbf{Mixture-of-Experts Integration Based Variants}} \\
GOAT & 83.46 & 77.45 & 98.56 & 97.18 & 95.32 & 77.39 & 96.62 & 89.43 \\
MoSLoRA & 83.34 & 77.39 & 98.46 & 98.12 & 95.46 & 77.75 & 96.86 & 89.63 \\
LoRAMoE & 1.41 & 9.26 & 56.19 & 20.92 & 16.17 & 1.43 & 19.59 & 17.85 \\
\bottomrule
\end{tabular*}
\end{table*}

\begin{table*}[t]
\centering
\footnotesize
\setlength{\tabcolsep}{5pt}
\renewcommand{\arraystretch}{1.1}
\caption{Performance on the seven image classification tasks with a learning rate of 5e-4.}
\label{tab:IC 5e-4}

\begin{tabular*}{0.9\textwidth}{@{\extracolsep{\fill}}lcccccccc@{}}
\toprule
Method & Cars & DTD & EuroSAT & GTSRB & RESISIC45 & SUN397 & SVHN & AVG \\
\midrule
LoRA & 86.78 & 78.62 & 98.74 & 98.93 & 96.27 & 78.68 & 97.20 & 90.75 \\
\midrule
\rowcolor{gray!10}
\multicolumn{9}{c}{\textbf{Rank Adjustment Based LoRA Variants}} \\
HiRA & 76.46 & 73.88 & 97.65 & 78.80 & 90.60 & 73.19 & 94.86 & 83.63 \\
RaSA & 86.36 & 78.62 & 98.59 & 98.94 & 96.05 & 78.68 & 97.16 & 90.63 \\
AdaLoRA & 81.77 & 76.86 & 98.48 & 96.99 & 94.16 & 77.08 & 96.53 & 88.84 \\
MELoRA & 2.46 & 11.06 & 74.24 & 39.00 & 38.25 & 6.01 & 20.44 & 27.35 \\
DenseLoRA & 87.10 & 78.72 & 98.83 & 99.17 & 96.32 & 79.01 & 97.41 & 90.94 \\
RandLoRA & 86.38 & 77.93 & 98.70 & 99.22 & 96.25 & 77.79 & 97.41 & 90.53 \\
ReLoRA & 85.59 & 79.31 & 98.67 & 98.49 & 95.90 & 77.82 & 97.07 & 90.41 \\
\midrule
\rowcolor{gray!10}
\multicolumn{9}{c}{\textbf{Optimization Process Adjustment Based LoRA Variants}} \\
DoRA & 86.72 & 78.94 & 98.61 & 99.06 & 96.05 & 78.82 & 97.27 & 90.78 \\
RsLoRA & 87.19 & 76.28 & 98.72 & 98.90 & 96.24 & 77.68 & 97.25 & 90.32 \\
LoRA+ & 1.11 & 3.24 & 33.13 & 9.90 & 9.44 & 1.23 & 19.72 & 11.11 \\
DeLoRA & 82.03 & 76.76 & 98.59 & 98.08 & 95.40 & 76.87 & 96.65 & 89.20 \\
\midrule
\rowcolor{gray!10}
\multicolumn{9}{c}{\textbf{Initialization Adjustment Based LoRA Variants}} \\
EVA & 83.17 & 73.62 & 98.41 & 98.40 & 95.62 & 75.15 & 97.13 & 88.79 \\
PiSSA & 85.10 & 73.67 & 98.48 & 98.77 & 95.56 & 75.53 & 97.15 & 89.18 \\
LoRA-GA & 77.12 & 60.53 & 97.52 & 98.20 & 93.17 & 65.33 & 96.44 & 84.04 \\
LoRA-One & 86.66 & 79.63 & 98.59 & 99.14 & 96.38 & 78.54 & 97.33 & 90.90 \\
NZLoRA & 83.92 & 71.54 & 98.57 & 98.74 & 95.54 & 73.31 & 96.83 & 88.35 \\
\midrule
\rowcolor{gray!10}
\multicolumn{9}{c}{\textbf{Mixture-of-Experts Integration Based Variants}} \\
GOAT & 84.33 & 77.23 & 98.28 & 96.30 & 94.86 & 76.38 & 28.61 & 79.43 \\
MoSLoRA & 85.23 & 78.56 & 98.63 & 98.75 & 96.06 & 78.31 & 97.10 & 90.38 \\
LoRAMoE & 87.00 & 4.41 & 59.41 & 6.98 & 11.98 & 1.02 & 19.59 & 14.89 \\
\bottomrule
\end{tabular*}
\end{table*}

\begin{table*}[t]
\centering
\footnotesize
\setlength{\tabcolsep}{5pt}
\renewcommand{\arraystretch}{1.1}
\caption{Performance on the seven image classification tasks with a learning rate of 1e-3.}
\label{tab:IC 1e-3}

\begin{tabular*}{0.9\textwidth}{@{\extracolsep{\fill}}lcccccccc@{}}
\toprule
Method & Cars & DTD & EuroSAT & GTSRB & RESISIC45 & SUN397 & SVHN & AVG \\
\midrule
LoRA & 87.41 & 77.07 & 98.74 & 99.03 & 96.19 & 77.71 & 97.27 & 90.49 \\
\midrule
\rowcolor{gray!10}
\multicolumn{9}{c}{\textbf{Rank Adjustment Based LoRA Variants}} \\
HiRA & 80.94 & 75.64 & 98.24 & 93.79 & 93.33 & 75.85 & 96.05 & 87.69 \\
RaSA & 87.71 & 77.93 & 98.69 & 98.92 & 96.35 & 78.35 & 97.30 & 90.75 \\
AdaLoRA & 83.73 & 78.51 & 98.46 & 98.11 & 95.14 & 77.86 & 96.77 & 89.80 \\
MELoRA & 1.79 & 11.76 & 72.09 & 27.47 & 33.59 & 2.69 & 31.39 & 25.83 \\
DenseLoRA & 88.15 & 80.21 & 98.74 & 98.92 & 96.41 & 78.65 & 97.26 & 91.19 \\
RandLoRA & 86.03 & 74.57 & 98.70 & 99.30 & 96.14 & 75.84 & 97.39 & 89.71 \\
ReLoRA & 85.66 & 77.34 & 98.61 & 98.82 & 95.84 & 76.62 & 97.08 & 90.00 \\
\midrule
\rowcolor{gray!10}
\multicolumn{9}{c}{\textbf{Optimization Process Adjustment Based LoRA Variants}} \\
DoRA & 88.07 & 77.71 & 98.69 & 98.92 & 96.43 & 77.73 & 97.35 & 90.70 \\
RsLoRA & 86.46 & 72.29 & 98.57 & 98.50 & 95.57 & 75.52 & 96.52 & 89.06 \\
LoRA+ & 99.00 & 3.35 & 37.30 & 11.03 & 8.14 & 30.00 & 19.59 & 11.53 \\
DeLoRA & 83.83 & 78.03 & 98.65 & 98.73 & 95.92 & 77.34 & 97.07 & 89.94 \\
\midrule
\rowcolor{gray!10}
\multicolumn{9}{c}{\textbf{Initialization Adjustment Based LoRA Variants}} \\
EVA & 83.97 & 66.76 & 98.22 & 98.15 & 94.86 & 74.21 & 96.88 & 87.58 \\
PiSSA & 83.40 & 67.61 & 98.41 & 98.38 & 94.73 & 71.77 & 96.75 & 87.29 \\
LoRA-GA & 58.91 & 44.95 & 96.57 & 95.15 & 89.60 & 48.79 & 95.29 & 75.61 \\
LoRA-One & 87.53 & 77.29 & 98.69 & 98.52 & 96.38 & 78.03 & 97.19 & 90.52 \\
NZLoRA & 77.13 & 59.47 & 97.46 & 97.97 & 92.94 & 65.55 & 95.33 & 83.69 \\
\midrule
\rowcolor{gray!10}
\multicolumn{9}{c}{\textbf{Mixture-of-Experts Integration Based Variants}} \\
GOAT & 1.06 & 6.28 & 69.50 & 11.47 & 8.84 & 67.34 & 32.97 & 28.21 \\
MoSLoRA & 86.94 & 78.72 & 98.63 & 98.92 & 95.95 & 78.50 & 97.34 & 90.71 \\
LoRAMoE & 92.00 & 4.31 & 19.02 & 13.71 & 6.51 & 38.00 & 19.59 & 9.21 \\
\bottomrule
\end{tabular*}
\end{table*}

\begin{table*}[t]
\centering
\footnotesize
\setlength{\tabcolsep}{5pt}
\renewcommand{\arraystretch}{1.1}
\caption{Performance on the seven image classification tasks with a learning rate of 2e-3.}
\label{tab:IC 2e-3}

\begin{tabular*}{0.9\textwidth}{@{\extracolsep{\fill}}lcccccccc@{}}
\toprule
Method & Cars & DTD & EuroSAT & GTSRB & RESISIC45 & SUN397 & SVHN & AVG \\
\midrule
LoRA & 86.01 & 69.95 & 98.59 & 98.53 & 95.22 & 75.09 & 96.20 & 88.51 \\
\midrule
\rowcolor{gray!10}
\multicolumn{9}{c}{\textbf{Rank Adjustment Based LoRA Variants}} \\
HiRA & 83.16 & 77.07 & 98.39 & 97.46 & 94.71 & 77.35 & 96.70 & 89.26 \\
RaSA & 87.36 & 74.89 & 98.57 & 98.80 & 96.16 & 77.75 & 97.30 & 90.12 \\
AdaLoRA & 85.05 & 78.35 & 98.80 & 98.65 & 95.70 & 78.44 & 96.90 & 90.27 \\
MELoRA & 1.58 & 8.62 & 61.13 & 24.15 & 17.40 & 1.26 & 22.17 & 19.47 \\
DenseLoRA & 55.00 & 3.14 & 98.70 & 98.74 & 5.54 & 72.00 & 19.59 & 32.43 \\
RandLoRA & 82.37 & 67.82 & 98.28 & 99.18 & 95.06 & 70.66 & 97.03 & 87.20 \\
ReLoRA & 82.30 & 69.26 & 97.98 & 98.32 & 94.44 & 71.15 & 96.27 & 87.10 \\
\midrule
\rowcolor{gray!10}
\multicolumn{9}{c}{\textbf{Optimization Process Adjustment Based LoRA Variants}} \\
DoRA & 86.66 & 70.43 & 98.48 & 98.93 & 94.98 & 75.32 & 96.70 & 88.79 \\
RsLoRA & 1.83 & 5.32 & 56.26 & 22.32 & 14.03 & 65.00 & 32.10 & 18.93 \\
LoRA+ & 66.00 & 2.13 & 35.69 & 7.95 & 2.24 & 25.00 & 19.59 & 9.79 \\
DeLoRA & 85.00 & 78.78 & 98.70 & 98.76 & 95.86 & 78.06 & 97.19 & 90.34 \\
\midrule
\rowcolor{gray!10}
\multicolumn{9}{c}{\textbf{Initialization Adjustment Based LoRA Variants}} \\
EVA & 80.00 & 57.39 & 97.28 & 94.82 & 92.90 & 70.24 & 94.23 & 83.84 \\
PiSSA & 77.99 & 54.68 & 96.91 & 96.72 & 92.60 & 64.44 & 94.33 & 82.52 \\
LoRA-GA & 2.72 & 13.72 & 83.43 & 49.11 & 50.32 & 4.59 & 53.62 & 36.79 \\
LoRA-One & 86.47 & 71.65 & 98.30 & 98.48 & 95.32 & 75.50 & 95.95 & 88.81 \\
NZLoRA & 2.50 & 12.02 & 67.85 & 41.00 & 41.37 & 4.29 & 46.39 & 30.77 \\
\midrule
\rowcolor{gray!10}
\multicolumn{9}{c}{\textbf{Mixture-of-Experts Integration Based Variants}} \\
GOAT & 83.00 & 3.30 & 40.72 & 9.94 & 3.98 & 31.00 & 19.63 & 11.24 \\
MoSLoRA & 87.20 & 77.93 & 98.70 & 99.11 & 96.33 & 78.68 & 97.15 & 90.73 \\
LoRAMoE & 88.00 & 2.39 & 25.39 & 7.18 & 5.35 & 38.00 & 19.55 & 8.73 \\
\bottomrule
\end{tabular*}
\end{table*}

\begin{table*}[t]
\centering
\footnotesize
\setlength{\tabcolsep}{5pt}
\renewcommand{\arraystretch}{1.1}
\caption{Performance on image classification tasks with a learning rate of 5e-3.}
\label{tab:visual-bench}

\begin{tabular*}{0.9\textwidth}{@{\extracolsep{\fill}}lcccccccc@{}}
\toprule
Method & Cars & DTD & EuroSAT & GTSRB & RESISIC45 & SUN397 & SVHN & AVG \\
\midrule
LoRA & 1.24 & 4.57 & 50.63 & 14.43 & 6.06 & 79.00 & 19.59 & 13.90 \\
\midrule
\rowcolor{gray!10}
\multicolumn{9}{c}{\textbf{Rank Adjustment Based LoRA Variants}} \\
HiRA & 85.45 & 79.10 & 98.67 & 98.73 & 96.19 & 78.78 & 97.28 & 90.60 \\
RaSA & 72.00 & 4.73 & 27.61 & 11.58 & 20.67 & 2.27 & 19.60 & 12.45 \\
AdaLoRA & 85.60 & 79.15 & 98.63 & 98.67 & 96.06 & 78.31 & 97.04 & 90.49 \\
MELoRA & 1.31 & 8.35 & 49.65 & 12.95 & 13.06 & 1.10 & 28.53 & 16.42 \\
DenseLoRA & 92.00 & 4.79 & 45.63 & 11.80 & 7.98 & 73.00 & 19.59 & 13.06 \\
RandLoRA & 70.51 & 51.65 & 97.19 & 98.50 & 91.35 & 55.24 & 95.77 & 80.03 \\
ReLoRA & 1.09 & 4.04 & 44.91 & 6.91 & 7.95 & 1.74 & 19.59 & 12.32 \\
\midrule
\rowcolor{gray!10}
\multicolumn{9}{c}{\textbf{Optimization Process Adjustment Based LoRA Variants}} \\
DoRA & 86.00 & 2.55 & 50.76 & 14.82 & 7.13 & 1.90 & 19.59 & 13.94 \\
RsLoRA & 72.00 & 5.80 & 31.15 & 10.93 & 13.44 & 34.00 & 22.60 & 12.14 \\
LoRA+ & 81.00 & 2.13 & 11.30 & 48.00 & 2.24 & 30.00 & 19.59 & 5.26 \\
DeLoRA & 86.25 & 79.36 & 98.78 & 98.60 & 96.13 & 78.51 & 97.36 & 90.71 \\
\midrule
\rowcolor{gray!10}
\multicolumn{9}{c}{\textbf{Initialization Adjustment Based LoRA Variants}} \\
EVA & 91.00 & 4.10 & 41.96 & 10.97 & 19.57 & 1.26 & 19.62 & 14.06 \\
PiSSA & 1.14 & 5.21 & 49.93 & 19.82 & 7.49 & 1.21 & 38.00 & 17.54 \\
LoRA-GA & 1.82 & 10.96 & 59.17 & 25.63 & 16.32 & 1.48 & 35.63 & 21.57 \\
LoRA-One & 67.00 & 5.59 & 48.48 & 5.98 & 8.43 & 72.00 & 27.11 & 13.85 \\
NZLoRA & 95.00 & 5.48 & 57.37 & 13.08 & 11.60 & 69.00 & 33.16 & 17.48 \\
\midrule
\rowcolor{gray!10}
\multicolumn{9}{c}{\textbf{Mixture-of-Experts Integration Based Variants}} \\
GOAT & 88.00 & 2.45 & 24.76 & 6.12 & 3.46 & 30.00 & 19.59 & 8.22 \\
MoSLoRA & 95.00 & 6.54 & 57.85 & 15.85 & 9.44 & 74.00 & 32.67 & 17.72 \\
LoRAMoE & 88.00 & 2.93 & 23.35 & 9.91 & 4.19 & 27.00 & 19.59 & 8.73 \\
\bottomrule
\end{tabular*}
\end{table*}

\begin{table*}[t]
\centering
\footnotesize
\setlength{\tabcolsep}{5pt}
\renewcommand{\arraystretch}{1.1}
\caption{Performance on GSM8k.}
\label{tab:gsm8k}

\begin{tabular*}{0.9\textwidth}{@{\extracolsep{\fill}}lcccccccc@{}}
\toprule
Method & 1e-6 & 1e-5 & 2e-5 & 5e-5 & 1e-4 & 2e-4 & 5e-4 & 1e-3 \\
\midrule
LoRA & 52.84 & 64.37 & 66.94 & 70.13 & 73.16 & 72.86 & 75.51 & 72.10 \\
\midrule
\rowcolor{gray!10}
\multicolumn{9}{c}{\textbf{Rank Adjustment Based LoRA Variants}} \\
HiRA & 6.44 & 7.44 & 30.02 & 49.73 & 54.81 & 62.09 & 66.49 & 72.10 \\
RaSA & 48.98 & 61.87 & 63.84 & 68.92 & 71.04 & 72.48 & 73.31 & 74.30 \\
AdaLoRA & 5.84 & 34.65 & 49.32 & 59.51 & 61.49 & 65.58 & 70.35 & 71.72 \\
MELoRA & 62.55 & 71.72 & 72.02 & 70.36 & 64.52 & 57.01 & 1.74 & 2.05 \\
DenseLoRA & 50.19 & 63.76 & 67.85 & 68.84 & 72.86 & 72.33 & 74.22 & 74.75 \\
RandLoRA & 46.32 & 64.75 & 68.92 & 74.22 & 74.53 & 75.59 & 71.27 & 65.43 \\
ReLoRA & 55.04 & 65.81 & 69.6 & 71.87 & 71.65 & 67.17 & 41.93 & 1.52 \\
\midrule
\rowcolor{gray!10}
\multicolumn{9}{c}{\textbf{Optimization Process Adjustment Based LoRA Variants}} \\
DoRA & 54.69 & 64.67 & 67.25 & 71.80 & 73.09 & 74.15 & 74.53 & 72.25 \\
RsLoRA & 54.51 & 65.73 & 69.45 & 71.87 & 73.92 & 74.68 & 69.75 & 67.55 \\
LoRA+ & 60.12 & 70.28 & 72.02 & 74.68 & 75.59 & 72.25 & 64.67 & 1.59 \\
DeLoRA & 30.86 & 39.68 & 42.34 & 49.22 & 66.57 & 70.20 & 70.74 & 72.02 \\
\midrule
\rowcolor{gray!10}
\multicolumn{9}{c}{\textbf{Initialization Adjustment Based LoRA Variants}} \\
EVA & 60.12 & 69.67 & 71.87 & 71.49 & 73.09 & 73.84 & 74.30 & 71.65 \\
PiSSA & 57.39 & 69.14 & 71.27 & 72.18 & 72.86 & 72.71 & 71.11 & 64.67 \\
LoRA-GA & 64.06 & 70.36 & 70.51 & 72.02 & 71.95 & 67.63 & 64.22 & 48.45 \\
LoRA-One & 52.39 & 64.59 & 67.70 & 70.28 & 72.18 & 73.01 & 74.68 & 71.27 \\
NZLoRA & 53.68 & 63.76 & 67.63 & 71.87 & 73.31 & 71.27 & 70.81 & 64.22 \\
\midrule
\rowcolor{gray!10}
\multicolumn{9}{c}{\textbf{Mixture-of-Experts Integration Based Variants}} \\
GOAT & 48.98 & 67.25 & 67.63 & 68.84 & 71.65 & 71.27 & 52.39 & 2.12 \\
MoSLoRA & 48.52 & 63.31 & 66.03 & 69.37 & 69.98 & 72.71 & 74.00 & 74.68 \\
LoRAMoE & 46.32 & 67.93 & 69.83 & 69.22 & 59.97 & 34.60 & 1.67 & 1.52 \\
\bottomrule
\end{tabular*}
\end{table*}

\begin{table*}[t]
\centering
\footnotesize
\setlength{\tabcolsep}{5pt}
\renewcommand{\arraystretch}{1.1}
\caption{Performance on HumanEval.}
\label{tab:humaneval}

\begin{tabular*}{0.9\textwidth}{@{\extracolsep{\fill}}lcccccccc@{}}
\toprule
Method & 1e-6 & 1e-5 & 2e-5 & 5e-5 & 1e-4 & 2e-4 & 5e-4 & 1e-3 \\
\midrule
LoRA & 9.30 & 41.31 & 41.46 & 44.82 & 45.27 & 48.09 & 48.17 & 42.07 \\
\midrule
\rowcolor{gray!10}
\multicolumn{9}{c}{\textbf{Rank Adjustment Based LoRA Variants}} \\
HiRA & 0.00 & 3.20 & 4.04 & 11.05 & 23.09 & 39.18 & 43.67 & 45.12 \\
RaSA & 6.48 & 40.47 & 43.67 & 42.91 & 46.04 & 47.71 & 49.31 & 49.31 \\
AdaLoRA & 0.08 & 4.19 & 7.24 & 26.52 & 39.25 & 41.62 & 45.73 & 44.74 \\
MELoRA & 18.14 & 45.81 & 45.88 & 42.45 & 35.59 & 26.45 & 0.00 & 0.00 \\
DenseLoRA & 7.77 & 38.41 & 42.61 & 45.05 & 45.20 & 46.19 & 48.02 & 46.72 \\
RandLoRA & 7.16 & 29.73 & 33.99 & 44.51 & 48.63 & 50.30 & 47.79 & 44.89 \\
ReLoRA & 44.21 & 44.97 & 47.03 & 42.07 & 30.26 & 10.44 & 1.07 & 0.00 \\
\midrule
\rowcolor{gray!10}
\multicolumn{9}{c}{\textbf{Optimization Process Adjustment Based LoRA Variants}} \\
DoRA & 9.07 & 42.99 & 42.91 & 46.34 & 46.72 & 46.57 & 47.18 & 44.74 \\
RsLoRA & 12.73 & 43.67 & 42.53 & 45.66 & 47.56 & 47.94 & 44.13 & 40.02 \\
LoRA+ & 28.28 & 47.64 & 46.57 & 46.04 & 46.19 & 44.36 & 37.35 & 0.00 \\
DeLoRA & 4.88 & 7.01 & 10.21 & 23.02 & 37.65 & 44.21 & 45.50 & 44.59 \\
\midrule
\rowcolor{gray!10}
\multicolumn{9}{c}{\textbf{Initialization Adjustment Based LoRA Variants}} \\
EVA & 15.55 & 39.63 & 46.42 & 46.49 & 45.43 & 45.20 & 45.96 & 44.89 \\
PiSSA & 7.01 & 28.51 & 26.14 & 45.20 & 45.66 & 47.41 & 41.54 & 35.37 \\
LoRA-GA & 37.04 & 44.59 & 44.28 & 45.96 & 46.65 & 43.75 & 39.41 & 30.72 \\
LoRA-One & 11.66 & 42.91 & 44.28 & 45.35 & 43.83 & 45.43 & 47.79 & 46.11 \\
NZLoRA & 7.16 & 24.70 & 36.74 & 45.81 & 46.80 & 47.64 & 46.27 & 39.10 \\
\midrule
\rowcolor{gray!10}
\multicolumn{9}{c}{\textbf{Mixture-of-Experts Integration Based Variants}} \\
GOAT & 5.49 & 42.53 & 43.83 & 45.58 & 46.19 & 45.05 & 0.00 & 0.00 \\
MoSLoRA & 5.79 & 43.06 & 44.97 & 46.04 & 45.27 & 47.79 & 46.49 & 46.80 \\
LoRAMoE & 9.91 & 45.05 & 43.05 & 43.45 & 37.20 & 0.00 & 0.00 & 0.00 \\
\bottomrule
\end{tabular*}
\end{table*}

\clearpage

\end{document}